\documentclass{article}

\PassOptionsToPackage{numbers, compress}{natbib}

\usepackage[preprint]{neurips_distshift_2022}

\usepackage[utf8]{inputenc} 
\usepackage[T1]{fontenc}    
\usepackage{hyperref}       
\usepackage{url}            
\usepackage{booktabs}       
\usepackage{amsfonts}       
\usepackage{nicefrac}       
\usepackage{microtype}      
\usepackage{xcolor}         
\usepackage{wrapfig}
\usepackage{multirow}
\usepackage{graphicx}
\usepackage{multicol}
\usepackage{multirow}
\usepackage{subfigure}
\title{Understanding the Robustness of Multi-Exit Models \\ under Common Corruptions}

\author{
Akshay Mehra\textsuperscript{1}\textsuperscript{*}\textsuperscript{+}, Skyler Seto\textsuperscript{2}\textsuperscript{*}, Navdeep Jaitly\textsuperscript{2}, and Barry-John Theobald\textsuperscript{2}\\
{\small \textsuperscript{1}Tulane University \quad \textsuperscript{2}Apple}\\
{\tt\small amehra@tulane.edu, \{sseto, njaitly, barryjohn\_theobald\}@apple.com}
}

\begin{document}

\maketitle
\def\thefootnote{*}\footnotetext{Equal Contribution}\def\thefootnote{\arabic{footnote}}
\def\thefootnote{+}\footnotetext{Work done while intern at Apple}\def\thefootnote{\arabic{footnote}}

\begin{abstract}
Multi-Exit models (MEMs) use an early-exit strategy to improve the accuracy and efficiency of deep neural networks (DNNs) by allowing samples to exit the network before the last layer.
However, the effectiveness of MEMs in the presence of distribution shifts remains largely unexplored.  Our work examines how distribution shifts generated by common image corruptions affect the accuracy/efficiency of MEMs.
We find that under common corruptions, early-exiting at the first correct exit reduces the inference cost and provides a significant boost in accuracy ($\geq$ 10\%) over exiting at the last layer.
However, with realistic early-exit strategies, which do not assume knowledge about the correct exits, MEMs still reduce inference cost but provide a marginal improvement in accuracy ($\approx$ 1\%) compared to exiting at the last layer.
Moreover, the presence of distribution shift widens the gap between an MEM's maximum classification accuracy and realistic early-exit strategies by 5\% on average compared with the gap on in-distribution data. 
Our empirical analysis shows that the lack of calibration due to a distribution shift increases the susceptibility of such early-exit strategies to exit early and increases misclassification rates.
Furthermore, the lack of calibration increases the inconsistency in the predictions of the model across exits, leading to both inefficient inference and more misclassifications compared with evaluation on in-distribution data. 
Finally, we propose two metrics, \emph{underthinking} and \emph{overthinking}, that quantify the different  behavior of practical early-exit strategy under distribution shifts, and provide insights into improving the practical utility of MEMs.

\end{abstract}

\section{Introduction}

Deep Neural Networks (DNNs) have made major advances towards solving problems in image recognition \cite{krizhevsky2017imagenet}, object detection \cite{zhao2019object}, and image generation \cite{goodfellow2020generative} through learning complex feature representations. Although these networks are powerful, they often contain millions of parameters.  As a result, these modern architectures are costly to evaluate and deploy from a financial, an environmental, and a computational standpoint  \cite{ahmad2019can, strubell2019energy}.  Several works have investigated whether using the full depth of a network is necessary and have found, through adaptive inference mechanisms, that computation can be saved at inference time by making input-specific predictions \cite{hong2020panda, huang2017multi, kaya2019shallow}.  These works, and others, introduced the concept of Multi-Exit models (MEMs) \cite{iuzzolino2021improving,kaya2019shallow,laskaridis2021adaptive,scardapane2020should,sun2021early,teerapittayanon2016branchynet,wolczyk2021zero,yoon2022hubert,zhou2020bert}, which introduce side branches (usually consisting of a feature reduction and classification layer) and halt computation (early-exit) at inference time based on characteristics of the input, thereby saving compute while still making correct predictions.  

The main assumption of MEMs is that each sample does not require the full depth of the network, so when a correct prediction can be made in a layer before the last layer, then a sample should exit at that layer to save inference cost.  MEMs are popular due to their effectiveness at reducing the average inference cost of DNNs while offering potential accuracy improvements, and they can be applied to most DNN architectures. Recent works have demonstrated their effectiveness in various applications, including computer vision \cite{kaya2019shallow,wolczyk2021zero}, natural language processing, \cite{liu2020fastbert,xin2020deebert,zhou2020bert} and speech recognition \cite{yoon2022hubert}.  In related works, MEMs have also been shown to be effective at improving performance against backdoor attacks \cite{kaya2019shallow}, and on adversarial examples \cite{hong2020panda,hu2019triple}. 

Although MEMs can improve accuracy and reduce inference costs across a range of tasks, most prior works study efficiency gains only within the same distribution.  However, it is known that large, over-parameterized models are susceptible to poor performance under distribution shifts \cite{bulusu2020anomalous}.  Further, a solution for improving robustness to distribution shifts in the real-world is to increase the model size and complexity \cite{diffenderfer2021winning, hendrycks2021many} contrary to the goals of deploying more efficient models. 

In contrast to prior works, we investigate whether robustness is in-built in DNNs via MEMs and study the question: \emph{Can MEMs improve accuracy and reduce inference cost in the presence of unseen distribution shifts?}  

In this work, we focus on MEMs based on shallow deep networks (SDNs) \cite{kaya2019shallow} that modify the architecture of a standard deep neural network by attaching classification layers similar to the last layer to intermediate layers. Our first contribution, is to demonstrate that early-exiting can significantly improve the accuracy while reducing the inference cost of deep neural networks under common corruptions since samples from the shifted distributions are correctly classified at multiple intermediate exits.
Moreover, early-exiting using the first correct exit (see App.~\ref{app:overview_strategies} for an overview of early-exit strategies) boosts the accuracy by more than 10\% while still saving on average 40\% of the compute on corrupted data compared to exiting at the last layer, which suggests an MEM approach for efficient inference under distribution shift is possible. 

To understand the behavior of MEMs in a practical setup, we evaluate them using heuristic-based early-exit strategies using confidence \cite{kaya2019shallow}, patience \cite{zhou2020bert}, and a nearest neighbor (NN)-based strategy motivated from \cite{papernot2018deep}, which do not require knowledge of the true label.  See App.~\ref{app:overview_strategies} for additional information.
Our empirical analysis demonstrates that, while these early-exit strategies can reduce the average inference cost, they only provide a marginal improvement in accuracy on corrupted data (by $\approx$ 1\%) compared to exiting at the last layer.
Moreover, while a gap between early-exit strategies and an MEM's maximum classification performance exists on clean data, we find that the presence of distribution shift widens this gap by 5\% on average.

We demonstrate that the reason for this increased gap in accuracy is due to the lack of calibration of exits on corrupted data.
We propose two metrics, namely \emph{underthinking} and \emph{overthinking}, to capture the difference in the behavior of practical early-exit strategies.
Our empirical analysis, using a VGG-16 and a ResNet-56 model with CIFAR-10 and CIFAR-100, shows that the lack of calibration increases \emph{underthinking}, where a sample exits the network before the first exit where it would have been correctly classified, and also exacerbates \emph{overthinking}, where despite using more computation, the model misclassifies the sample.
Underthinking increases the misclassification rate where as overthinking leads to both increased misclassification and inefficient inference. 

Lastly, to improve the calibration of MEMs,  we study AugMix \cite{hendrycks2019augmix} based SDN training and show a significant decrease in under/overthinking and reduction in the performance gap between practical and oracle-based early-exit strategies.  We additionally experiment with the impact of adapting batch norm parameters at inference time as in \cite{benz2021revisiting} on MEMs and identify improvements in accuracy, but more subtle improvements in under/overthinking.
Thus, our work highlights the potential of early-exiting and shortcomings of current MEMs that limit their practical utility under distribution shifts and shows that training/early-exit strategies that take distribution shift into account should be considered with MEMs before deploying them in the wild.

\section{Related work}
\label{sec:background}
{\bf Multi-exit models:} Several dynamic approaches have been developed that aim to decrease the computational cost of DNNs while improving their performance via input adaptive inference \cite{bengio2013estimating,davis2013low,mcgill2017deciding}. 
MEMs \cite{kaya2019shallow,zhou2020bert,wolczyk2021zero,iuzzolino2021improving,yoon2022hubert, teerapittayanon2016branchynet,scardapane2020should,sun2021early, laskaridis2021adaptive} and adaptive neural networks (ANNs) \cite{liu2021anytime, veit2018convolutional, wang2018skipnet} are popular examples of this approach.  ANNs utilize a routing algorithm that skips parts of the network during inference time, while MEMs append multiple classification branches to the model and utilize an early-exiting strategy during inference to save computational costs.
The early-exit strategy is a crucial aspect of MEMs that decides which exit to use for a data sample.
Popular exit strategies used in MEMs are 
based on \emph{confidence} \cite{kaya2019shallow}, and \emph{patience} \cite{zhou2020bert}. 
Additionally, we propose an early-exit strategy that uses consistency in the predictions of the nearest neighbors \cite{papernot2018deep} of a test sample to decide the exit (see App.~\ref{app:overview_strategies} for details).

{\bf Benchmarking robustness to common corruptions:} Machine learning (ML) has been shown to be successful in scenarios when training and test distributions are the same. 
However, this basic assumption of training and test data being samples from the same underlying distribution might not hold in practice, which makes it essential to study the performance of various ML algorithms/models in the presence of distribution shifts \cite{szegedy2013intriguing,sinha2017certifying,mehra2021robust,mehra2021understanding, sun2021certified, cohen2019certified, hendrycks2018benchmarking,mehra2022domain}. 
Recently, \cite{hendrycks2019robustness} proposed corrupted versions of the popular benchmark datasets, CIFAR-10/100-C comprising of fifteen corruptions with varying levels of severity to evaluate the robustness of models trained on CIFAR-10/100.
We use these to study the behavior of MEMs and evaluate their readiness for deployment in the wild.  While robustness and calibration have been studied for standard DNNs \cite{ovadia2019can}, and recently for pruned models \cite{diffenderfer2021winning}, little work has explored the effect of such corruptions on adaptive inference models. 

{\bf Leveraging intermediate features for improved robustness:} While few works have studied directly using  intermediate representations to improve network robustness under distribution shifts, several works have studied the importance of such feature representations for such purposes.  Early works investigating the importance of intermediate representations applied linear probes to the intermediate layers and studied the ability of such layers, or even neurons within the layers to perform tasks different from the  task the model was trained for \cite{kim2018interpretability, zhou2018interpreting}.  Recent works have also studied their  use for transfer learning \cite{evci2022head2toe}. Howver, these works differ from the work here in that they do not aim to improve performance and efficiency on the original task the model is trained for. 

Other works combine information from intermediate representations to detect OOD and adversarial samples \cite{dong2022neural, lee2018simple, lin2021mood, papernot2018deep} .  However, except for \cite{lin2021mood} which performs OOD detection and stops computation at early exits, all approaches use the full network to make predictions rather than exiting early or saving compute.  Relating to OOD detection, prior works examine the impact of common corruptions on batch normalization statistics \cite{benz2021revisiting, liu2021ttt, schneider2020improving, wang2020tent}.  They find the mean and variance of batches of data from the corrupted set differs from clean samples, and propose adapting the batch normalizatoin statistics at test time.  These works differ from this study as prior works examine detection of distribution shifts with semantic and covariate shifts and do not make predictions.
\section{Understanding the behavior of MEMs under common corruptions}
Here we evaluate MEM architectures with six exits based on SDNs \cite{kaya2019shallow} using VGG-16/ResNet56 architectures with three early-exit strategies on CIFAR-10/100 and their corrupted versions.
We follow the SDN training procedure as proposed by \cite{kaya2019shallow}, which jointly optimizes the backbone and early-exit classifiers. 
See App.~\ref{app:additional_experiments} for full experiments using both datasets/models and AugMix-based \cite{hendrycks2019augmix} SDN training. See App.~\ref{app:experimental_details} for details on architectures/hyperparameters used in our work. 

{\bf Potential of MEMs for improving accuracy and efficiency.}
\label{sec:oracle}

\begin{wrapfigure}[14]{r}{0.45\textwidth}
\vspace{-0.3cm}
\centering
\includegraphics[width=0.25\textwidth]{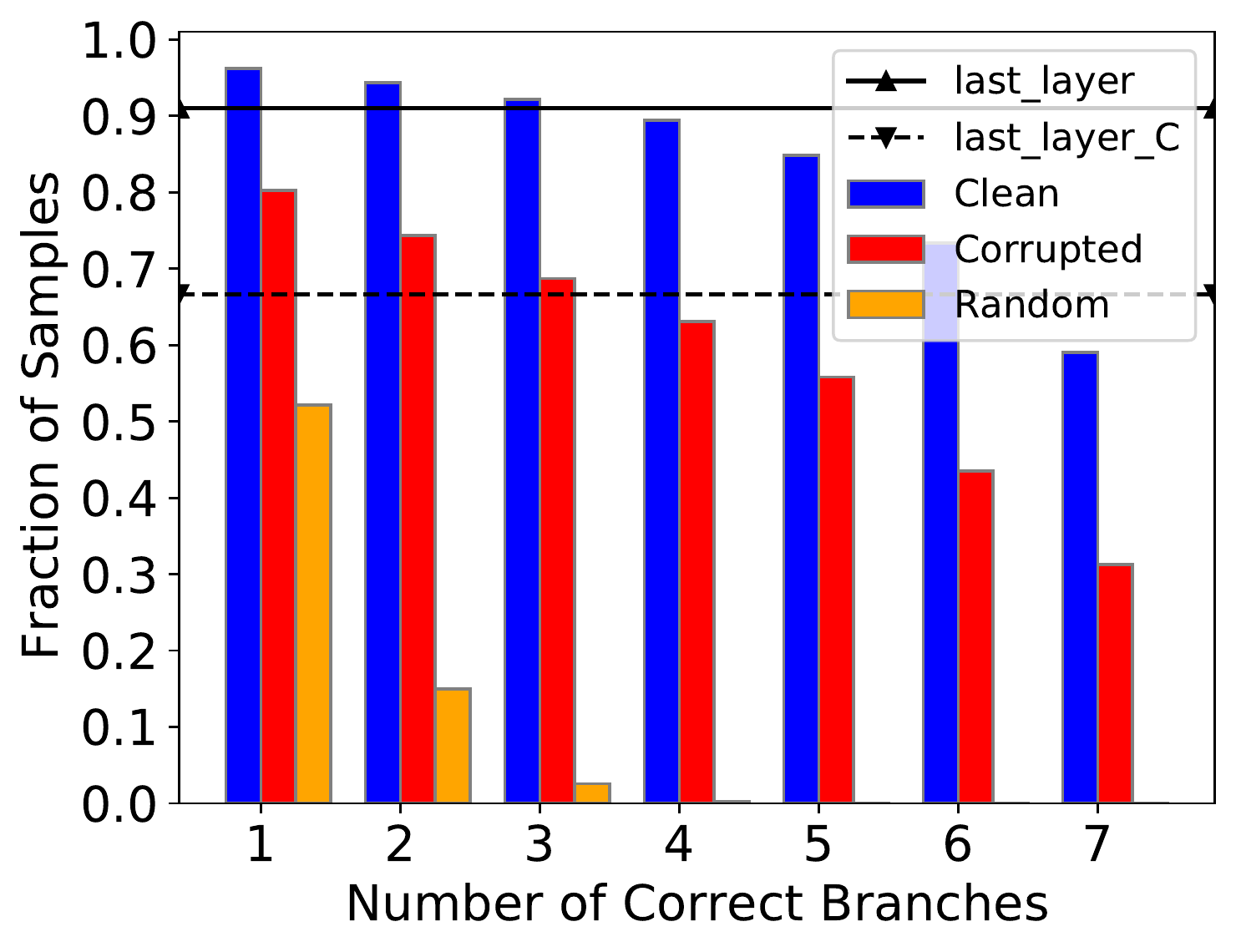}
\caption{Histogram of the number of correct branches (x-axis) for samples predicted by a ResNet-56 MEM on CIFAR-10. \_C denotes corrupted data.}
\vspace{-3mm}
\label{fig:correct_pred_resnet}
\end{wrapfigure}

We compute the number of exits in which a MEM correctly classifies a sample to demonstrate the potential of MEMs for improving accuracy and efficiency. 
The results in Figure~\ref{fig:correct_pred_resnet} show that a large fraction of samples are correctly classified by multiple exits.

\begin{figure*}[tb]
  \centering{
  \subfigure[Accuracy]{\includegraphics[width=0.32\columnwidth]{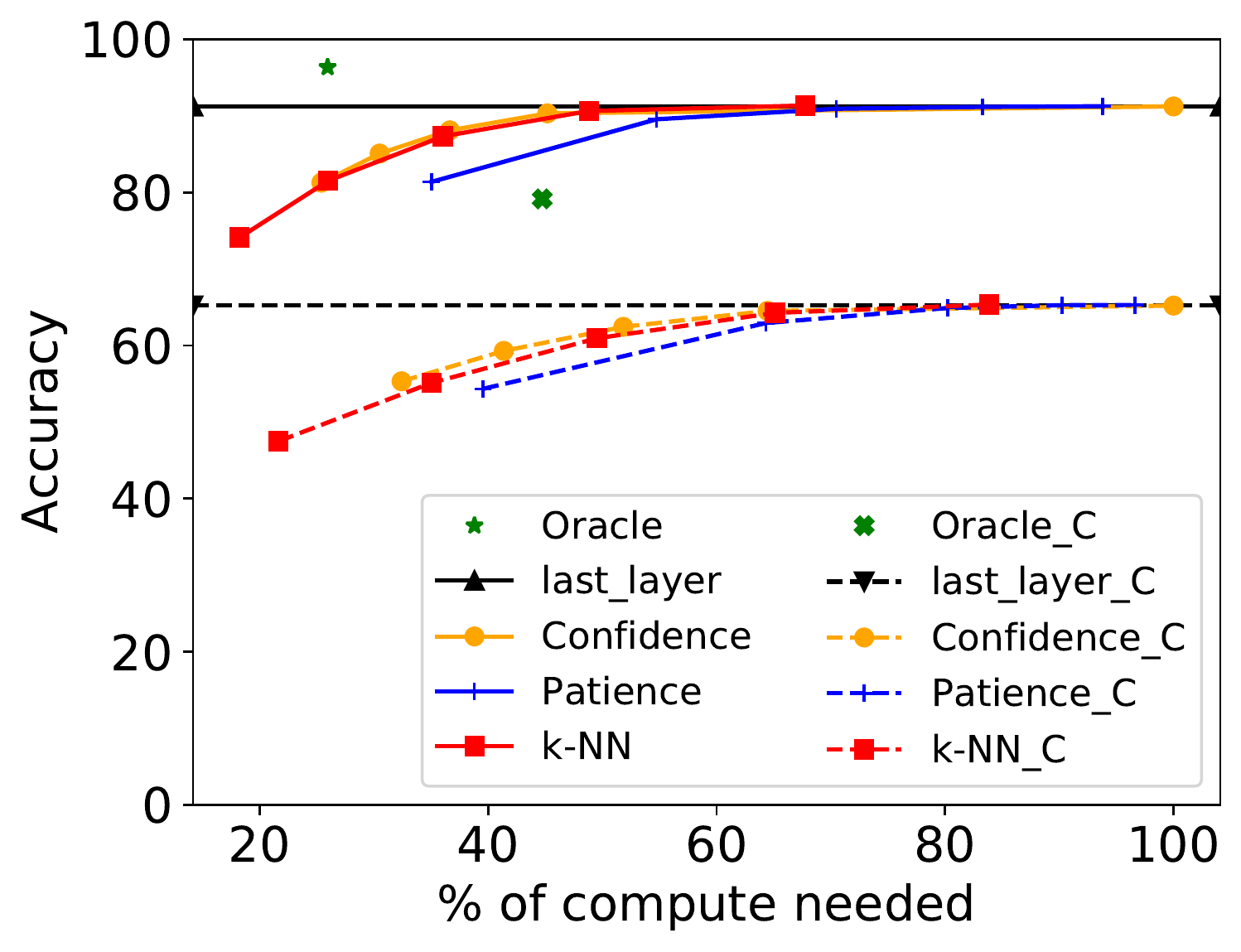}}
  \subfigure[Underthinking]{\includegraphics[width=0.32\columnwidth]{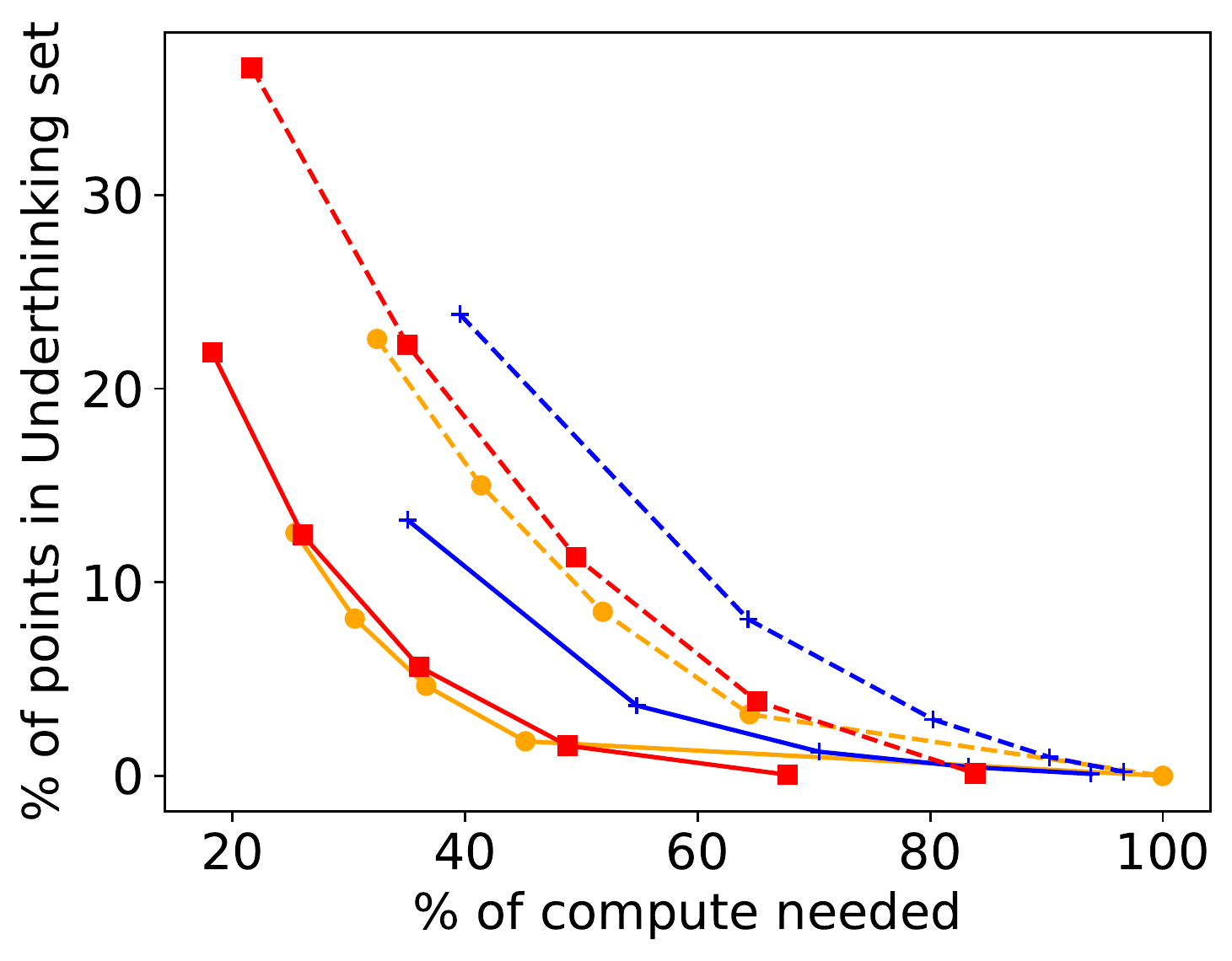}}
  \subfigure[Overthinking]{\includegraphics[width=0.32\columnwidth]{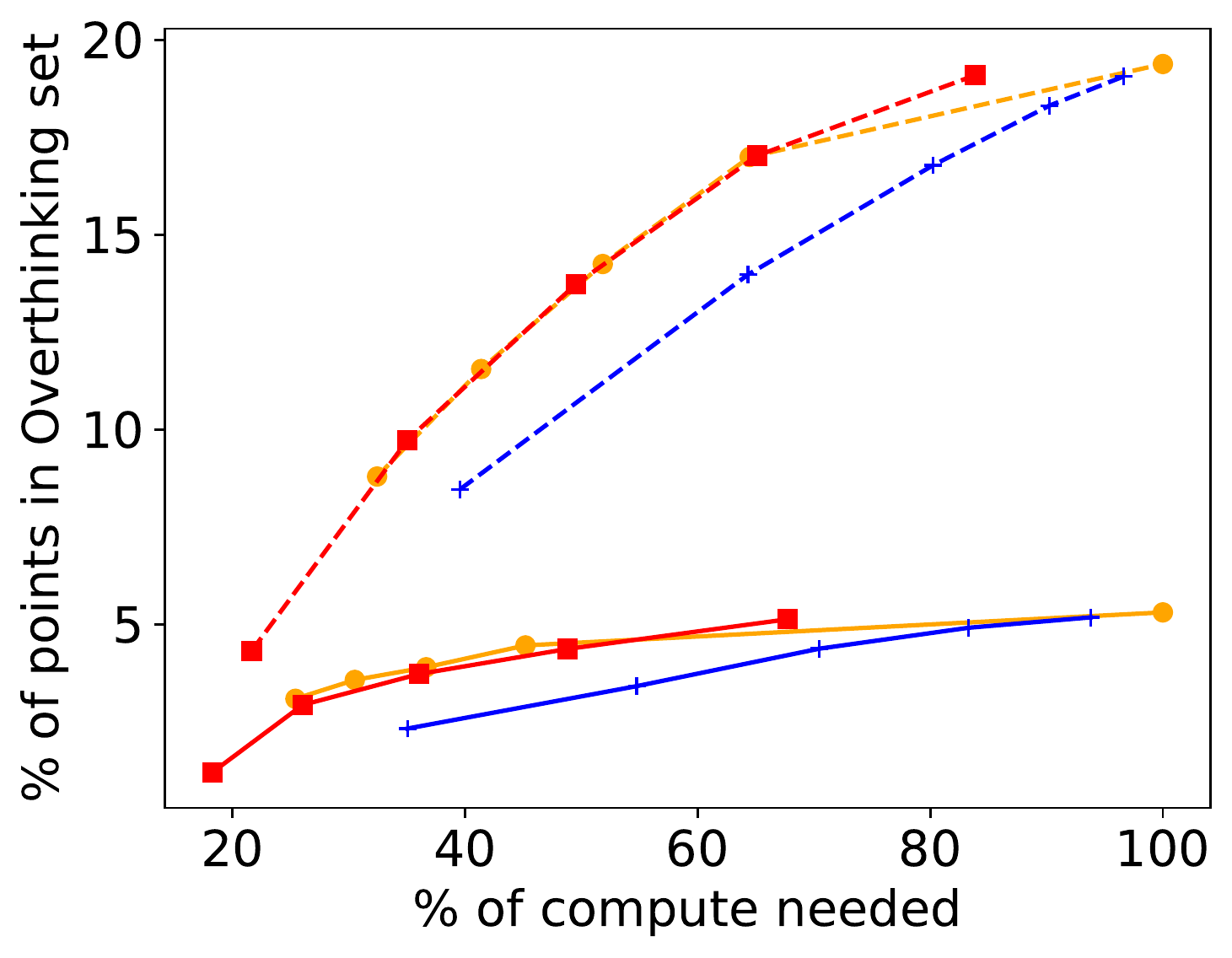}}
  }
  \caption{Accuracy, underthinking, and overthinking vs. the amount of compute used for MEMs (using ResNet-56 backbone model on CIFAR-10) with different early-exit strategies on clean and corrupted (denoted by \_C) datasets. The percentage of samples in the underthinking and overthinking sets are measured relative to the samples' first correct exit (set $\mathcal{O}$).
  }
  \label{fig:comparison_dist_shift}
\end{figure*}

In particular, since 5\% more of the samples are correctly classified twice in the network over only at the last exit, MEMs have a higher chance at improving accuracy on corrupted distributions. Additionally, performance of the trained networks are better than the probability of a correct classification from a random\footnote{Random network probabilities are simulated according to a binomial distribution.} network with equal number of exits.  Moreover, a sample exiting at its first correct exit provides a reduction in the inference cost of 40\% on average while boosting the accuracy by about 10\% as shown in Table~\ref{Table:SDN_vs_DNN_orcale} in the Appendix. While performance gains in Figure~\ref{fig:correct_pred_resnet} require knowledge of the label of the test sample, significant accuracy boosts suggest  implicit knowledge in the intermediate exits that can improve the performance of DNNs against common corruptions without any specialized data augmentation during training \cite{hendrycks2019augmix}. Similar findings have been observed in prior works in a different context \cite{evci2022head2toe,kim2018interpretability,papernot2018deep}.
{\bf Evaluation of MEMs with practical early-exit strategies.}
\label{sec:underthinking}
We now evaluate the performance of MEMs using practical early-exit strategies based on confidence \cite{kaya2019shallow}, patience, \cite{zhou2020bert} and nearest neighbor (NN) (see App.~\ref{app:overview_strategies}).
We use various exit thresholds for these strategies to obtain an accuracy versus efficiency curve as shown in Figure~\ref{fig:comparison_dist_shift}(a).
For the confidence-based strategy, we use thresholds $ \in \{0.6, 0.7, 0.8, 0.9, 1.0\} $, for patience we use $t \in \{1,2,3,4,5\}$ and for NN-based we use confidence values $\tau \in \{0.2, 0.4, 0.6, 0.8, 0.99\}$ and $k=$ 50 and 200 neighbors for CIFAR-10 and CIFAR-100, respectively.
We find that while MEMs can reduce the amount of compute needed on corrupted data, they require more compute than for clean data to achieve the same level of performance as the last layer, and only a small improvement in the accuracy is achieved ($\approx$ 1\%) compared to exiting at the last layer.
As shown in Figure~\ref{fig:comparison_dist_shift}(a), the gap between the accuracy of the oracle-based and the practical early-exit strategies widens on corrupted data by 5\% on average compared to the gap on in-distribution data.  Using a larger threshold (i.e.\ more compute) does not reduce this gap. 
Similar behavior is observed on other models/datasets (see App.\ref{app:additional_experiments}), which limits the utility of MEMs for improving accuracy on corrupted data.

{\bf Understanding the behavior of practical exit strategies.} 
We propose two metrics, underthinking and overthinking, 
to quantify the behavior of a practical early-exit strategy. 
To define these metrics, let $\mathcal{A}$ be the set of all samples in a dataset, set $\mathcal{O}$ be the subset of samples in $\mathcal{A}$ that are correctly classified by at least one exit, let $\ell_\tau(x)$ be the exit selected by an exit strategy $\ell_\tau$, with a threshold $\tau$ for a sample $(x,y)$, let $\ell^*(x,y)$ be the first correct exit for the sample and let $f^i(x)$ be the prediction at exit $i$.

Then the \emph{underthinking} set $UT(\ell_\tau) = \left\{x: x \in \mathcal{O} \; s.t. \; \ell_\tau(x) < \ell^*(x,y)\right\}$ i.e., the subset of samples in $\mathcal{O}$ for which the practical exit strategy exited \emph{before} the first correct exit for the sample.
Thus, all samples in $UT$ are misclassified by the practical exit strategy.

The \emph{overthinking} set is defined as $OT(\ell_\tau) = \left\{x: x \in \mathcal{O} \; s.t. \; \ell_\tau(x) > \ell^*(x,y)\; \mathrm{and} \;y\neq f^{\ell_\tau(x)}(x)\right\}$, i.e.\ the subset of samples in $\mathcal{O}$ for which the practical exit strategy exited \emph{after} the first correct exit but the sample was misclassified at that exit.
This definition of $OT$ is related to the notion of destructive overthinking proposed by \cite{kaya2019shallow} but it is more general since \cite{kaya2019shallow} only defined destructive overthinking considering samples which were misclassified at the last layer but our definition considers misclassification at an exit suggested by any practical exit strategy.

We find that distribution shift significantly increases underthinking and overthinking with all practical early-exit strategies as shown in Figure~\ref{fig:comparison_dist_shift}(b,c). We observe similar behavior across datasets and model architectures as shown in App.~\ref{app:additional_experiments}.
Increased underthinking shows that practical exit strategies stop the computation prematurely, leading to increased misclassification.
While increased overthinking highlights the failure of early-exit strategies to exit at an appropriate exit and results in an incorrect classification and wasted compute.
Moreover, using more compute decreases underthinking but increases overthinking leaving the accuracy gap between oracle-based and practical exit strategies the same as shown in Figure~\ref{fig:comparison_dist_shift}(a). 
Thus, these two metrics help understand the reason for the difference in the behavior of the practical and oracle-based early-exit strategies.

\begin{figure*}[tb]
\centering
\includegraphics[width=0.32\textwidth]{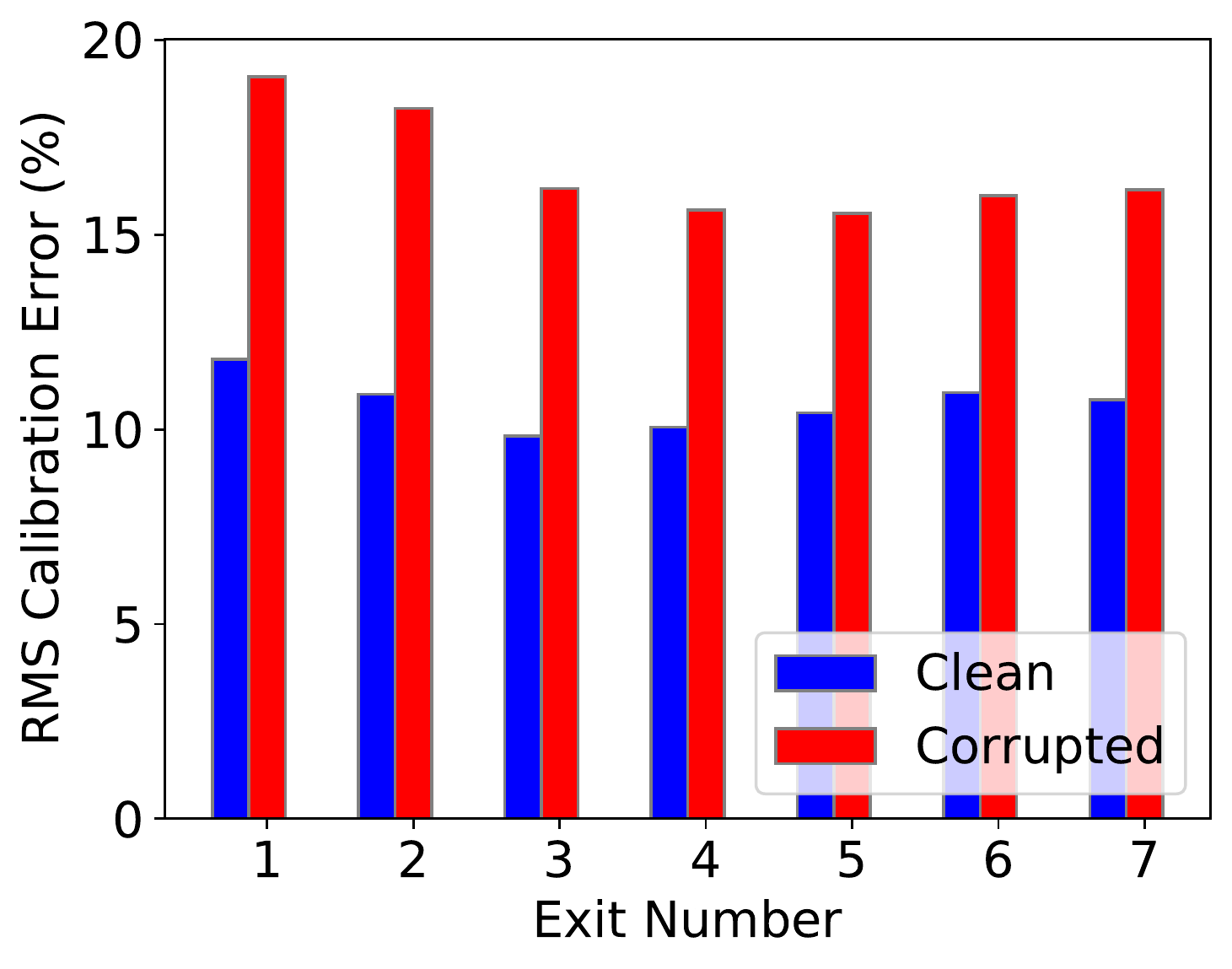}
\includegraphics[width=0.32\textwidth]{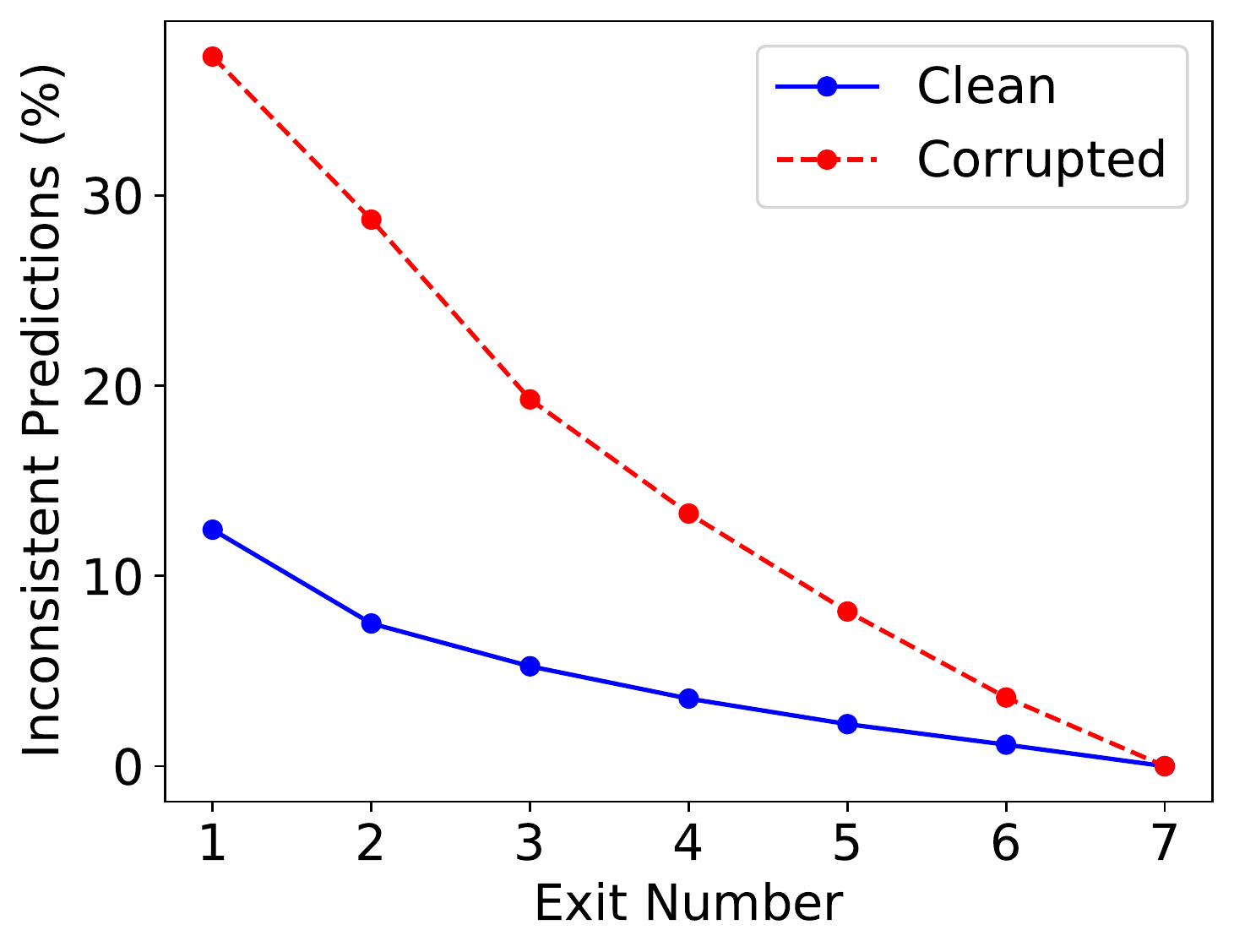}
\caption{Higher calibration error (left) and inconsistency in predictions (right) for exits of the MEMs (ResNet-56 on CIFAR-10) under corruptions. 
}
\label{fig:calibration_inconsistent_preds}
\end{figure*}

{\bf Understanding why under/overthinking increases on corrupted data with practical early-exit strategies.}
We find poor calibration of MEMs in the presence of distribution shift to be the primary cause for increased underthinking.
As shown in Figure~\ref{fig:calibration_inconsistent_preds} (left), we find that root mean squared (RMS) calibration error \cite{hendrycks2019augmix}
of all exits is higher on corrupted data compared to clean data.

This lack of calibration suggests that the confidence of the model is not representative of the true uncertainty estimates, especially on earlier exits for corrupted data.
Thus, exiting based on confidence in early layers can make a sample exit the network before the first exit that would otherwise correctly classify the sample, leading to underthinking.

To understand underthinking with NN-based strategy we show the t-SNE embeddings of clean and corrupted data from the layer before the softmax in each exit of the MEM in Fig~\ref{fig:knn_tsne} in Appendix \ref{app:additional_experiments}.
We find that the nearest neighbors of the corrupted samples often tend to be the training samples from an incorrect class. 
This leads to underthinking since a corrupted sample lying close to the samples from the same wrong class can lead to high agreement between the labels of the neighbors and that of the corrupted sample. This high agreement makes the NN-based strategy exit the network earlier than the first correct exit, leading to underthinking.

The lack of calibration in presence of distribution shifts further leads to increasing the inconsistency in the predictions of the model across exits.
To compute prediction inconsistency of a model we measure the proportion of samples correctly classified at an exit that are misclassified by a later exit. 
A higher number suggests that 
even if an exit strategy selects an exit later than the exit suggested by the oracle-based strategy, the sample may be misclassified, increasing overthinking.
Figure~\ref{fig:calibration_inconsistent_preds}(right) shows MEMs have a higher inconsistency in their predictions on corrupted data. 

\begin{figure*}[tb]
  \centering{
  \subfigure[Accuracy]{\includegraphics[width=0.24\columnwidth]{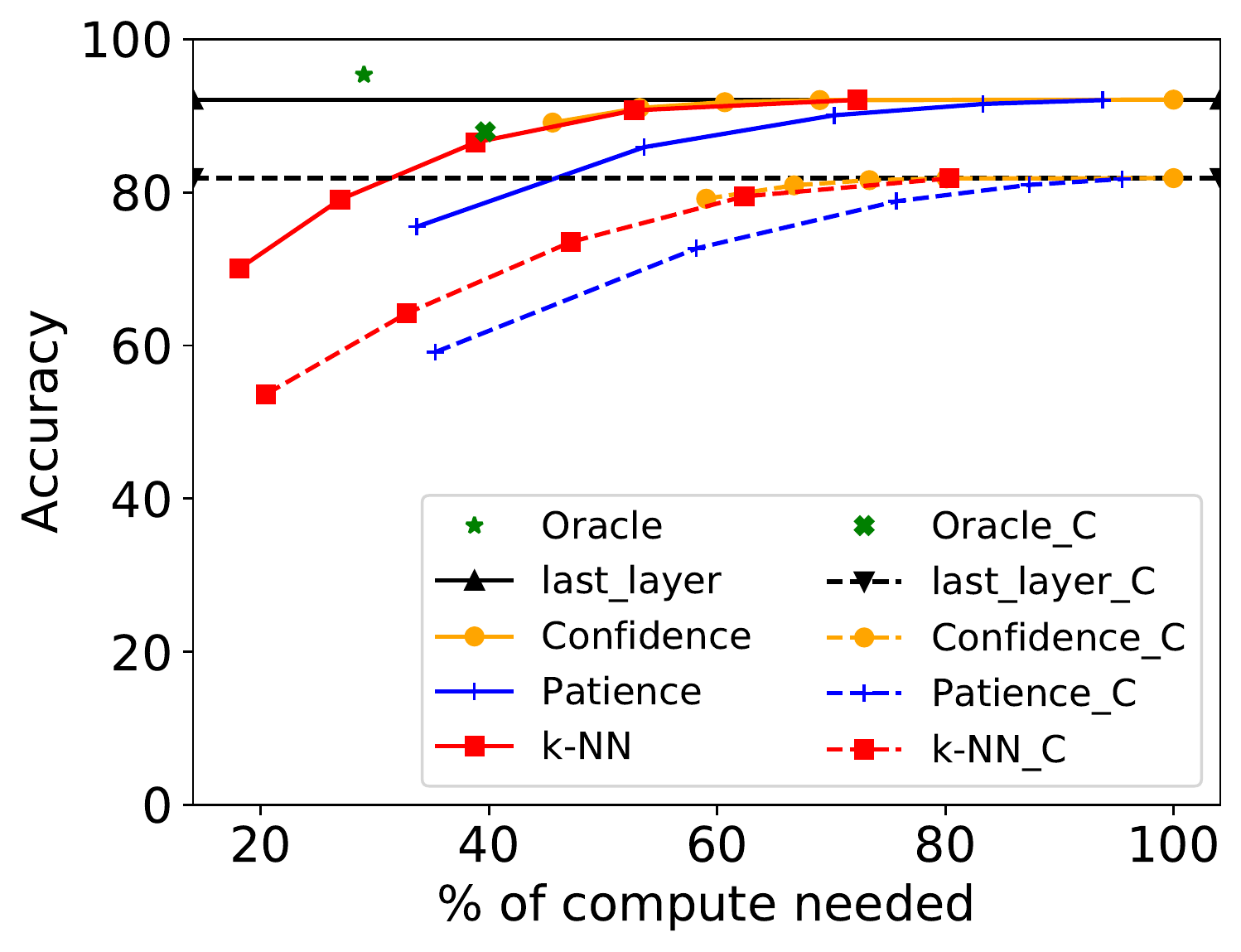}}
  \subfigure[Calibration]{\includegraphics[width=0.24\columnwidth]{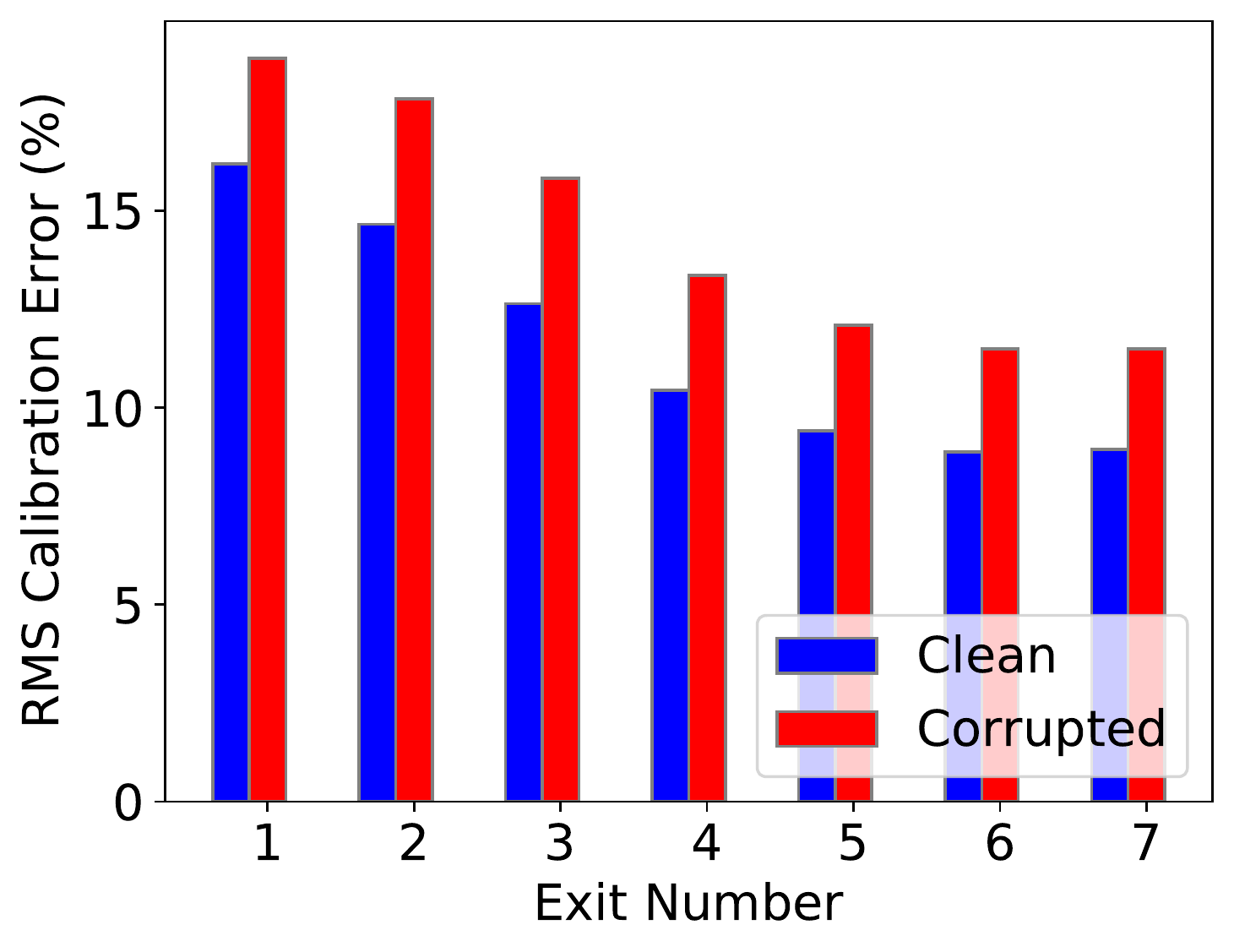}}
  \subfigure[Underthinking]{\includegraphics[width=0.24\columnwidth]{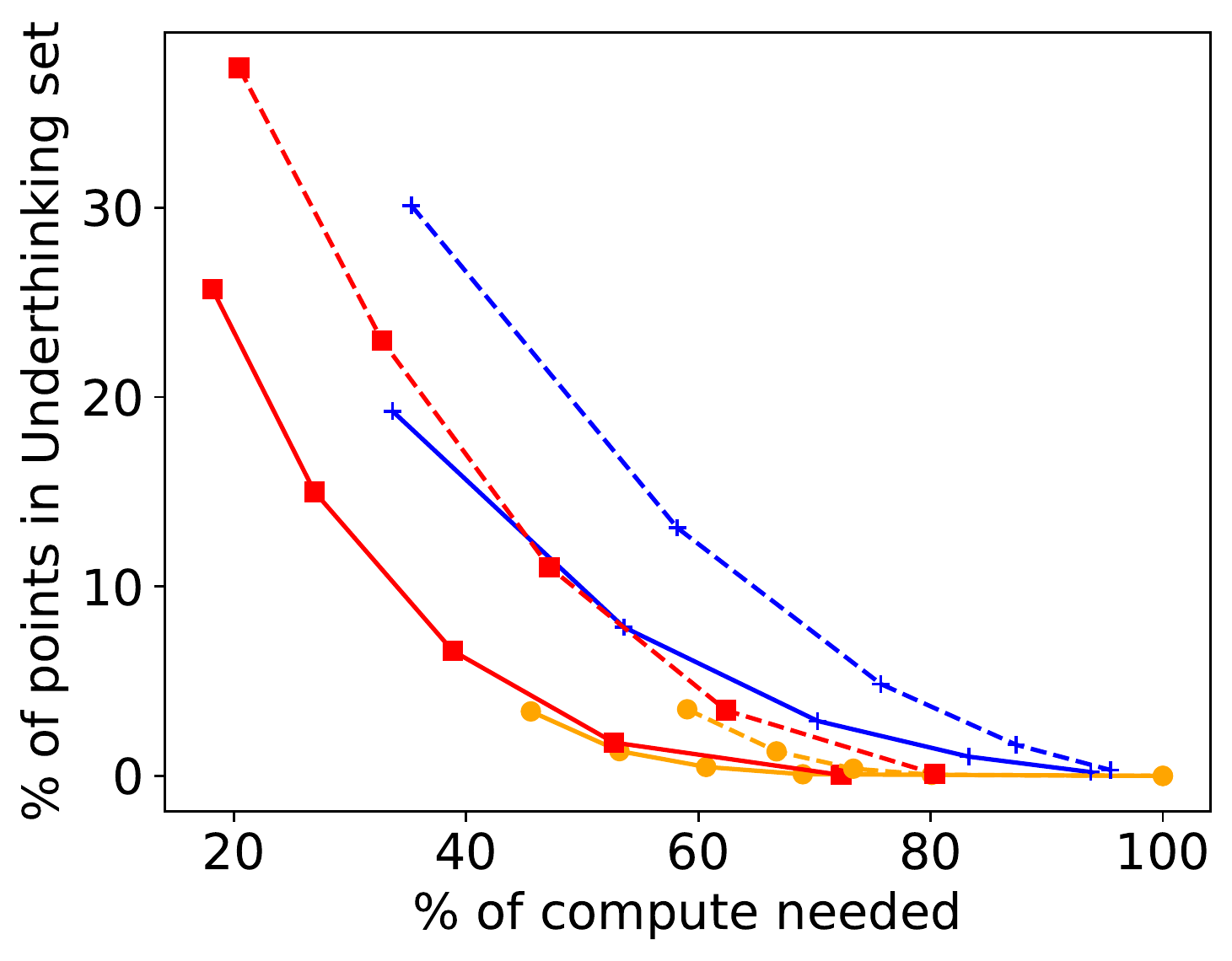}}
  \subfigure[Overthinking]{\includegraphics[width=0.24\columnwidth]{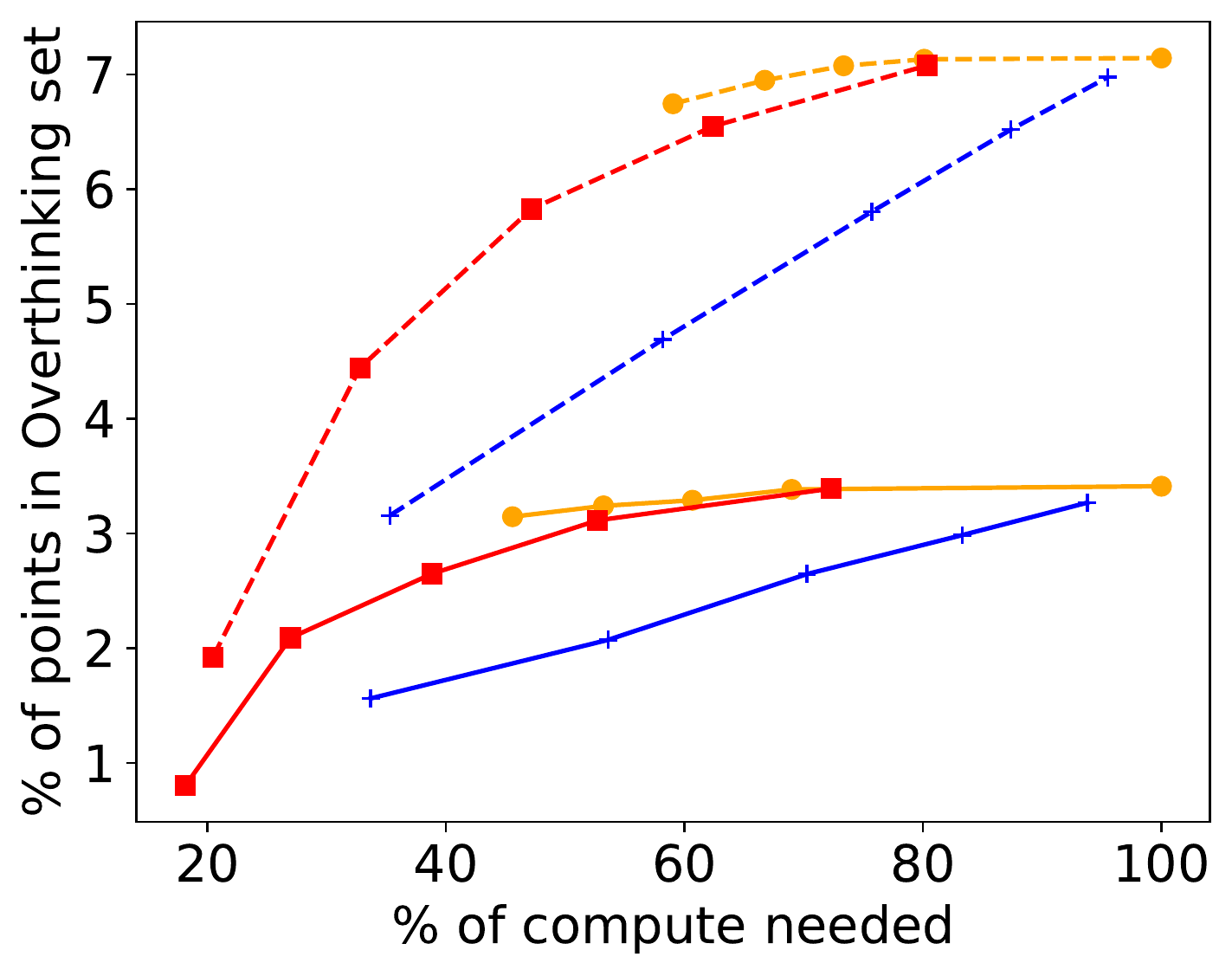}}
  }
  \caption{Accuracy, underthinking, and overthinking vs. the amount of compute used for MEMs trained with AugMix (using ResNet-56 backbone model on CIFAR-10) with practical exit strategies on clean and corrupted (denoted by \_C) datasets.  The percentage of samples in the underthinking and overthinking sets are measured relative to the samples' first correct exit (set $\mathcal{O}$). 
  }
  \label{fig:augmix}
\end{figure*}

\begin{figure*}[tb]
  \centering{
  \subfigure[Accuracy]{\includegraphics[width=0.24\columnwidth]{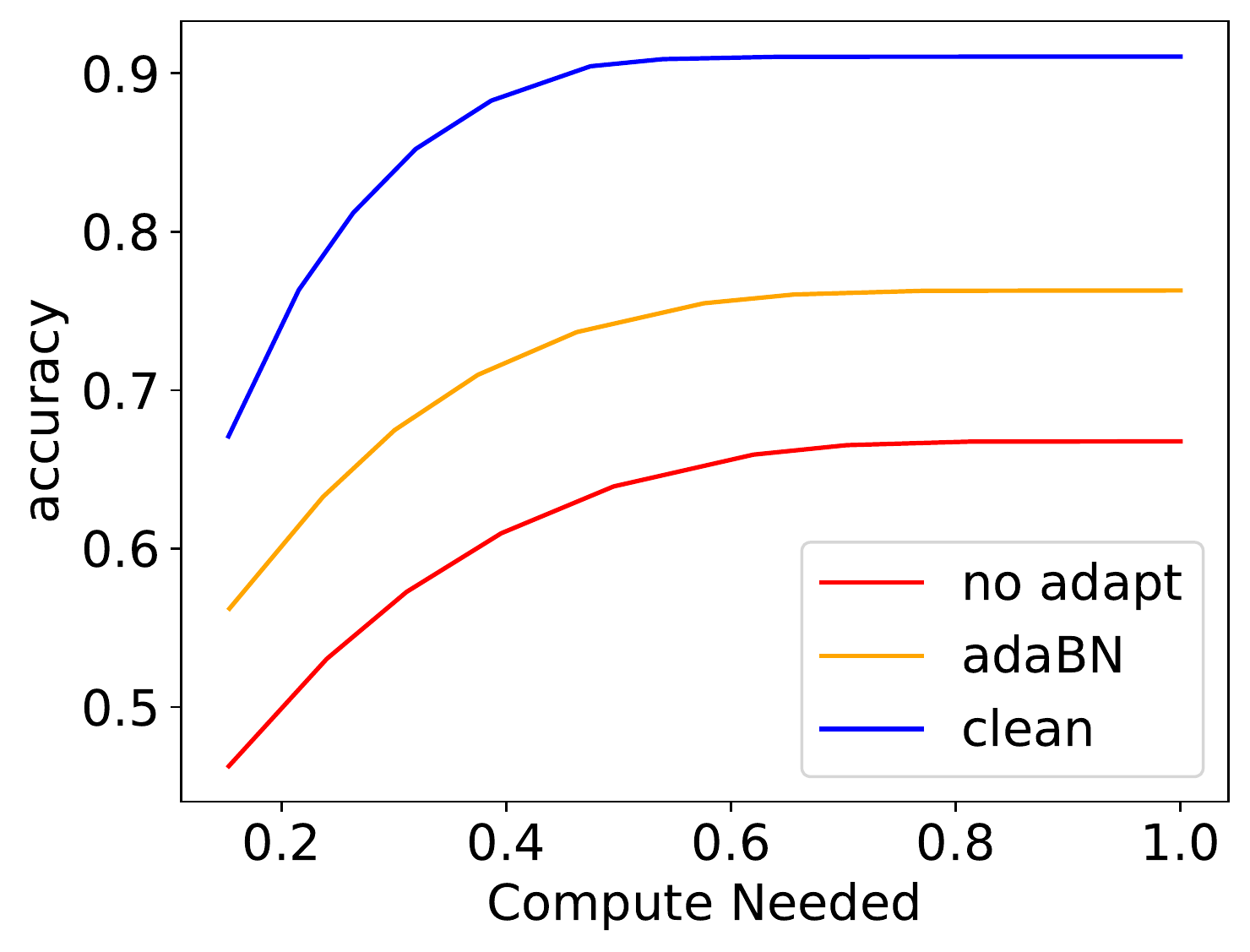}}
  \subfigure[Calibration]{\includegraphics[width=0.24\columnwidth]{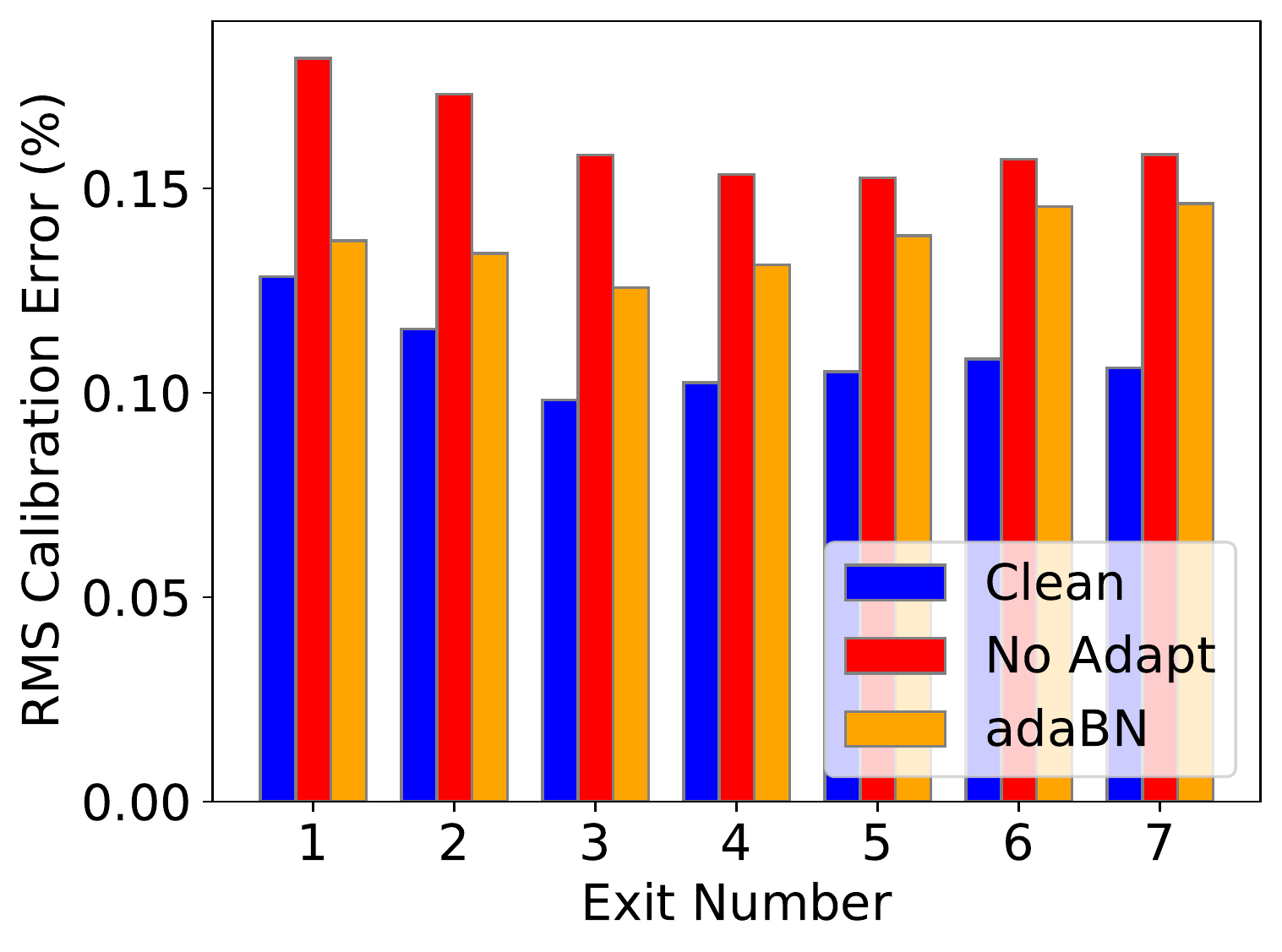}}
  \subfigure[Underthinking]{\includegraphics[width=0.24\columnwidth]{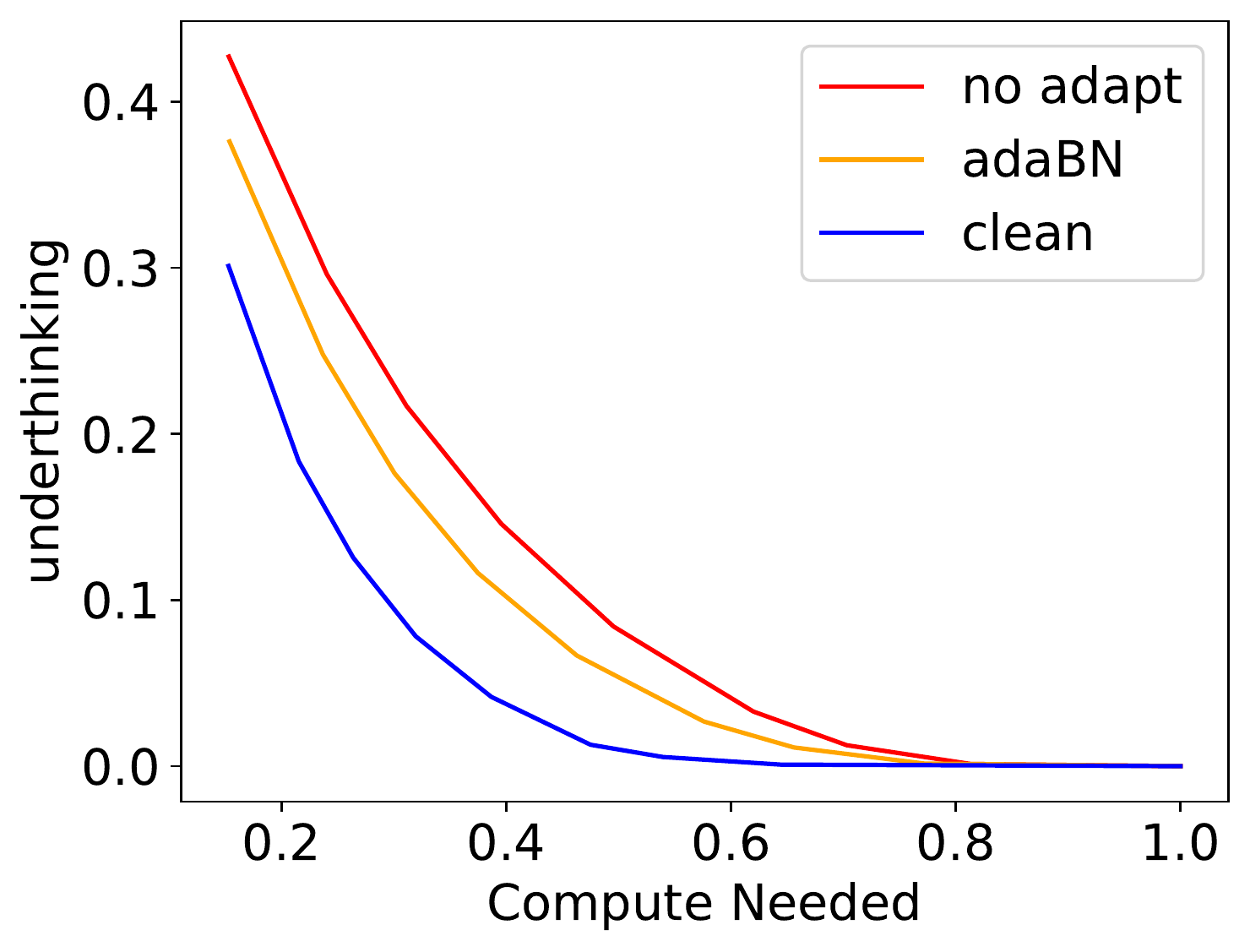}}
  \subfigure[Overthinking]{\includegraphics[width=0.24\columnwidth]{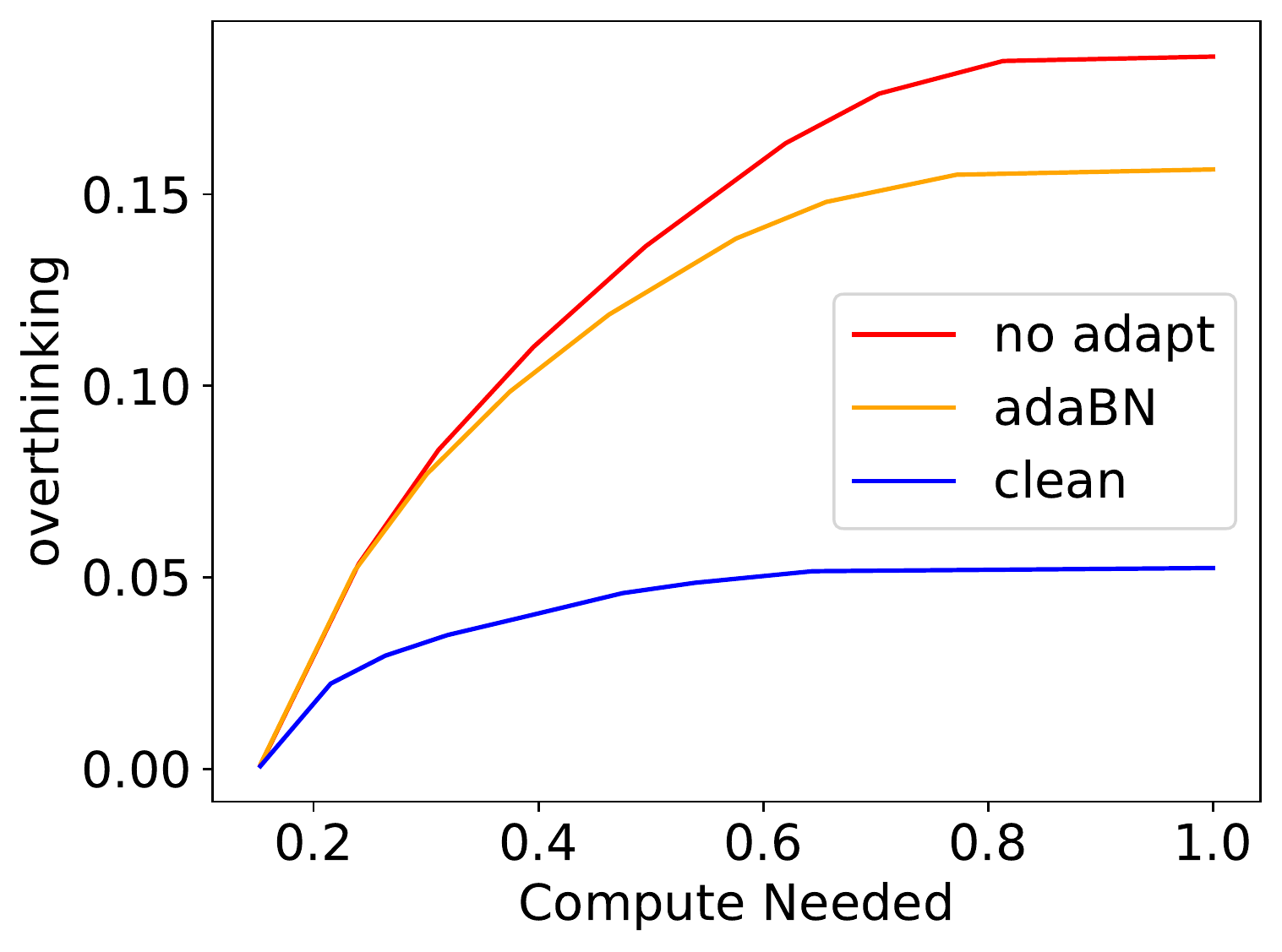}}
  }
  \caption{Accuracy, underthinking, and overthinking vs. the amount of compute used for MEMs with adapted batchnorm parameters (using ResNet-56 backbone model on CIFAR-10) with confidence early-exit strategies on clean and corrupted (denoted by \_C) datasets. Models evaluated on clean data do not use adapted batchnorm parameters.  The percentage of samples in the underthinking and overthinking sets are measured relative to the samples' first correct exit (set $\mathcal{O}$). 
  }
  \label{fig:adaBN}
\end{figure*}

Lastly, to improve calibration error and decrease the inconsistency in the predictions of MEMs on corrupted data, we use AugMix \cite{hendrycks2019augmix} to train the SDN model. To train SDNs with AugMix, we add the JSD regularzation loss for each exit of the network as the regularizer.  Results for the ResNet MEM on CIFAR-10 are shown in Figure~\ref{fig:augmix}, and indicate that AugMix improves calibration of the models, leading to a decrease in under/overthinking, improvement in the accuracy of practical and oracle-based strategies and reduces their performance gap on corrupted data.  Further results for other models and datasets are in App.~\ref{app:additional_experiments}.  We also experiment with adapting batch normalization statistics \cite{benz2021revisiting} at inference time for improving corruption robustness and reducing inference costs of MEMs.  Our results in Figure~\ref{fig:adaBN} show significant increases in accuracy, but marginal changes to under/over-thinking, calibration (especially in later exits).  Further results for other models and datasets are in App.~\ref{app:additional_experiments}.

\vspace{-0.2cm}
\section{Conclusion}
\vspace{-0.1cm}
We studied the behavior of MEMs in presence of distribution shifts. 
We showed that multiple exits in a MEM can correctly classify samples from corrupted data distributions 
demonstrating the possibility of 
early-exiting for improving accuracy and efficiency of DNNs
under distribution shift. 
We also proposed two metrics to quantify the reason practical early-exit strategies suffer at improving accuracy and efficiency under distribution shifts and highlighted the lack of calibration and the inconsistency in the predictions of the exits  to be the primary causes for under/overthinking in MEMs. 

\clearpage
\bibliographystyle{plain}
{\small \bibliography{neurips_distshift_2022}}

\begin{thebibliography}{10}

\bibitem{ahmad2019can}
Subutai Ahmad and Luiz Scheinkman.
\newblock How can we be so dense? the benefits of using highly sparse
  representations.
\newblock {\em arXiv preprint arXiv:1903.11257}, 2019.

\bibitem{bengio2013estimating}
Yoshua Bengio, Nicholas L{\'e}onard, and Aaron Courville.
\newblock Estimating or propagating gradients through stochastic neurons for
  conditional computation.
\newblock {\em arXiv preprint arXiv:1308.3432}, 2013.

\bibitem{benz2021revisiting}
Philipp Benz, Chaoning Zhang, Adil Karjauv, and In~So Kweon.
\newblock Revisiting batch normalization for improving corruption robustness.
\newblock In {\em Proceedings of the IEEE/CVF Winter Conference on Applications
  of Computer Vision}, pages 494--503, 2021.

\bibitem{bulusu2020anomalous}
Saikiran Bulusu, Bhavya Kailkhura, Bo~Li, Pramod~K Varshney, and Dawn Song.
\newblock Anomalous example detection in deep learning: A survey.
\newblock {\em IEEE Access}, 8:132330--132347, 2020.

\bibitem{cohen2019certified}
Jeremy Cohen, Elan Rosenfeld, and Zico Kolter.
\newblock Certified adversarial robustness via randomized smoothing.
\newblock In {\em International Conference on Machine Learning}, pages
  1310--1320. PMLR, 2019.

\bibitem{davis2013low}
Andrew Davis and Itamar Arel.
\newblock Low-rank approximations for conditional feedforward computation in
  deep neural networks.
\newblock {\em arXiv preprint arXiv:1312.4461}, 2013.

\bibitem{diffenderfer2021winning}
James Diffenderfer, Brian Bartoldson, Shreya Chaganti, Jize Zhang, and Bhavya
  Kailkhura.
\newblock A winning hand: Compressing deep networks can improve
  out-of-distribution robustness.
\newblock {\em Advances in Neural Information Processing Systems}, 34:664--676,
  2021.

\bibitem{dong2022neural}
Xin Dong, Junfeng Guo, Ang Li, Wei-Te Ting, Cong Liu, and HT~Kung.
\newblock Neural mean discrepancy for efficient out-of-distribution detection.
\newblock In {\em Proceedings of the IEEE/CVF Conference on Computer Vision and
  Pattern Recognition}, pages 19217--19227, 2022.

\bibitem{evci2022head2toe}
Utku Evci, Vincent Dumoulin, Hugo Larochelle, and Michael~C Mozer.
\newblock Head2toe: Utilizing intermediate representations for better transfer
  learning.
\newblock In {\em International Conference on Machine Learning}, pages
  6009--6033. PMLR, 2022.

\bibitem{goodfellow2020generative}
Ian Goodfellow, Jean Pouget-Abadie, Mehdi Mirza, Bing Xu, David Warde-Farley,
  Sherjil Ozair, Aaron Courville, and Yoshua Bengio.
\newblock Generative adversarial networks.
\newblock {\em Communications of the ACM}, 63(11):139--144, 2020.

\bibitem{hendrycks2021many}
Dan Hendrycks, Steven Basart, Norman Mu, Saurav Kadavath, Frank Wang, Evan
  Dorundo, Rahul Desai, Tyler Zhu, Samyak Parajuli, Mike Guo, et~al.
\newblock The many faces of robustness: A critical analysis of
  out-of-distribution generalization.
\newblock In {\em Proceedings of the IEEE/CVF International Conference on
  Computer Vision}, pages 8340--8349, 2021.

\bibitem{hendrycks2019robustness}
Dan Hendrycks and Thomas Dietterich.
\newblock Benchmarking neural network robustness to common corruptions and
  perturbations.
\newblock {\em Proceedings of the International Conference on Learning
  Representations}, 2019.

\bibitem{hendrycks2018benchmarking}
Dan Hendrycks and Thomas~G Dietterich.
\newblock Benchmarking neural network robustness to common corruptions and
  surface variations.
\newblock {\em arXiv preprint arXiv:1807.01697}, 2018.

\bibitem{hendrycks2019augmix}
Dan Hendrycks, Norman Mu, Ekin~Dogus Cubuk, Barret Zoph, Justin Gilmer, and
  Balaji Lakshminarayanan.
\newblock Augmix: A simple data processing method to improve robustness and
  uncertainty.
\newblock In {\em International Conference on Learning Representations}, 2019.

\bibitem{hong2020panda}
Sanghyun Hong, Yi{\u{g}}itcan Kaya, Ionu{\c{t}}-Vlad Modoranu, and Tudor
  Dumitra{\c{s}}.
\newblock A panda? no, it's a sloth: Slowdown attacks on adaptive multi-exit
  neural network inference.
\newblock {\em arXiv preprint arXiv:2010.02432}, 2020.

\bibitem{hu2019triple}
Ting-Kuei Hu, Tianlong Chen, Haotao Wang, and Zhangyang Wang.
\newblock Triple wins: Boosting accuracy, robustness and efficiency together by
  enabling input-adaptive inference.
\newblock In {\em International Conference on Learning Representations}, 2019.

\bibitem{huang2017multi}
Gao Huang, Danlu Chen, Tianhong Li, Felix Wu, Laurens Van Der~Maaten, and
  Kilian~Q Weinberger.
\newblock Multi-scale dense networks for resource efficient image
  classification.
\newblock {\em arXiv preprint arXiv:1703.09844}, 2017.

\bibitem{iuzzolino2021improving}
Michael Iuzzolino, Michael~C Mozer, and Samy Bengio.
\newblock Improving anytime prediction with parallel cascaded networks and a
  temporal-difference loss.
\newblock {\em Advances in Neural Information Processing Systems},
  34:27631--27644, 2021.

\bibitem{johnson2019billion}
Jeff Johnson, Matthijs Douze, and Herv{\'e} J{\'e}gou.
\newblock Billion-scale similarity search with {GPUs}.
\newblock {\em IEEE Transactions on Big Data}, 7(3):535--547, 2019.

\bibitem{kaya2019shallow}
Yigitcan Kaya, Sanghyun Hong, and Tudor Dumitras.
\newblock Shallow-deep networks: Understanding and mitigating network
  overthinking.
\newblock In {\em International conference on machine learning}, pages
  3301--3310. PMLR, 2019.

\bibitem{kim2018interpretability}
Been Kim, Martin Wattenberg, Justin Gilmer, Carrie Cai, James Wexler, Fernanda
  Viegas, et~al.
\newblock Interpretability beyond feature attribution: Quantitative testing
  with concept activation vectors (tcav).
\newblock In {\em International conference on machine learning}, pages
  2668--2677. PMLR, 2018.

\bibitem{krizhevsky2017imagenet}
Alex Krizhevsky, Ilya Sutskever, and Geoffrey~E Hinton.
\newblock Imagenet classification with deep convolutional neural networks.
\newblock {\em Communications of the ACM}, 60(6):84--90, 2017.

\bibitem{laskaridis2021adaptive}
Stefanos Laskaridis, Alexandros Kouris, and Nicholas~D Lane.
\newblock Adaptive inference through early-exit networks: Design, challenges
  and directions.
\newblock In {\em Proceedings of the 5th International Workshop on Embedded and
  Mobile Deep Learning}, pages 1--6, 2021.

\bibitem{lee2018simple}
Kimin Lee, Kibok Lee, Honglak Lee, and Jinwoo Shin.
\newblock A simple unified framework for detecting out-of-distribution samples
  and adversarial attacks.
\newblock {\em Advances in neural information processing systems}, 31, 2018.

\bibitem{lin2021mood}
Ziqian Lin, Sreya~Dutta Roy, and Yixuan Li.
\newblock Mood: Multi-level out-of-distribution detection.
\newblock In {\em Proceedings of the IEEE/CVF Conference on Computer Vision and
  Pattern Recognition}, pages 15313--15323, 2021.

\bibitem{liu2020fastbert}
Weijie Liu, Peng Zhou, Zhe Zhao, Zhiruo Wang, Haotang Deng, and Qi~Ju.
\newblock Fastbert: a self-distilling bert with adaptive inference time.
\newblock {\em arXiv preprint arXiv:2004.02178}, 2020.

\bibitem{liu2021ttt}
Yuejiang Liu, Parth Kothari, Bastien van Delft, Baptiste Bellot-Gurlet, Taylor
  Mordan, and Alexandre Alahi.
\newblock Ttt++: When does self-supervised test-time training fail or thrive?
\newblock {\em Advances in Neural Information Processing Systems},
  34:21808--21820, 2021.

\bibitem{liu2021anytime}
Zhuang Liu, Zhiqiu Xu, Hung-Ju Wang, Trevor Darrell, and Evan Shelhamer.
\newblock Anytime dense prediction with confidence adaptivity.
\newblock In {\em International Conference on Learning Representations}, 2021.

\bibitem{mcgill2017deciding}
Mason McGill and Pietro Perona.
\newblock Deciding how to decide: Dynamic routing in artificial neural
  networks.
\newblock In {\em International Conference on Machine Learning}, pages
  2363--2372. PMLR, 2017.

\bibitem{mehra2022domain}
Akshay Mehra, Bhavya Kailkhura, Pin-Yu Chen, and Jihun Hamm.
\newblock Do domain generalization methods generalize well?

\bibitem{mehra2021robust}
Akshay Mehra, Bhavya Kailkhura, Pin-Yu Chen, and Jihun Hamm.
\newblock How robust are randomized smoothing based defenses to data poisoning?
\newblock In {\em Proceedings of the IEEE/CVF Conference on Computer Vision and
  Pattern Recognition}, pages 13244--13253, 2021.

\bibitem{mehra2021understanding}
Akshay Mehra, Bhavya Kailkhura, Pin-Yu Chen, and Jihun Hamm.
\newblock Understanding the limits of unsupervised domain adaptation via data
  poisoning.
\newblock {\em Advances in Neural Information Processing Systems},
  34:17347--17359, 2021.

\bibitem{ovadia2019can}
Yaniv Ovadia, Emily Fertig, Jie Ren, Zachary Nado, David Sculley, Sebastian
  Nowozin, Joshua Dillon, Balaji Lakshminarayanan, and Jasper Snoek.
\newblock Can you trust your model's uncertainty? evaluating predictive
  uncertainty under dataset shift.
\newblock {\em Advances in neural information processing systems}, 32, 2019.

\bibitem{papernot2018deep}
Nicolas Papernot and Patrick McDaniel.
\newblock Deep k-nearest neighbors: Towards confident, interpretable and robust
  deep learning.
\newblock {\em arXiv preprint arXiv:1803.04765}, 2018.

\bibitem{scardapane2020should}
Simone Scardapane, Michele Scarpiniti, Enzo Baccarelli, and Aurelio Uncini.
\newblock Why should we add early exits to neural networks?
\newblock {\em Cognitive Computation}, 12(5):954--966, 2020.

\bibitem{schneider2020improving}
Steffen Schneider, Evgenia Rusak, Luisa Eck, Oliver Bringmann, Wieland Brendel,
  and Matthias Bethge.
\newblock Improving robustness against common corruptions by covariate shift
  adaptation.
\newblock {\em Advances in Neural Information Processing Systems},
  33:11539--11551, 2020.

\bibitem{sinha2017certifying}
Aman Sinha, Hongseok Namkoong, Riccardo Volpi, and John Duchi.
\newblock Certifying some distributional robustness with principled adversarial
  training.
\newblock {\em arXiv preprint arXiv:1710.10571}, 2017.

\bibitem{strubell2019energy}
Emma Strubell, Ananya Ganesh, and Andrew McCallum.
\newblock Energy and policy considerations for deep learning in nlp.
\newblock In {\em Proceedings of the 57th Annual Meeting of the Association for
  Computational Linguistics}, pages 3645--3650, 2019.

\bibitem{sun2021certified}
Jiachen Sun, Akshay Mehra, Bhavya Kailkhura, Pin-Yu Chen, Dan Hendrycks, Jihun
  Hamm, and Z~Morley Mao.
\newblock Certified adversarial defenses meet out-of-distribution corruptions:
  Benchmarking robustness and simple baselines.
\newblock {\em arXiv preprint arXiv:2112.00659}, 2021.

\bibitem{sun2021early}
Tianxiang Sun, Yunhua Zhou, Xiangyang Liu, Xinyu Zhang, Hao Jiang, Zhao Cao,
  Xuanjing Huang, and Xipeng Qiu.
\newblock Early exiting with ensemble internal classifiers.
\newblock {\em arXiv preprint arXiv:2105.13792}, 2021.

\bibitem{szegedy2013intriguing}
Christian Szegedy, Wojciech Zaremba, Ilya Sutskever, Joan Bruna, Dumitru Erhan,
  Ian Goodfellow, and Rob Fergus.
\newblock Intriguing properties of neural networks.
\newblock {\em arXiv preprint arXiv:1312.6199}, 2013.

\bibitem{teerapittayanon2016branchynet}
Surat Teerapittayanon, Bradley McDanel, and Hsiang-Tsung Kung.
\newblock Branchynet: Fast inference via early exiting from deep neural
  networks.
\newblock In {\em 2016 23rd International Conference on Pattern Recognition
  (ICPR)}, pages 2464--2469. IEEE, 2016.

\bibitem{veit2018convolutional}
Andreas Veit and Serge Belongie.
\newblock Convolutional networks with adaptive inference graphs.
\newblock In {\em Proceedings of the European Conference on Computer Vision
  (ECCV)}, pages 3--18, 2018.

\bibitem{wang2020tent}
Dequan Wang, Evan Shelhamer, Shaoteng Liu, Bruno Olshausen, and Trevor Darrell.
\newblock Tent: Fully test-time adaptation by entropy minimization.
\newblock In {\em International Conference on Learning Representations}, 2020.

\bibitem{wang2018skipnet}
Xin Wang, Fisher Yu, Zi-Yi Dou, Trevor Darrell, and Joseph~E Gonzalez.
\newblock Skipnet: Learning dynamic routing in convolutional networks.
\newblock In {\em Proceedings of the European Conference on Computer Vision
  (ECCV)}, pages 409--424, 2018.

\bibitem{wolczyk2021zero}
Maciej Wo{\l}czyk, Bartosz W{\'o}jcik, Klaudia Ba{\l}azy, Igor~T Podolak, Jacek
  Tabor, Marek {\'S}mieja, and Tomasz Trzcinski.
\newblock Zero time waste: Recycling predictions in early exit neural networks.
\newblock {\em Advances in Neural Information Processing Systems},
  34:2516--2528, 2021.

\bibitem{xin2020deebert}
Ji~Xin, Raphael Tang, Jaejun Lee, Yaoliang Yu, and Jimmy Lin.
\newblock Deebert: Dynamic early exiting for accelerating bert inference.
\newblock {\em arXiv preprint arXiv:2004.12993}, 2020.

\bibitem{yoon2022hubert}
Ji~Won Yoon, Beom~Jun Woo, and Nam~Soo Kim.
\newblock Hubert-ee: Early exiting hubert for efficient speech recognition.
\newblock {\em arXiv preprint arXiv:2204.06328}, 2022.

\bibitem{zhao2019object}
Zhong-Qiu Zhao, Peng Zheng, Shou-tao Xu, and Xindong Wu.
\newblock Object detection with deep learning: A review.
\newblock {\em IEEE transactions on neural networks and learning systems},
  30(11):3212--3232, 2019.

\bibitem{zhou2018interpreting}
Bolei Zhou, David Bau, Aude Oliva, and Antonio Torralba.
\newblock Interpreting deep visual representations via network dissection.
\newblock {\em IEEE transactions on pattern analysis and machine intelligence},
  41(9):2131--2145, 2018.

\bibitem{zhou2020bert}
Wangchunshu Zhou, Canwen Xu, Tao Ge, Julian McAuley, Ke~Xu, and Furu Wei.
\newblock Bert loses patience: Fast and robust inference with early exit.
\newblock {\em Advances in Neural Information Processing Systems},
  33:18330--18341, 2020.

\end{thebibliography}

\clearpage
\appendix
\begin{center}
{\LARGE \bf Appendix}
\end{center}
We present a brief overview of the different early-exit strategies used with MEMs in Appendix~\ref{app:overview_strategies} followed by additional experimental results on the evaluation of practical early-exit strategies in Appendix~\ref{app:additional_experiments}. We conclude by providing details of the datasets and hyperparameters used in our experiments Appendix~\ref{app:experimental_details}. 

\section{Overview of various early-exit strategies}
\label{app:overview_strategies}
Here, we provide a brief overview of the existing early-exit strategies used in our work along with the details of the proposed nearest neighbor based strategy. Throughout this work, we refer to the confidence-based, patience-based, and nearest neighbor-based strategies as practical exit strategies in contrast to the oracle-based strategy as they are heuristic-based and do not require knowlegde of the true label.  The confidence-based strategy is referenced similarly in prior works \cite{kaya2019shallow}.

{\bf Oracle-based exit strategy:}  This strategy returns the earliest exit that correctly classifies a sample. This strategy requires knowing the label of the test sample and is not a practical early-exit strategy. If a point is never correct in the network the strategy exits at the final layer. 

{\bf Confidence-based strategy:} This strategy was proposed in \cite{kaya2019shallow} and uses the estimated probability of a sample belonging to a class as the confidence of the classifier. In particular, at an exit $i$, the strategy checks if the prediction at this exit i.e. $\max_{j \in \mathcal{Y}} f^i_j(x)$ is greater than a threshold, where $\mathcal{Y}$ denotes the set of labels and $f^i$ is the classifier at exit $i$. If none of the exits are confident enough, the most confident exit is used for the sample.

{\bf Patience-based strategy:}
The patience-based early-exit strategy was proposed in \cite{zhou2020bert} which allows a sample to exit the network if $t$ consecutive exits produce the same prediction. If any $t$ consecutive exits do not produce the same predictions, the sample exits at the last layer.

{\bf Nearest neighbor-based strategy:}
This strategy measures the confidence in the exit's predictions based on the support from the in-distribution data, which was used to train the ME model. Given a sample at inference time, the strategy first obtains the prediction of the exit $i$ in the same manner as the confidence-based strategy, i.e. the class that has the highest softmax probability. Then the strategy computes the prediction of the exit $i$ on k-nearest neighbors of the test sample in the training data and computes the number of neighbors whose predictions match the prediction of the new test sample (called $\tau$). If $\tau$ is greater than a certain threshold then the strategy allows the test sample to exit the network using the exit $i$. If the threshold is not met at any of the exits then the strategy uses the last layer as the exit. 
We use Faiss \cite{johnson2019billion}, for efficient nearest neighbor search and used {\tt faiss.IndexFlatL2} as the indexing method with normalized representation space (layer before the softmax layer at exit $i$) distance.

\section{Additional experiments}
\label{app:additional_experiments}
\begin{figure*}[tb]
  \centering{
  \subfigure[VGG on CIFAR-10]{\includegraphics[width=0.24\columnwidth]{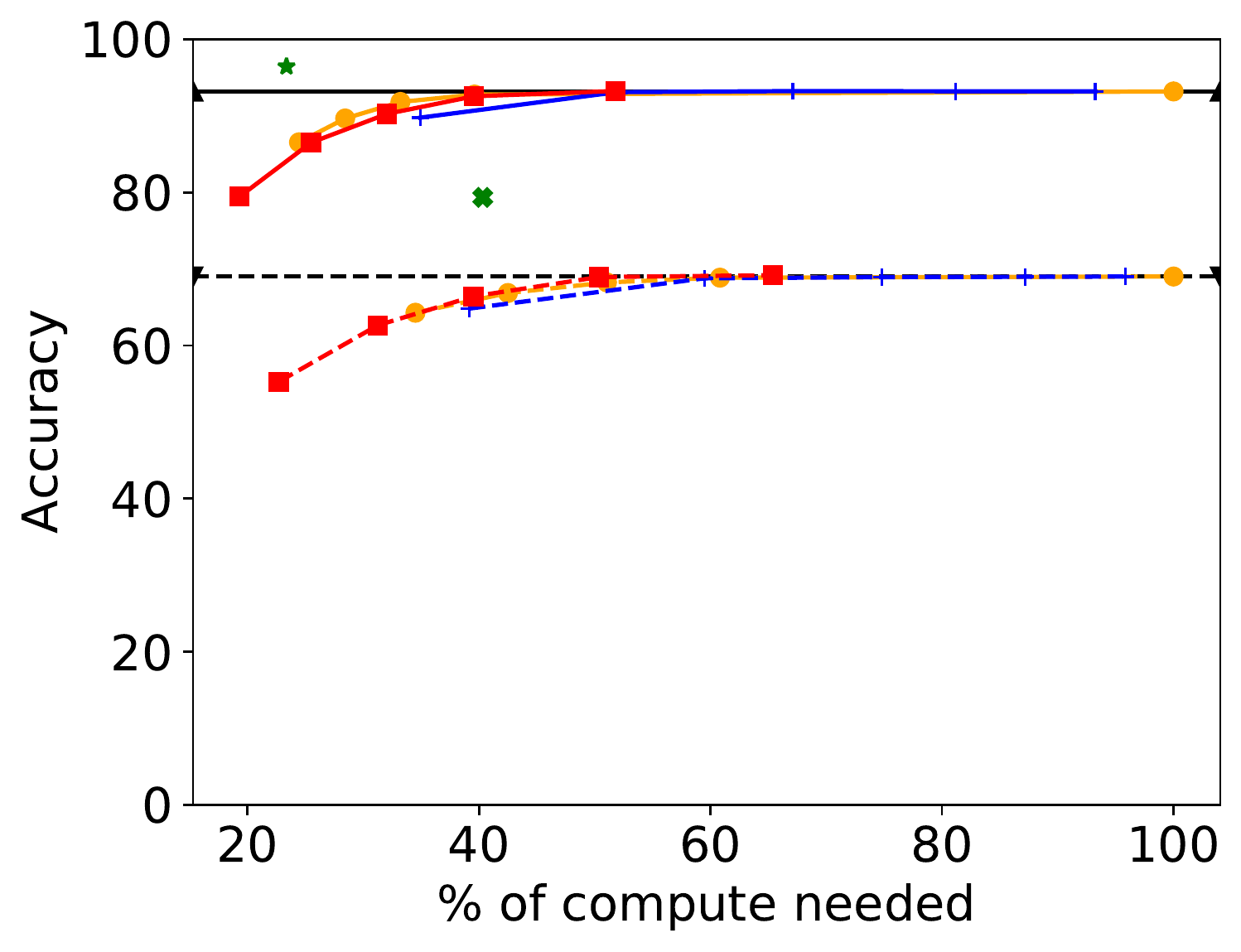}}
  \subfigure[ResNet on CIFAR-10]{\includegraphics[width=0.24\columnwidth]{Images/accuracy_vs_efficiency_resnet_cifar10_augmixFalse.pdf}}
  \subfigure[VGG on CIFAR-100]{\includegraphics[width=0.24\columnwidth]{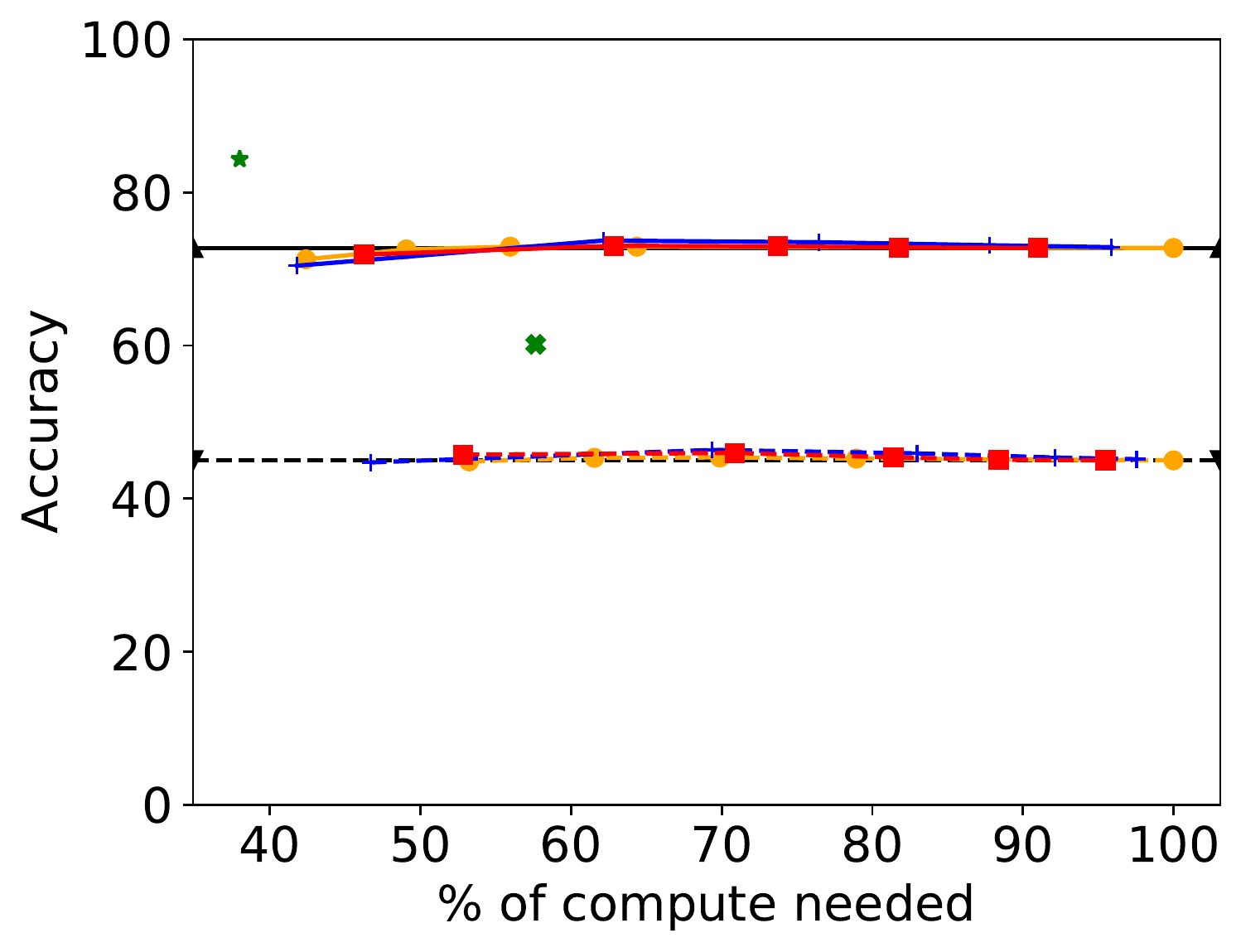}}
  \subfigure[ResNet on CIFAR-100]{\includegraphics[width=0.24\columnwidth]{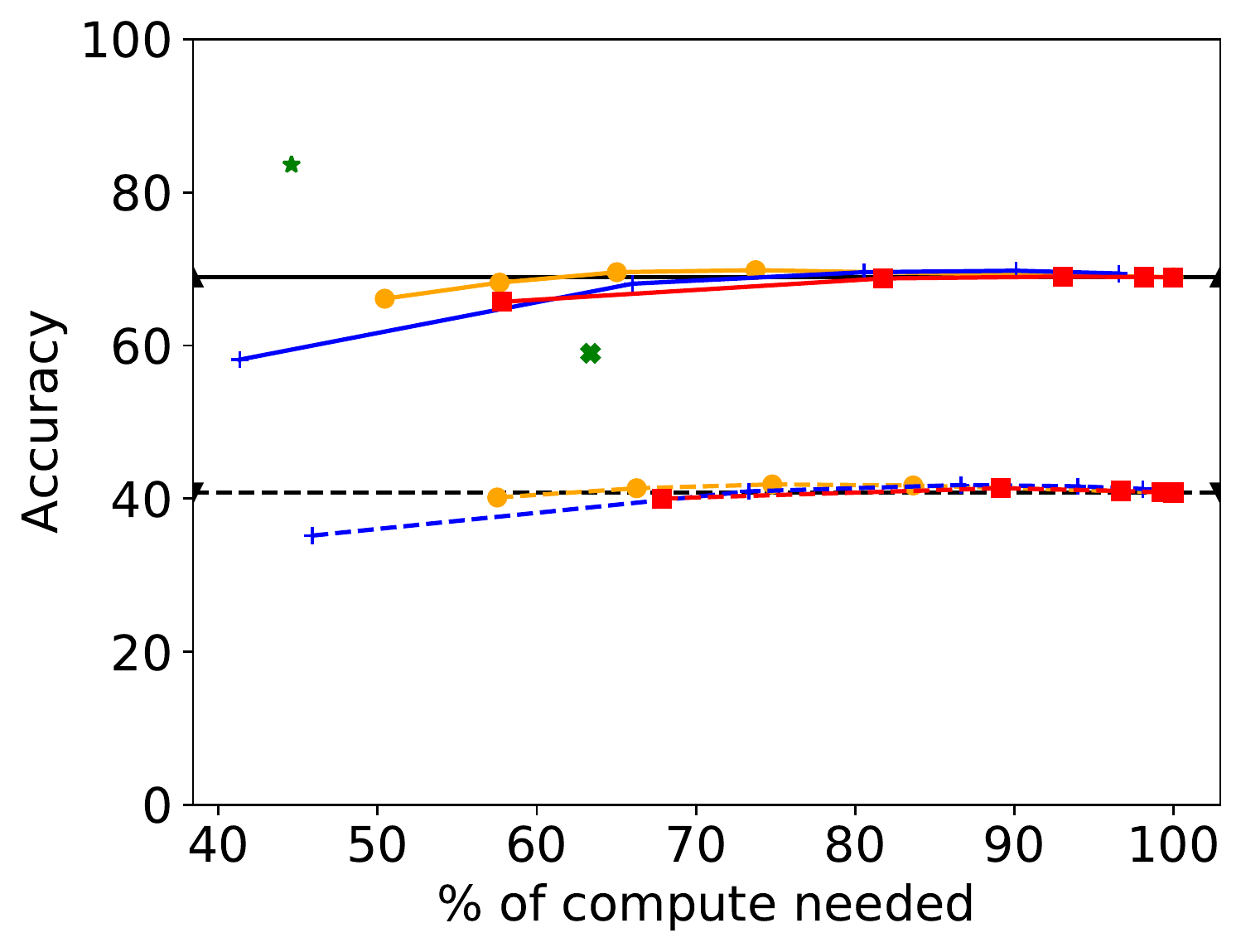}}
  }
  \caption{Comparison of accuracy of MEMs with two different architectures (VGG/ResNet-56) using oracle-based and various practical early-exit strategies on clean and corrupted CIFAR-10/100 datasets. 
  }
  \label{fig:accuracy_dist_shift}
\end{figure*}

\begin{figure*}[tb]
  \centering{
  \subfigure[VGG on CIFAR-10]{\includegraphics[width=0.24\columnwidth]{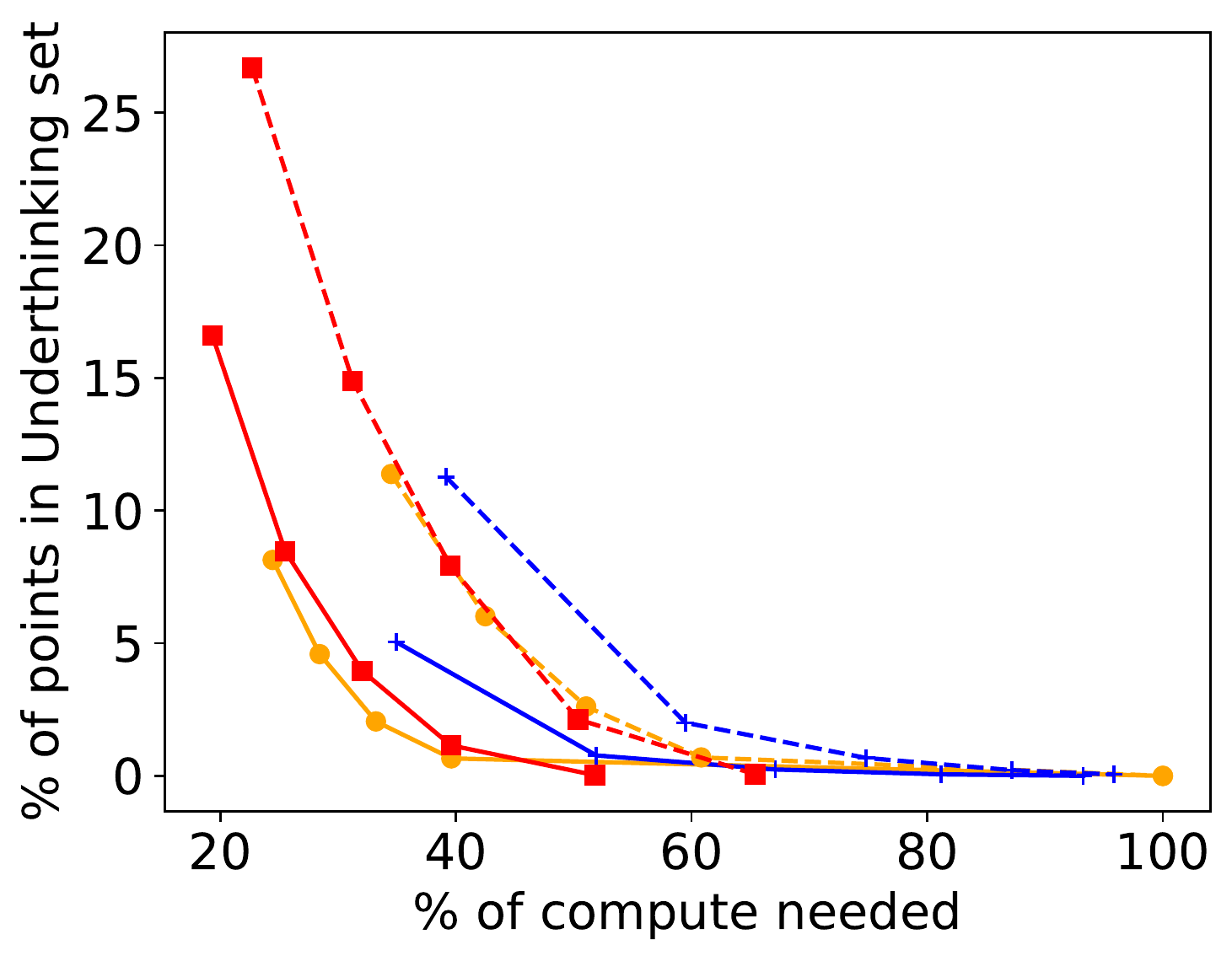}}
  \subfigure[ResNet on CIFAR-10]{\includegraphics[width=0.24\columnwidth]{Images/underthinking_vs_efficiency_resnet_cifar10_augmixFalse.pdf}}
  \subfigure[VGG on CIFAR-100]{\includegraphics[width=0.24\columnwidth]{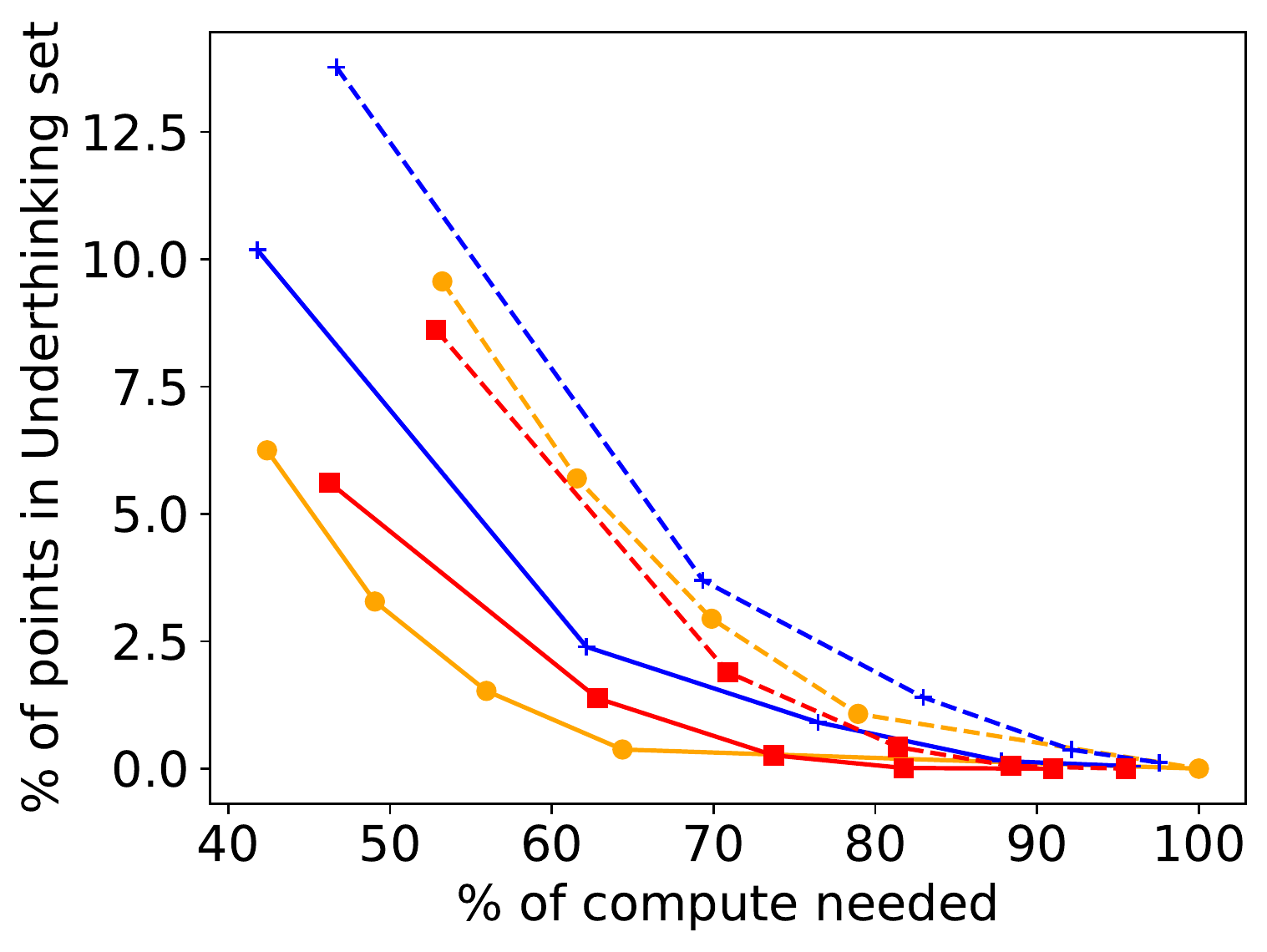}}
  \subfigure[ResNet on CIFAR-100]{\includegraphics[width=0.24\columnwidth]{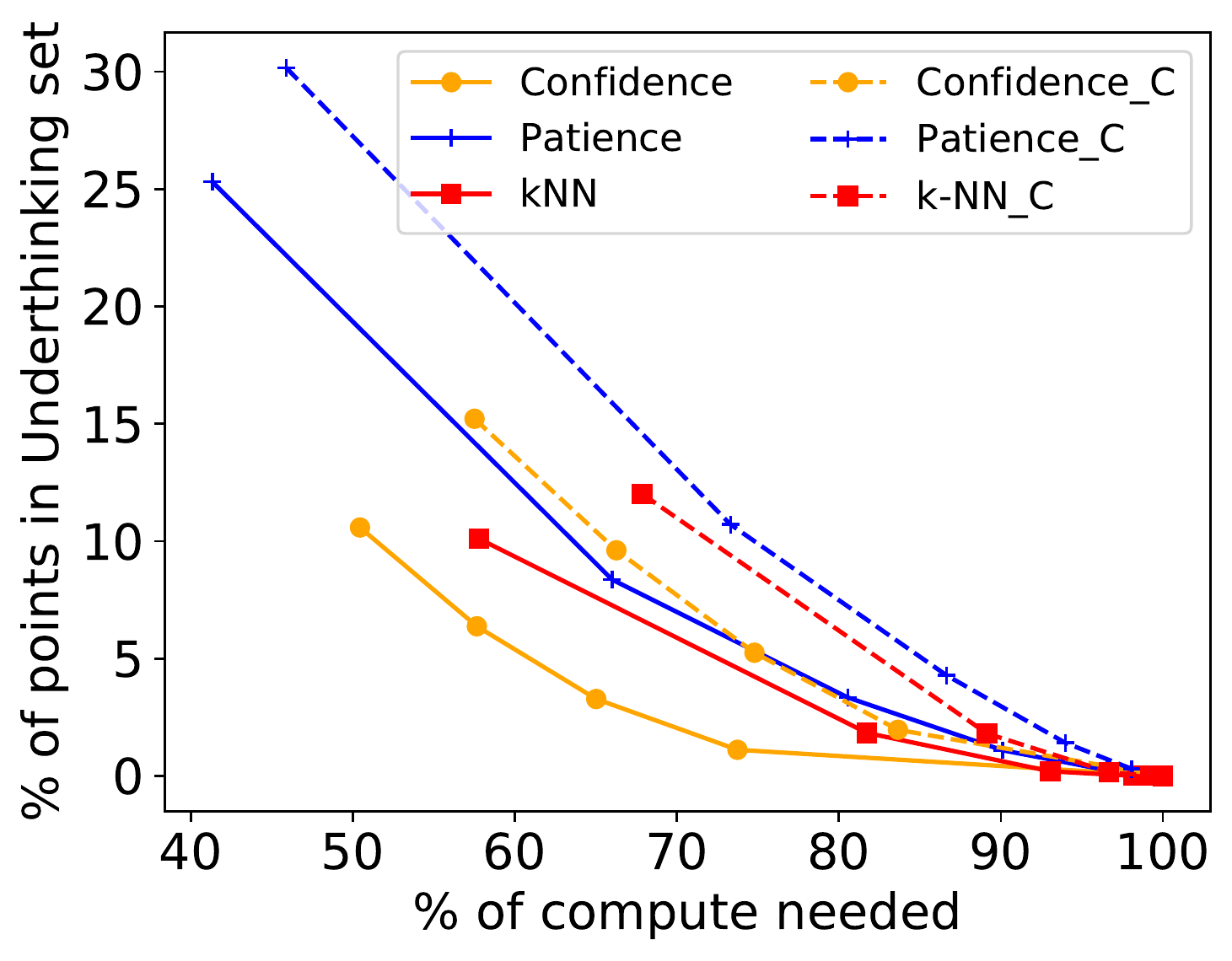}}
  }
  \caption{Increase in the size of the underthinking set $UT$ of MEMs with two different architectures (VGG/ResNet-56) using various practical early-exit strategies on clean and corrupted CIFAR-10/100 datasets.
  }
  \label{fig:underthinking_dist_shift}
\end{figure*}

\begin{figure*}[tb]
  \centering{
  \subfigure[VGG on CIFAR-10]{\includegraphics[width=0.24\columnwidth]{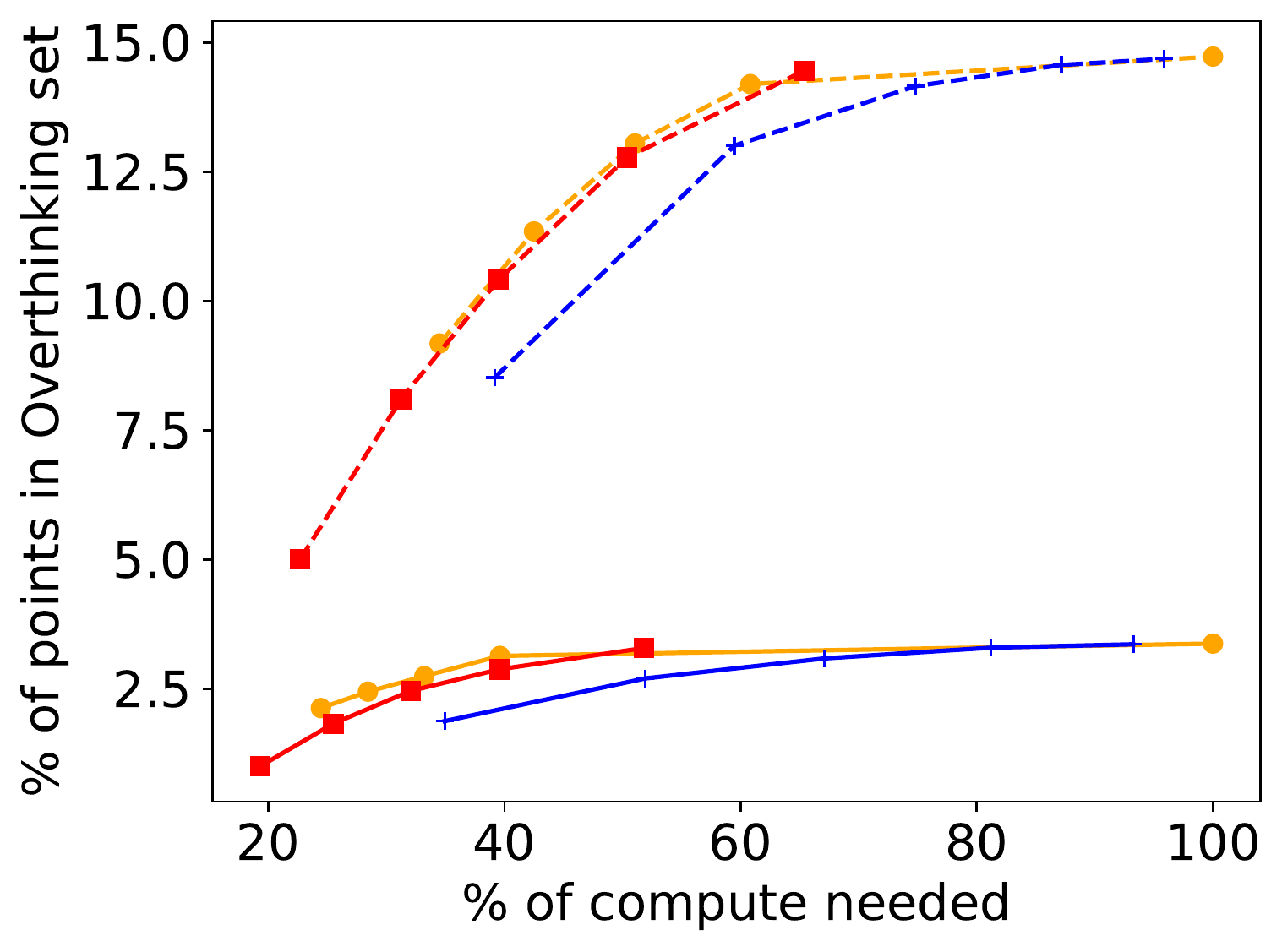}}
  \subfigure[ResNet on CIFAR-10]{\includegraphics[width=0.24\columnwidth]{Images/overthinking_vs_efficiency_resnet_cifar10_augmixFalse.pdf}}
  \subfigure[VGG on CIFAR-100]{\includegraphics[width=0.24\columnwidth]{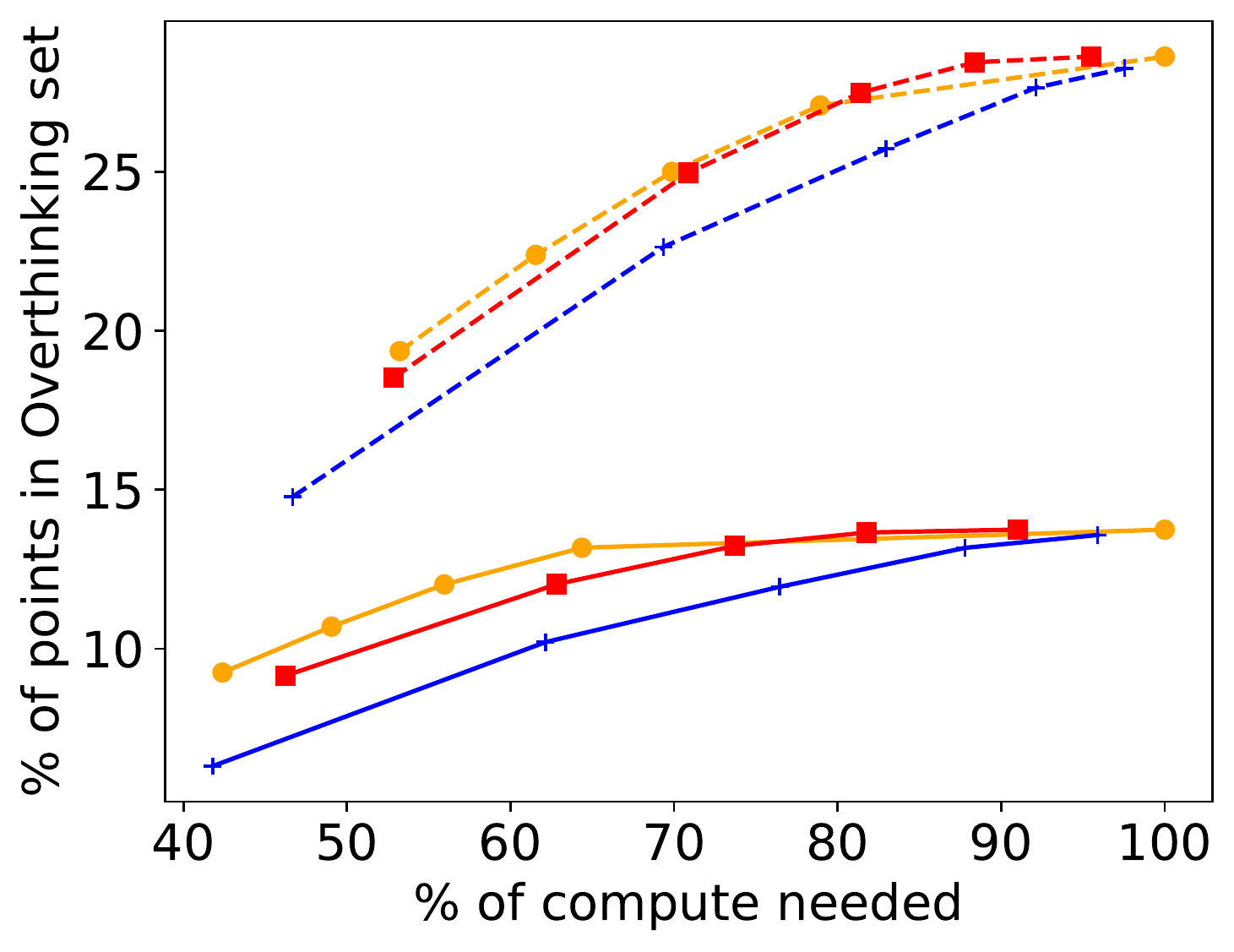}}
  \subfigure[ResNet on CIFAR-100]{\includegraphics[width=0.24\columnwidth]{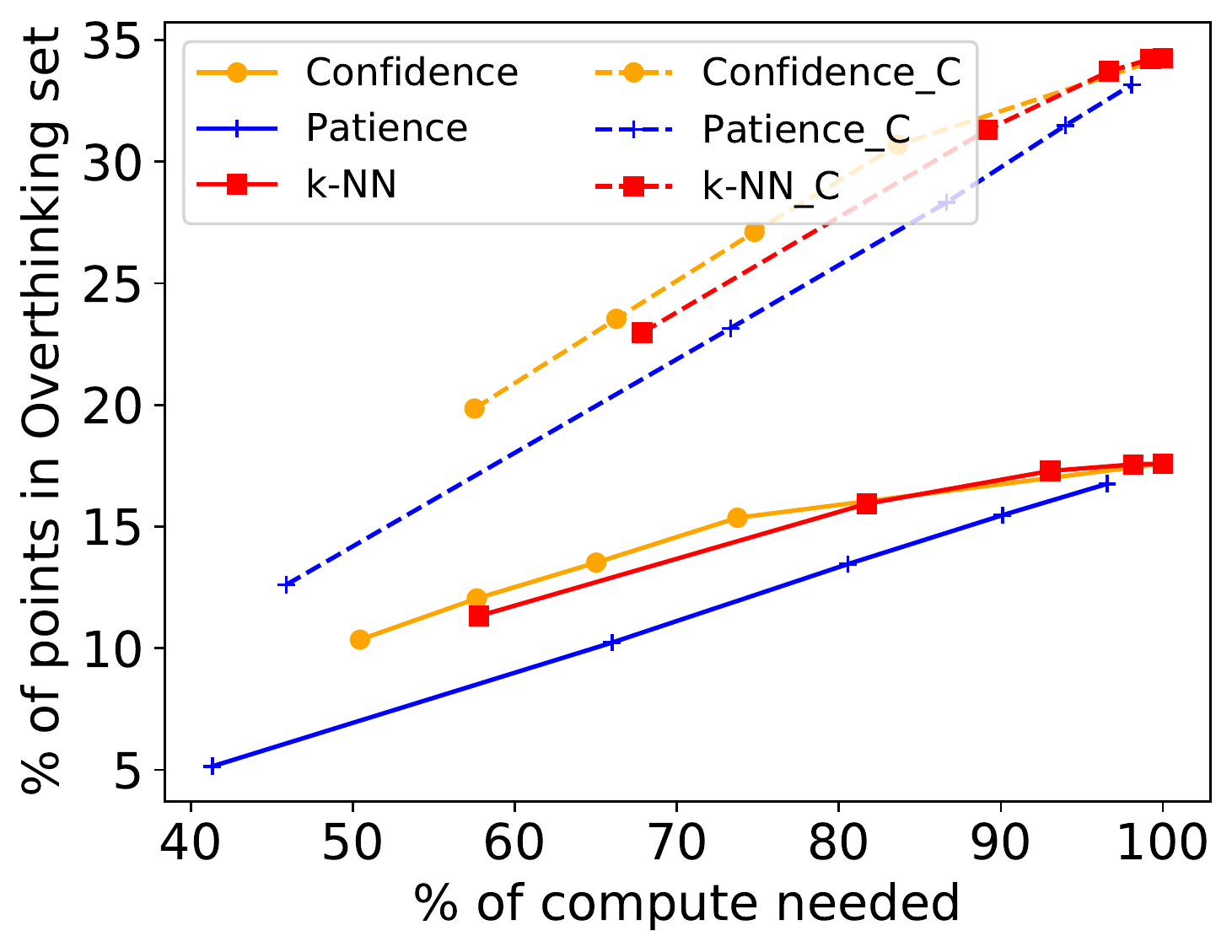}}
  }
  \caption{Increase in the size of the overthinking set $OT$ of MEMs with two different architectures (VGG/ResNet-56) using various practical early-exit strategies on clean and corrupted CIFAR-10/100 datasets.
  }
  \label{fig:overthinking_dist_shift}
\end{figure*}

In this section, we discuss the behavior of various practical early-exit strategies on MEMs with different backbone architectures, namely VGG-16 and ResNet-56 trained using CIFAR-10/100 datasets. 
Similar to the results reported in the main paper in Figure~\ref{fig:comparison_dist_shift}(a), we find that the gap between the accuracy obtained by using oracle-based and practical early-exit strategies widens in presence of distribution shift. 
As shown in Figure~\ref{fig:accuracy_dist_shift}, the problem becomes more severe on the CIFAR-100 dataset where the gap widens by more than 5\% compared to the gap between the oracle-based and practical early-exit strategies on in-distribution data. 
The significant increase in the size of the underthinking and overthinking sets in presence of distribution shift, in Figs.~\ref{fig:underthinking_dist_shift} and~\ref{fig:overthinking_dist_shift} demonstrates the reason for the increased gap similar to Figure~\ref{fig:comparison_dist_shift}(b,c).

\begin{figure*}[tb]
  \centering{
  \subfigure[VGG on CIFAR-10]{\includegraphics[width=0.24\columnwidth]{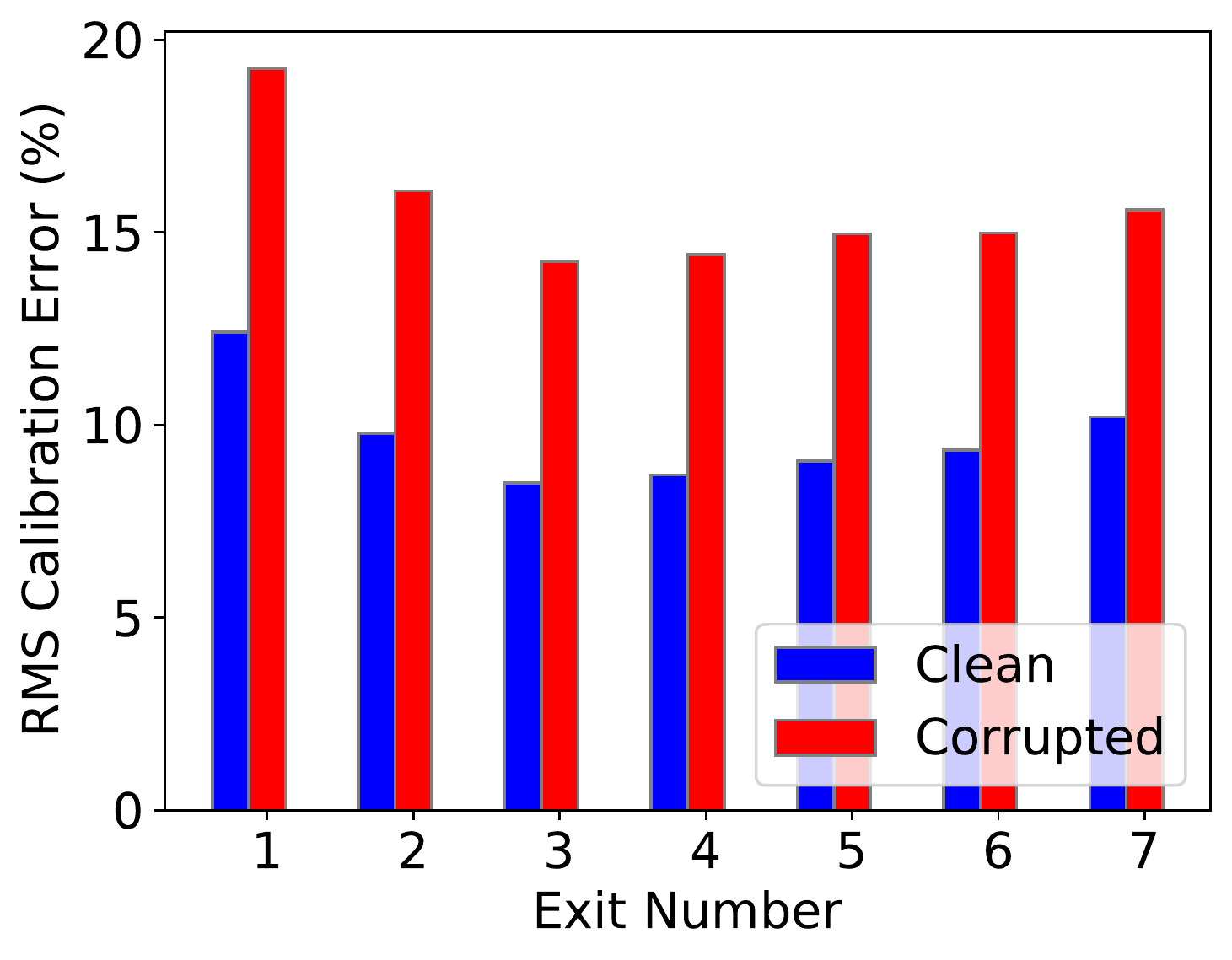}}
  \subfigure[ResNet on CIFAR-10]{\includegraphics[width=0.24\columnwidth]{Images/exit_wise_calibration_resnet_cifar10_augmixFalse.pdf}}
  \subfigure[VGG on CIFAR-100]{\includegraphics[width=0.24\columnwidth]{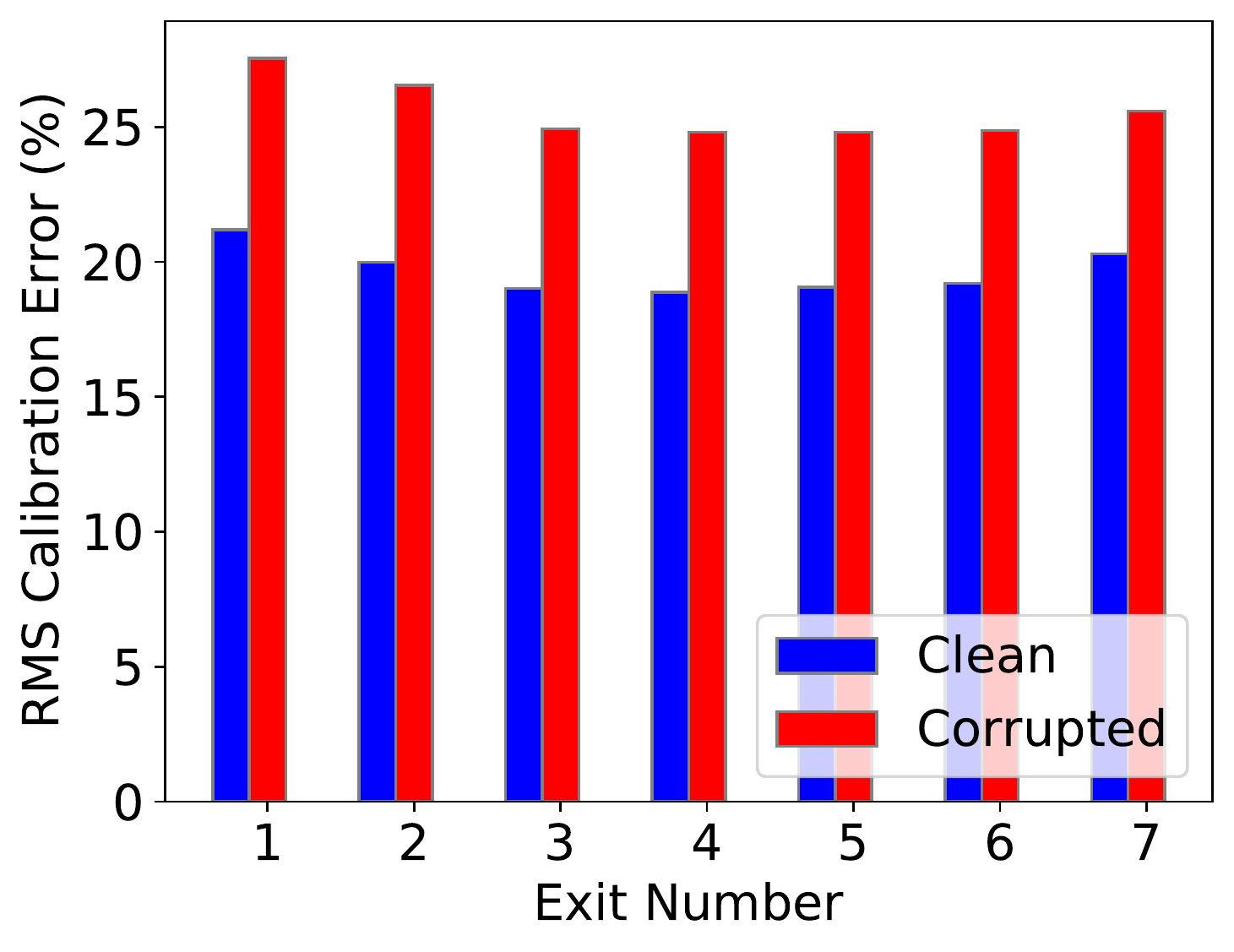}}
  \subfigure[ResNet on CIFAR-100]{\includegraphics[width=0.24\columnwidth]{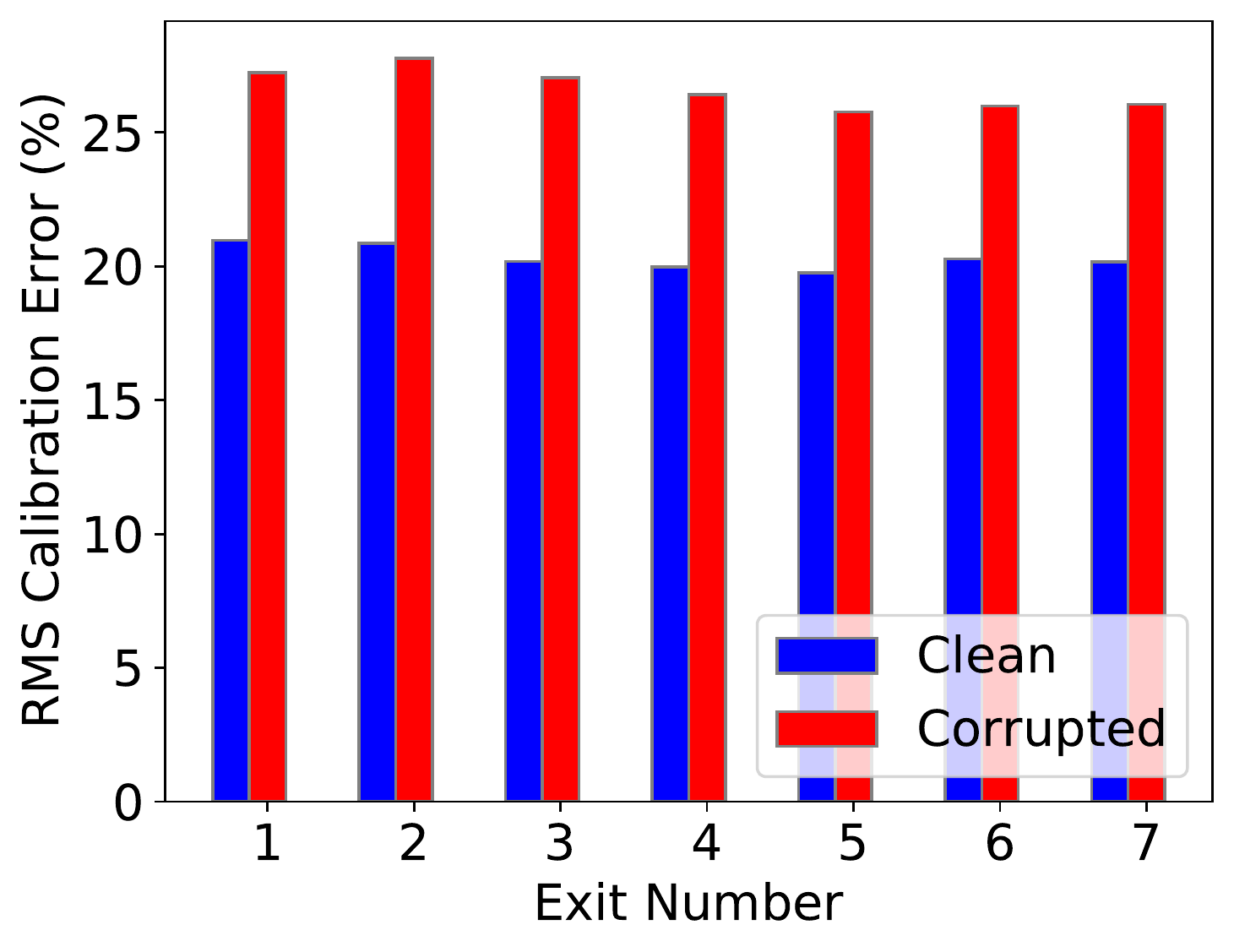}}
  }
  \caption{Worsening of the RMS calibration error of all exits in the MEMs with two different architectures (VGG/ResNet-56) in the presence of distribution shifts leading to increased underthinking.
  }
  \label{fig:calibration_dist_shift}
\end{figure*}

\begin{figure*}[tb]
  \centering{
  \subfigure[VGG on CIFAR-10]{\includegraphics[width=0.24\columnwidth]{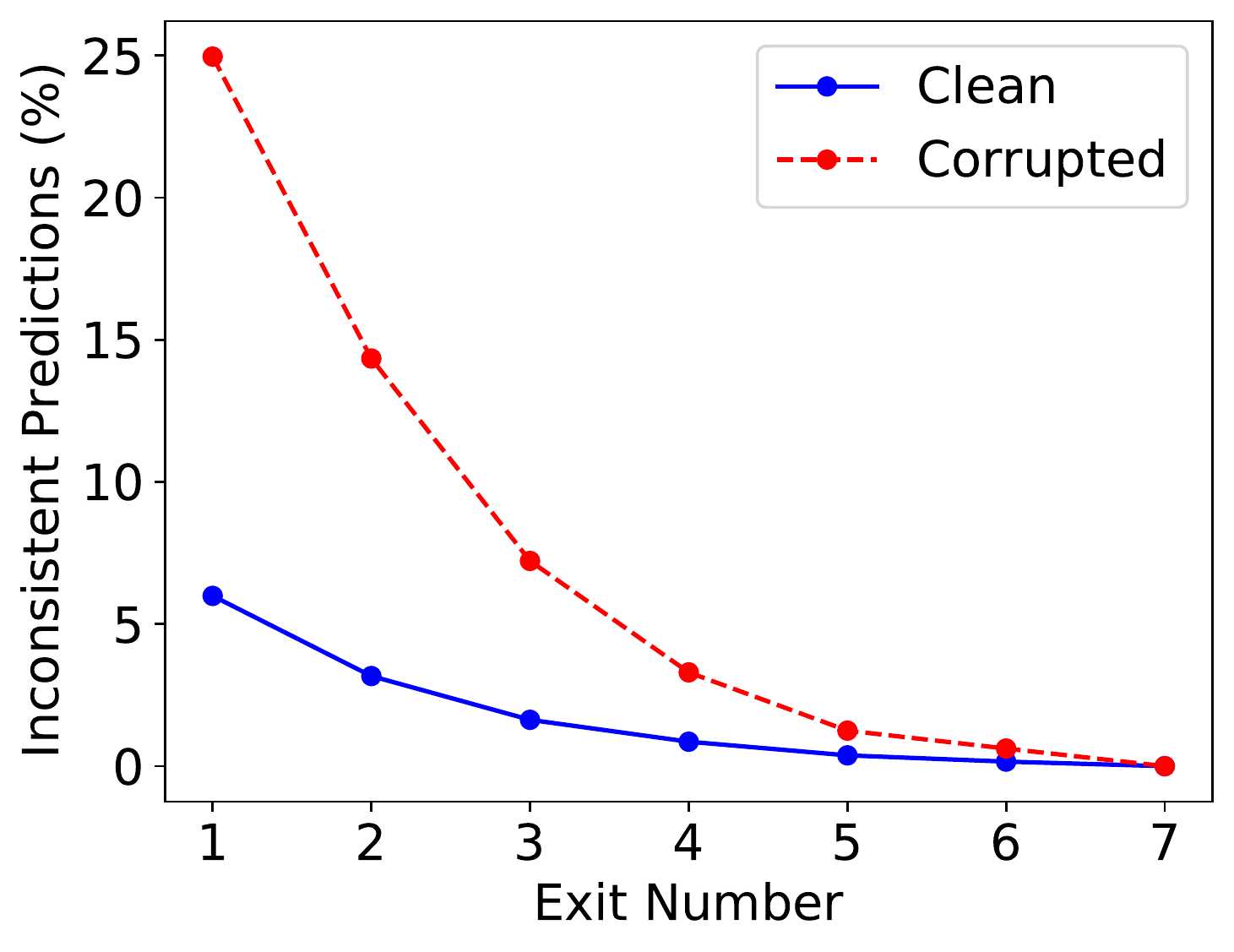}}
  \subfigure[ResNet on CIFAR-10]{\includegraphics[width=0.24\columnwidth]{Images/exit_wise_inconsistent_preds_resnet_cifar10_augmixFalse.pdf}}
  \subfigure[VGG on CIFAR-100]{\includegraphics[width=0.24\columnwidth]{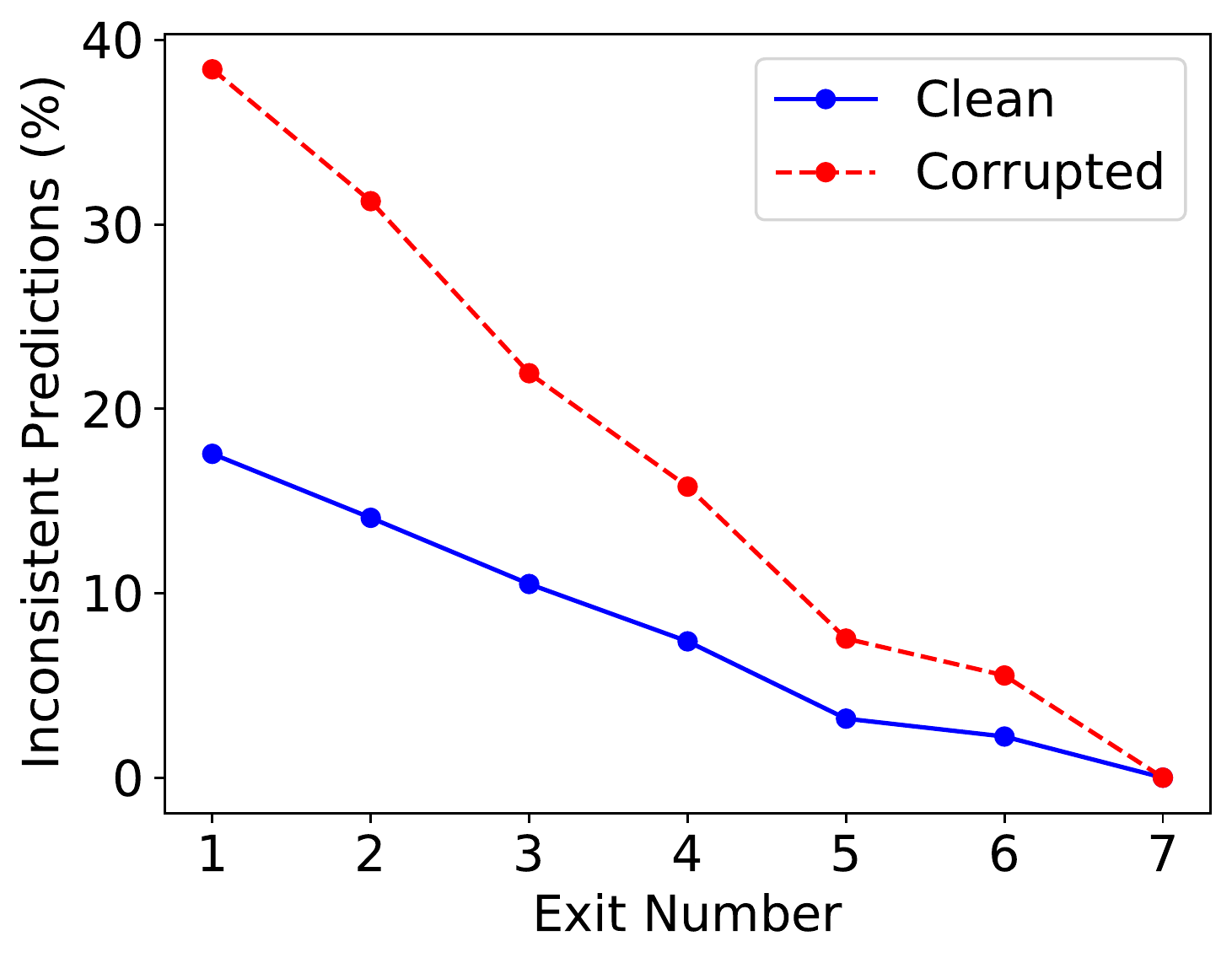}}
  \subfigure[ResNet on CIFAR-100]{\includegraphics[width=0.24\columnwidth]{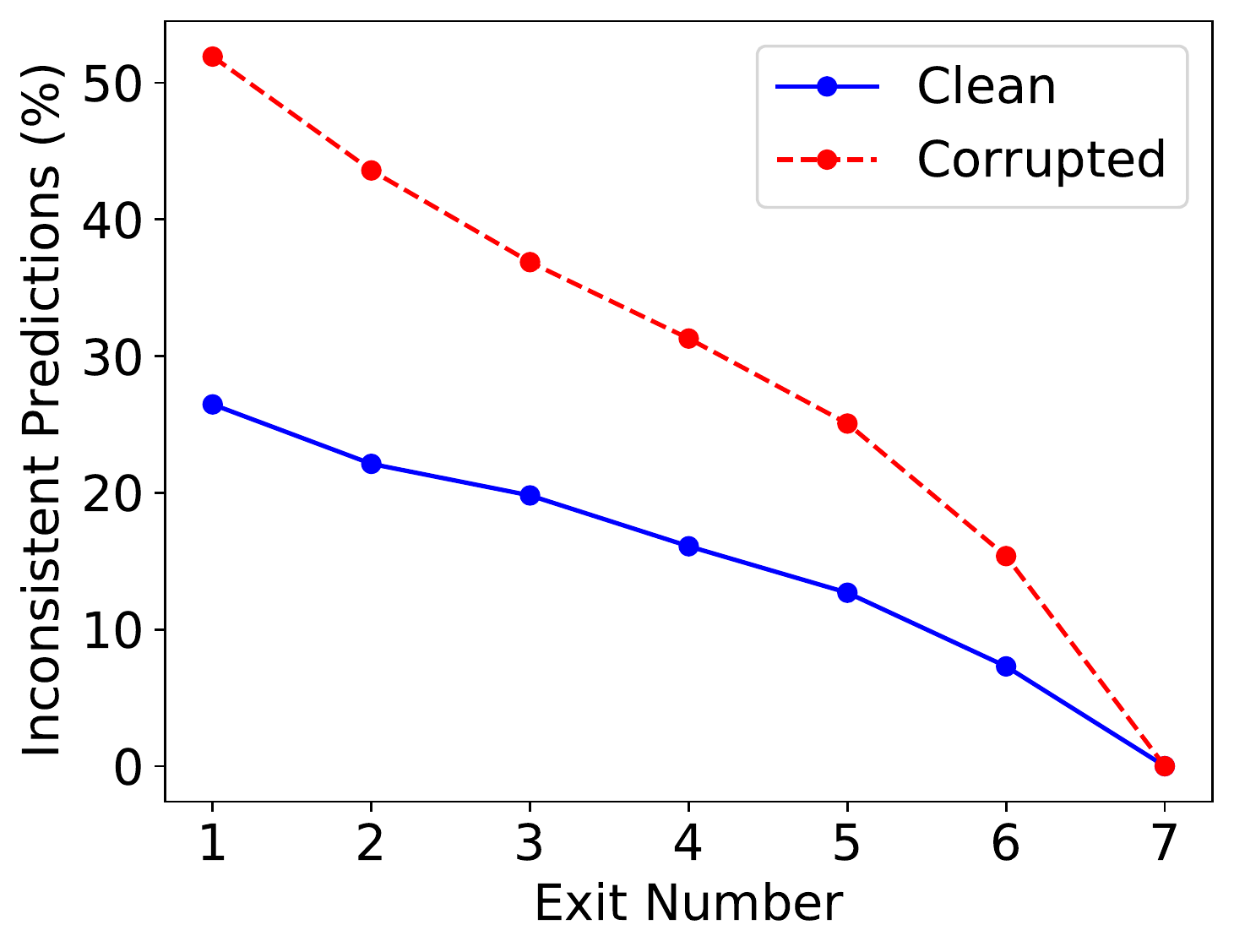}}
  }
  \caption{Increase in the inconsistency of the predictions of all exits in the MEMs with two different architectures (VGG/ResNet-56) in the presence of distribution shifts, leading to increased overthinking.
  }
  \label{fig:inconsistent_predictions_dist_shift}
\end{figure*}

We also evaluate the exit-wise calibration error and the inconsistency in the predictions of the model at different exits to understand the reason for increased underthinking and overthinking. 
Similar to Figure~\ref{fig:calibration_inconsistent_preds}, we observe in Figure~\ref{fig:calibration_dist_shift} that the calibration error increases in presence of distribution shift for all the exits.
This suggests that softmax probabilities of the earlier exits cannot be trusted and techniques that improve the calibration of the models in presence of distribution shift can be used to reduce underthinking.
We see similar behavior in the number of inconsistent predictions made by the model (see Figure~\ref{fig:inconsistent_predictions_dist_shift}), which is the percentage of samples that are correctly classified at an exit but got misclassified at a later exit. A high percentage of inconsistent predictions in a model creates difficulties for the early-exit strategies since failure to stop the computation at a particular exit that correctly classified a sample might lead to misclassification if the later exit does not correctly classify the sample. 
Thus, new training strategies which are able to reduce inconsistent predictions are desirable since once a sample is correctly classified at an exit it could remain correctly classified at all later exits. This would lead to decreased overthinking and increased accuracy of the MEMs.

\begin{figure*}[tb]
  \centering{
  \subfigure[VGG on CIFAR-10]{\includegraphics[width=0.24\columnwidth]{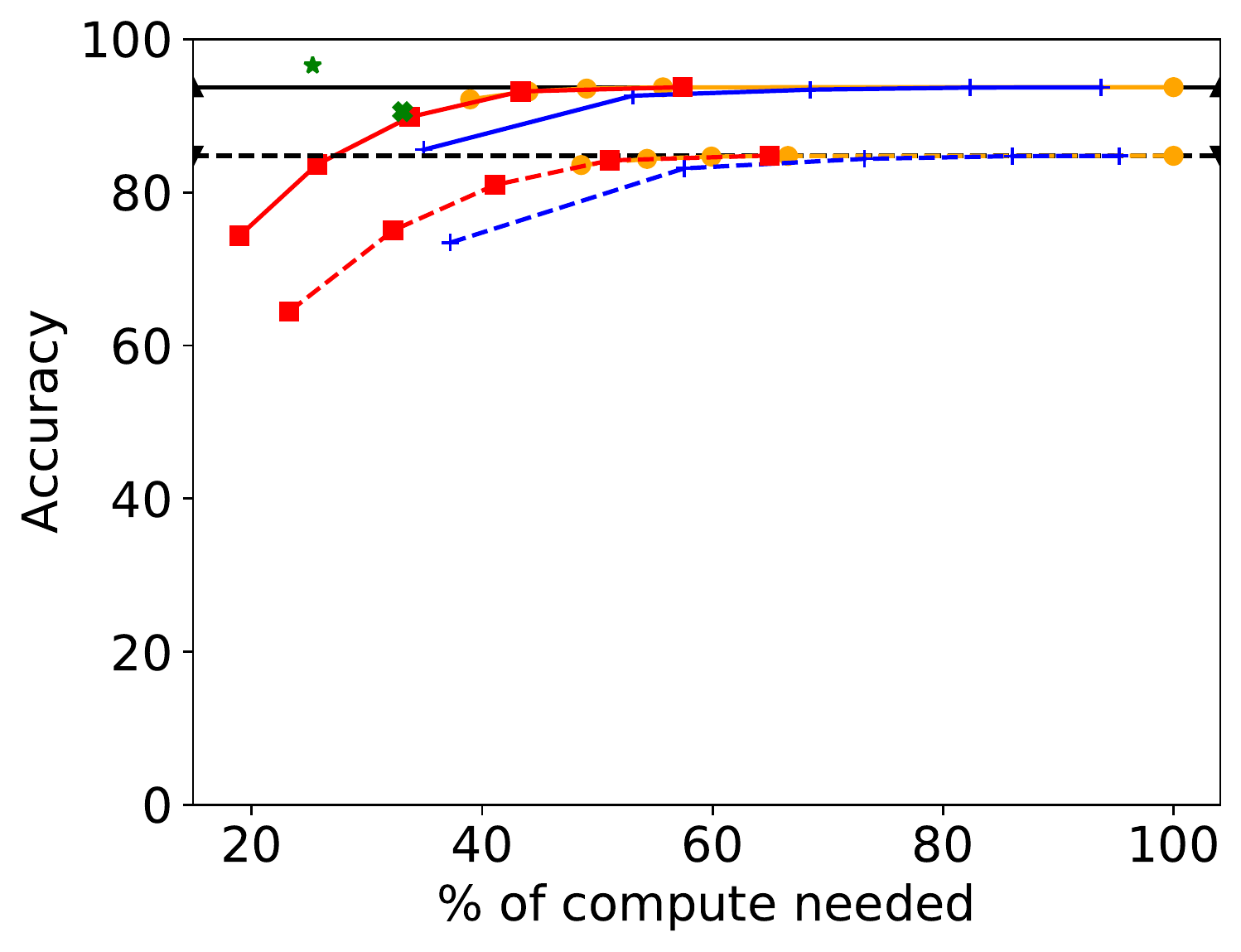}}
  \subfigure[ResNet on CIFAR-10]{\includegraphics[width=0.24\columnwidth]{Images/accuracy_vs_efficiency_resnet_cifar10_augmixTrue.pdf}}
  \subfigure[VGG on CIFAR-100]{\includegraphics[width=0.24\columnwidth]{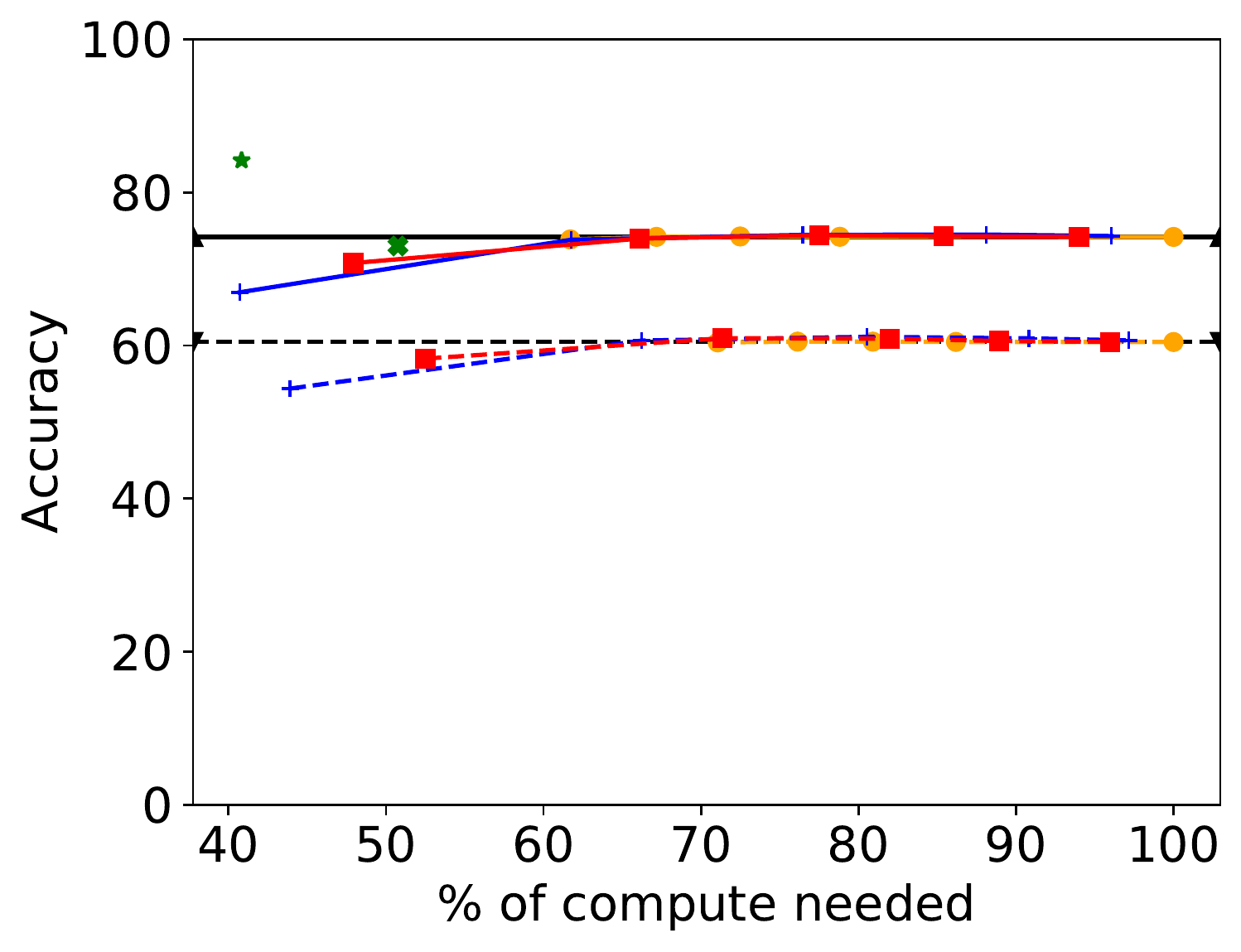}}
  \subfigure[ResNet on CIFAR-100]{\includegraphics[width=0.24\columnwidth]{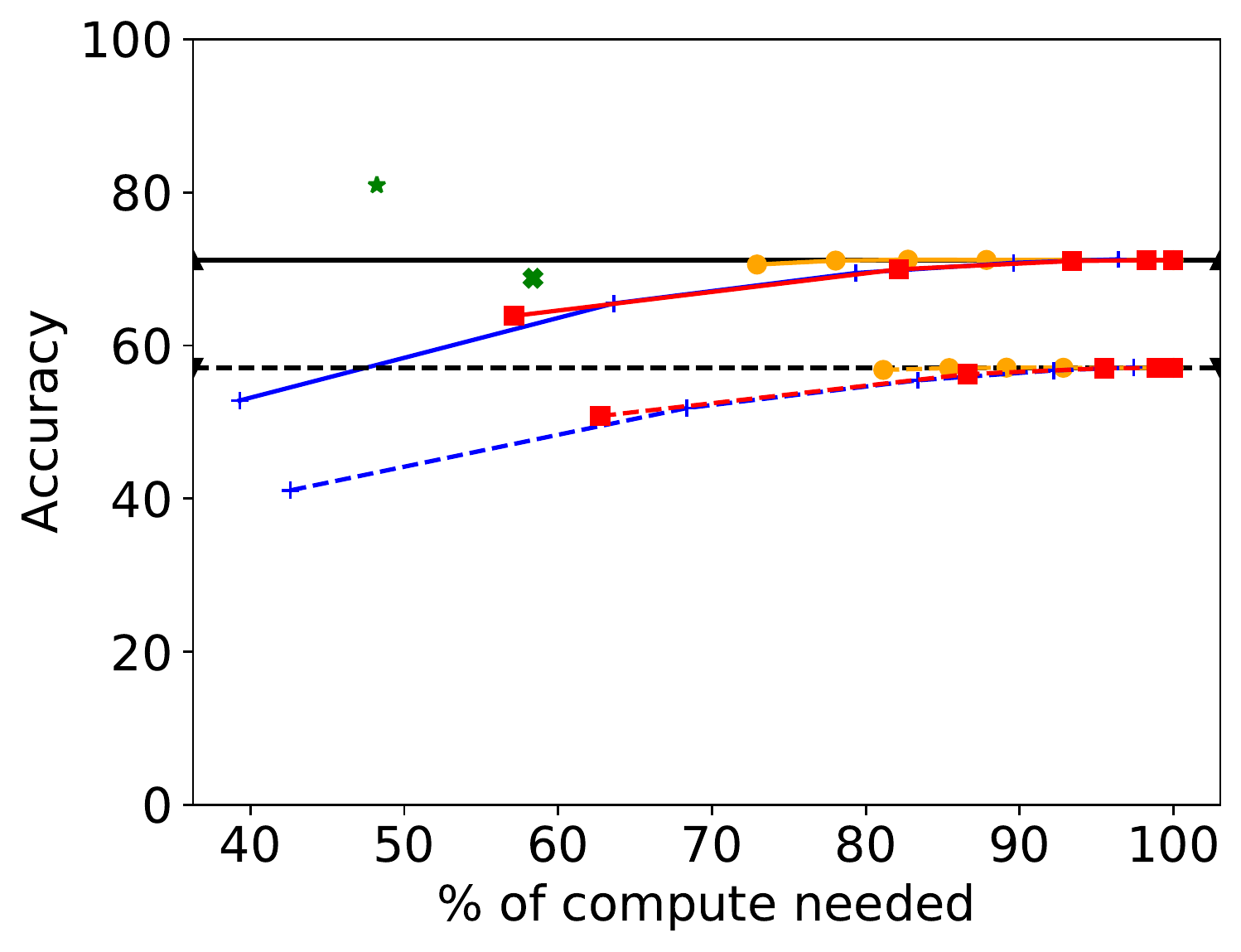}}
  }
  \caption{Comparison of accuracy of MEMs trained with AugMix with two different architectures (VGG/ResNet-56) using oracle-based and various practical early-exit strategies on clean and corrupted CIFAR-10/100 datasets. 
  }
  \label{fig:augmix_accuracy_dist_shift}
\end{figure*}

\begin{figure*}[tb]
  \centering{
  \subfigure[VGG on CIFAR-10]{\includegraphics[width=0.24\columnwidth]{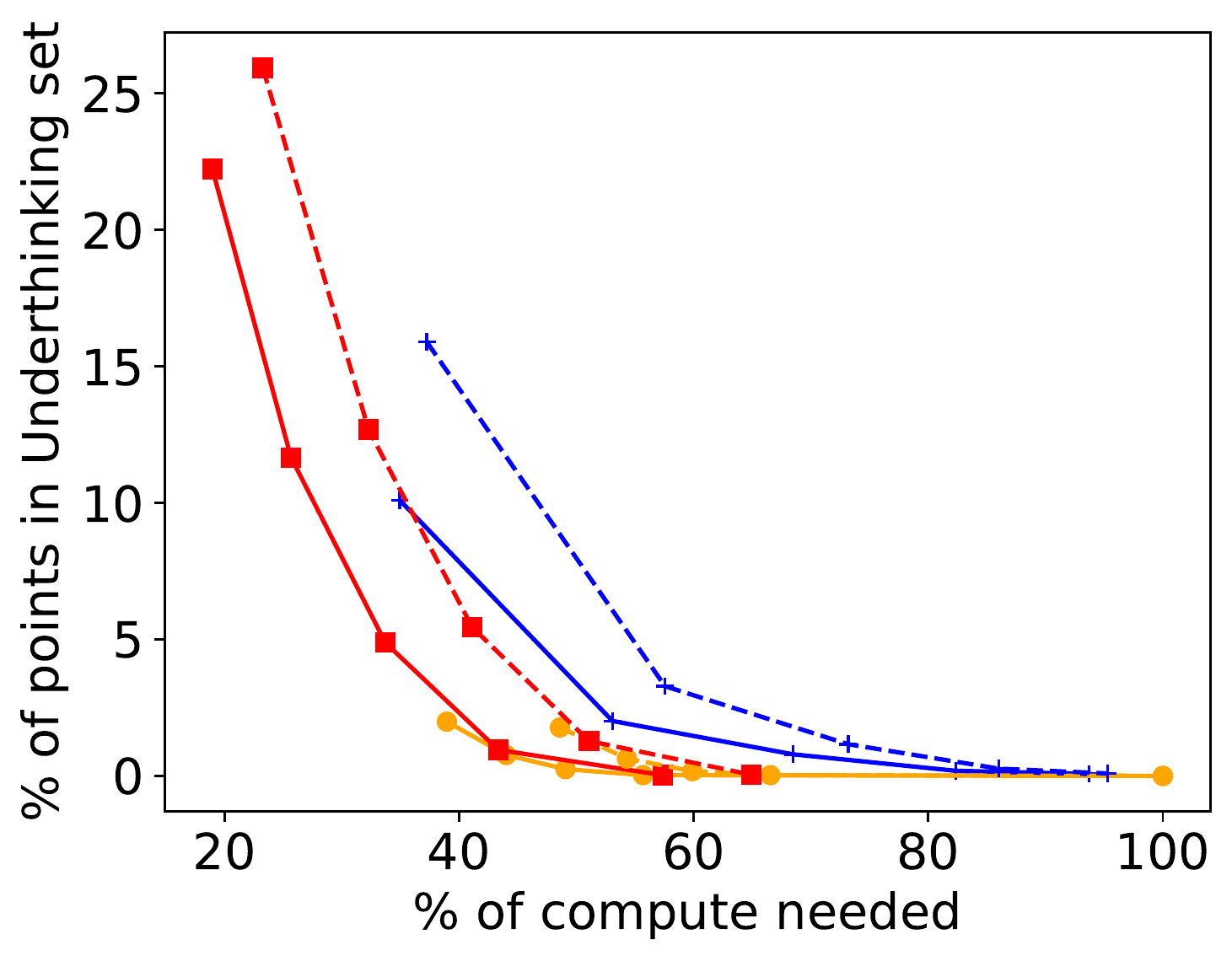}}
  \subfigure[ResNet on CIFAR-10]{\includegraphics[width=0.24\columnwidth]{Images/underthinking_vs_efficiency_resnet_cifar10_augmixTrue.pdf}}
  \subfigure[VGG on CIFAR-100]{\includegraphics[width=0.24\columnwidth]{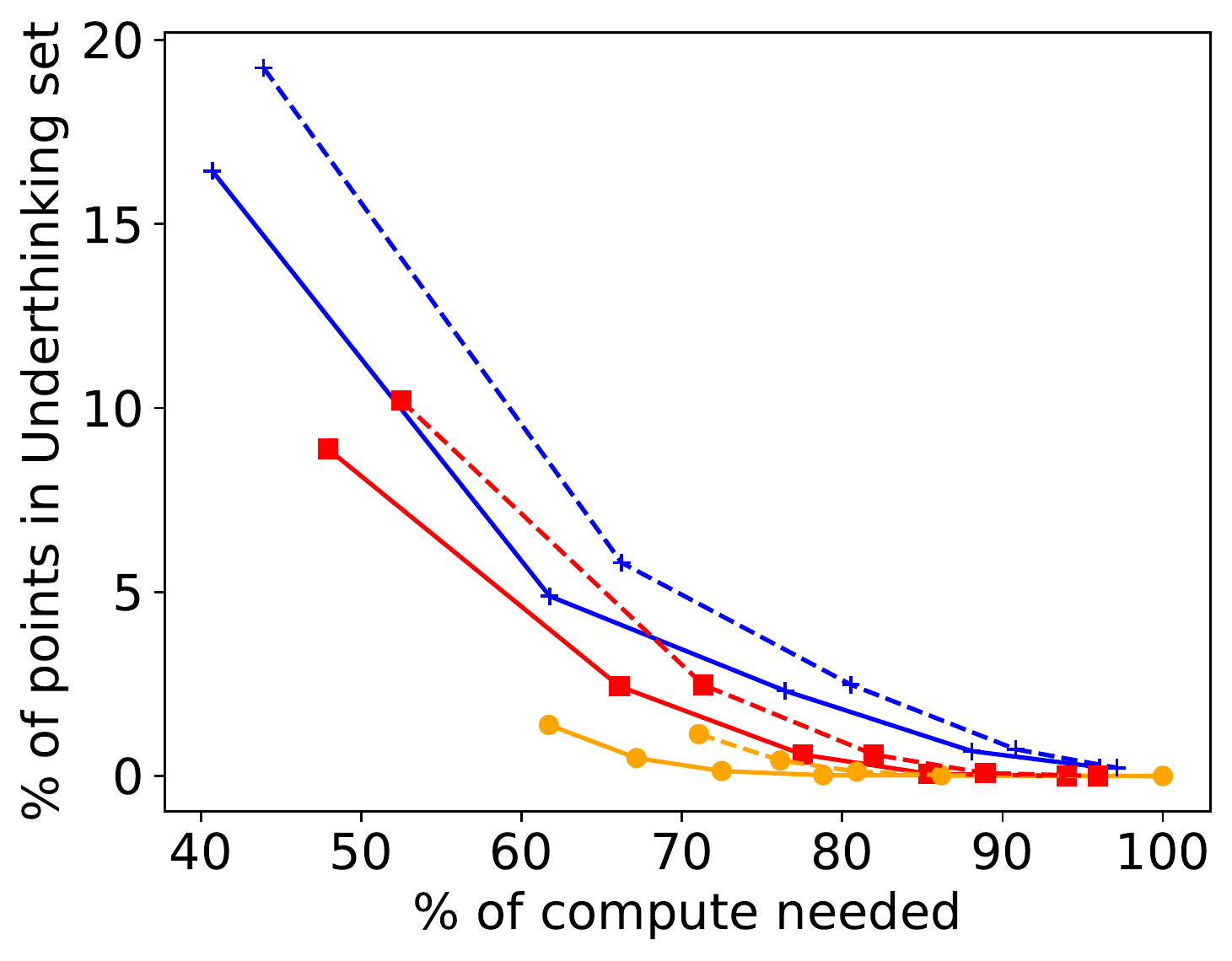}}
  \subfigure[ResNet on CIFAR-100]{\includegraphics[width=0.24\columnwidth]{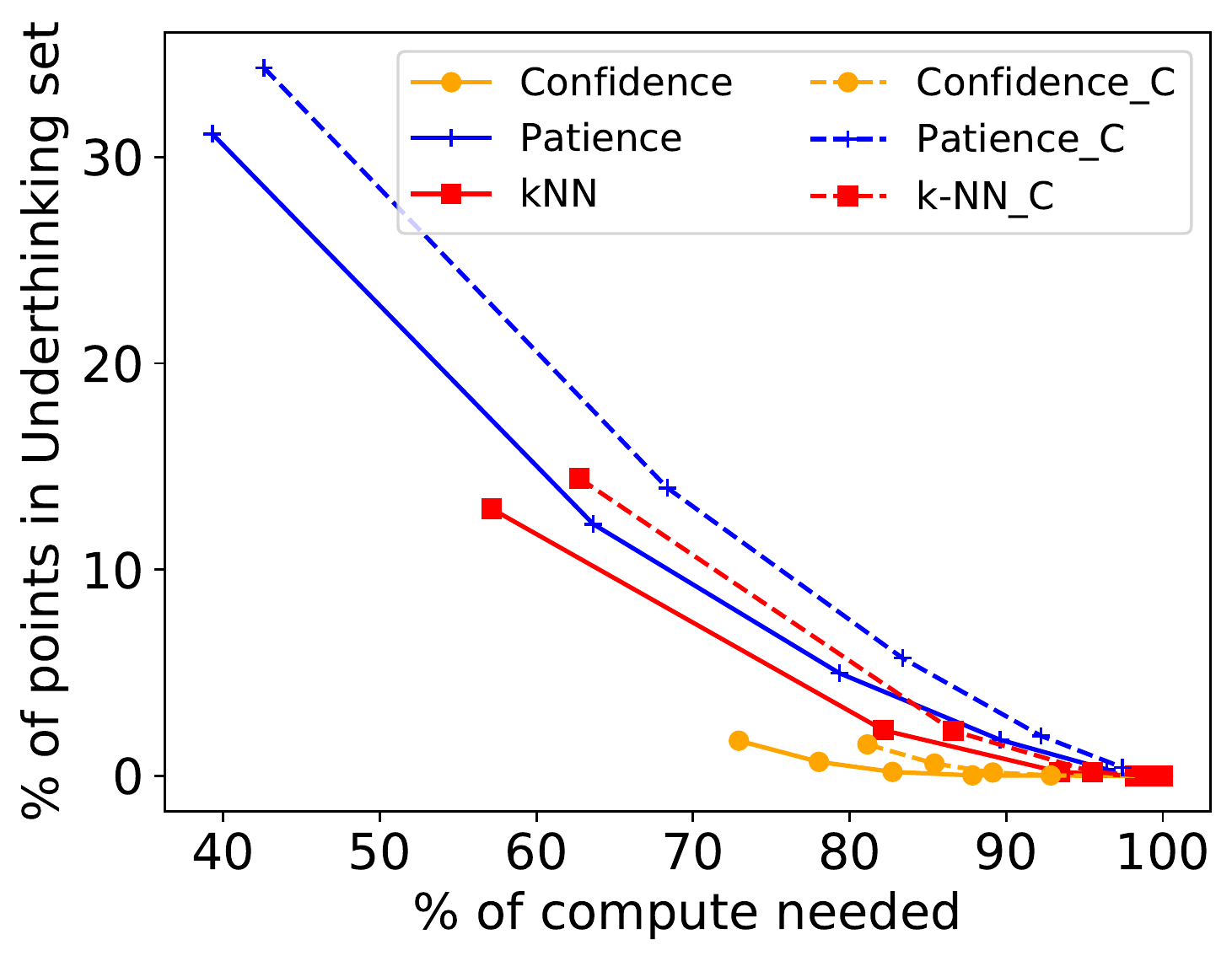}}
  }
  \caption{Decrease in the size of the underthinking set $UT$ of MEMs when trained with AugMix with two different architectures (VGG/ResNet-56) using various practical early-exit strategies on clean and corrupted CIFAR-10/100 datasets.
  }
  \label{fig:augmix_underthinking_dist_shift}
\end{figure*}

\begin{figure*}[tb]
  \centering{
  \subfigure[VGG on CIFAR-10]{\includegraphics[width=0.24\columnwidth]{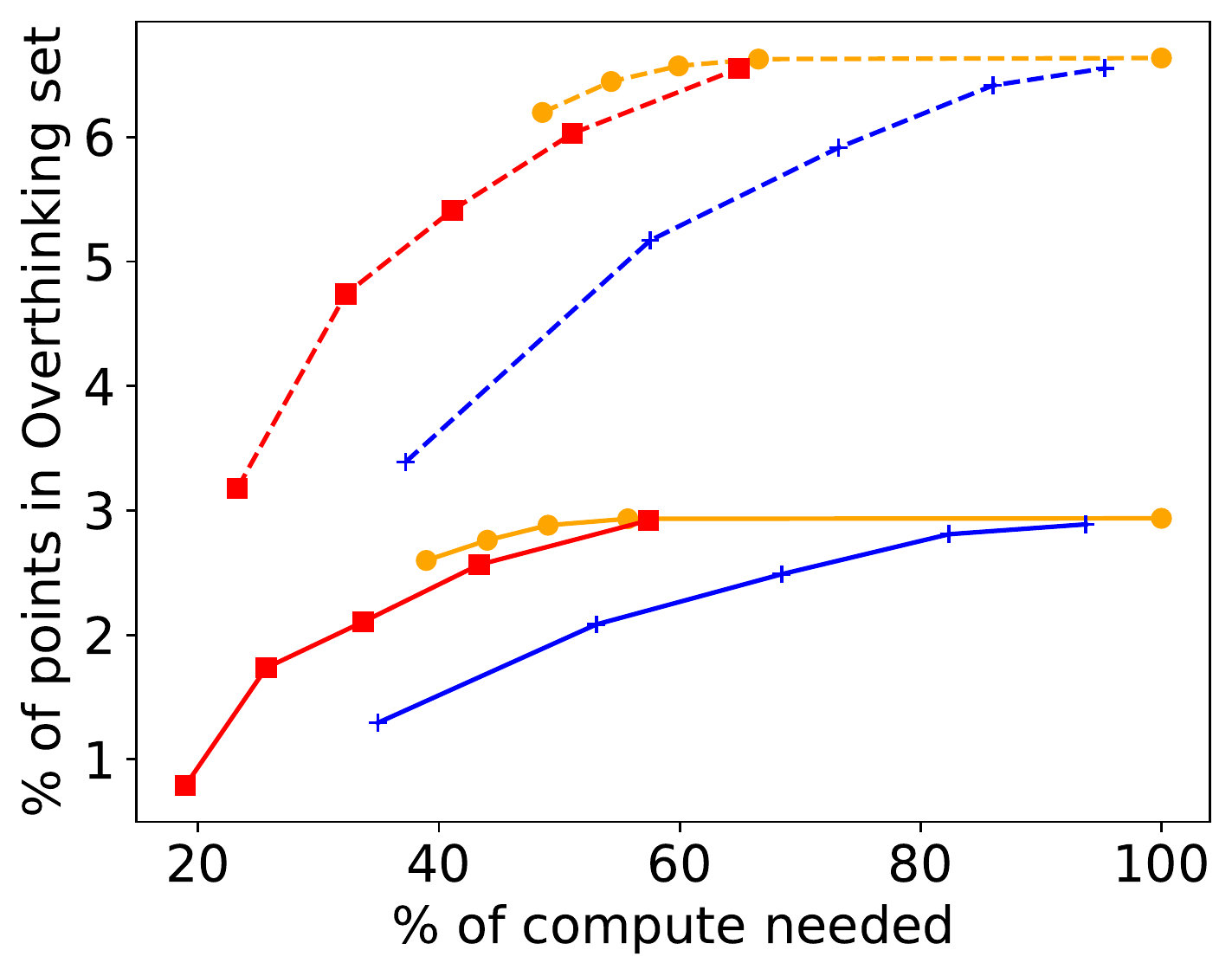}}
  \subfigure[ResNet on CIFAR-10]{\includegraphics[width=0.24\columnwidth]{Images/overthinking_vs_efficiency_resnet_cifar10_augmixTrue.pdf}}
  \subfigure[VGG on CIFAR-100]{\includegraphics[width=0.24\columnwidth]{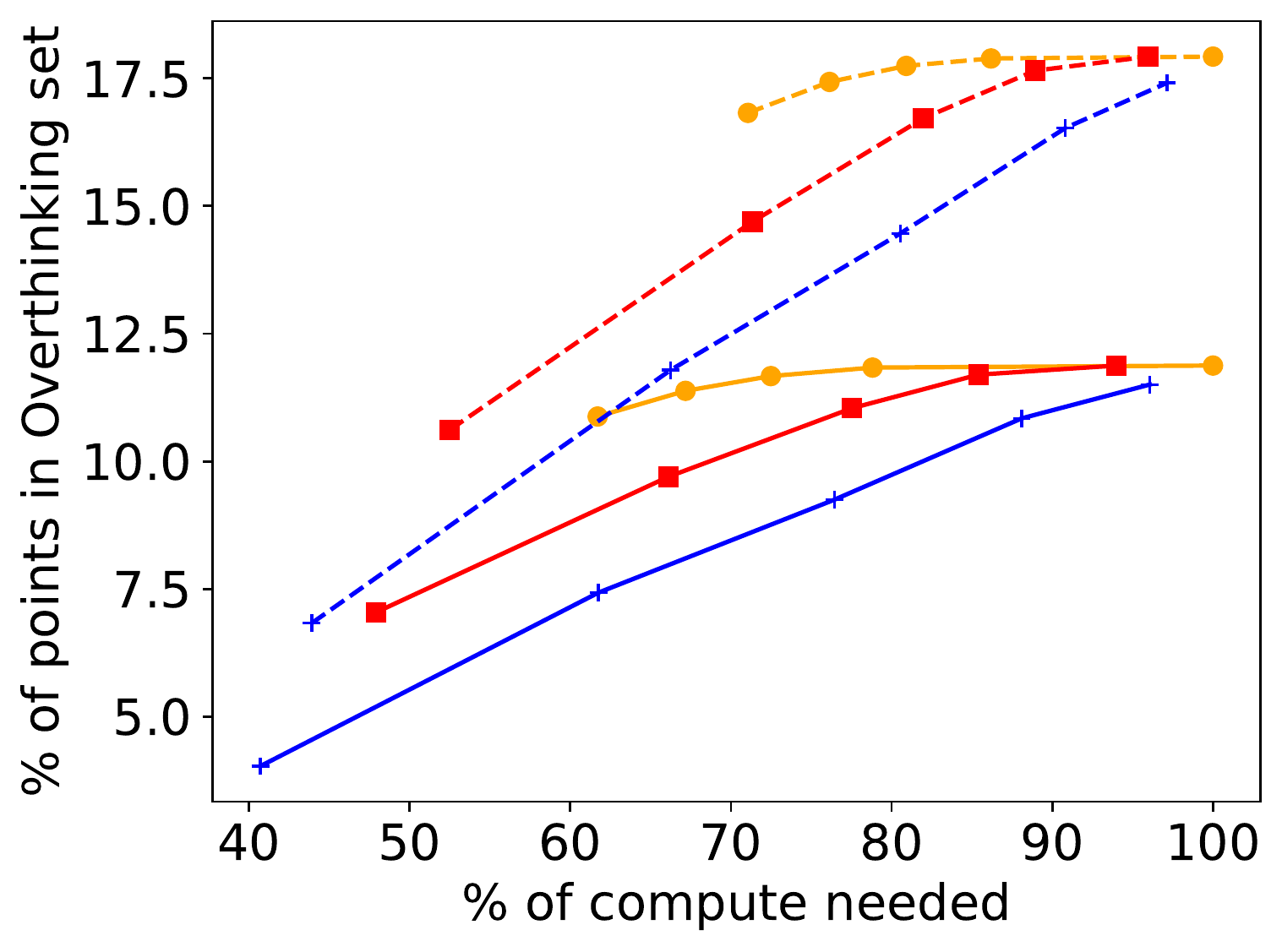}}
  \subfigure[ResNet on CIFAR-100]{\includegraphics[width=0.24\columnwidth]{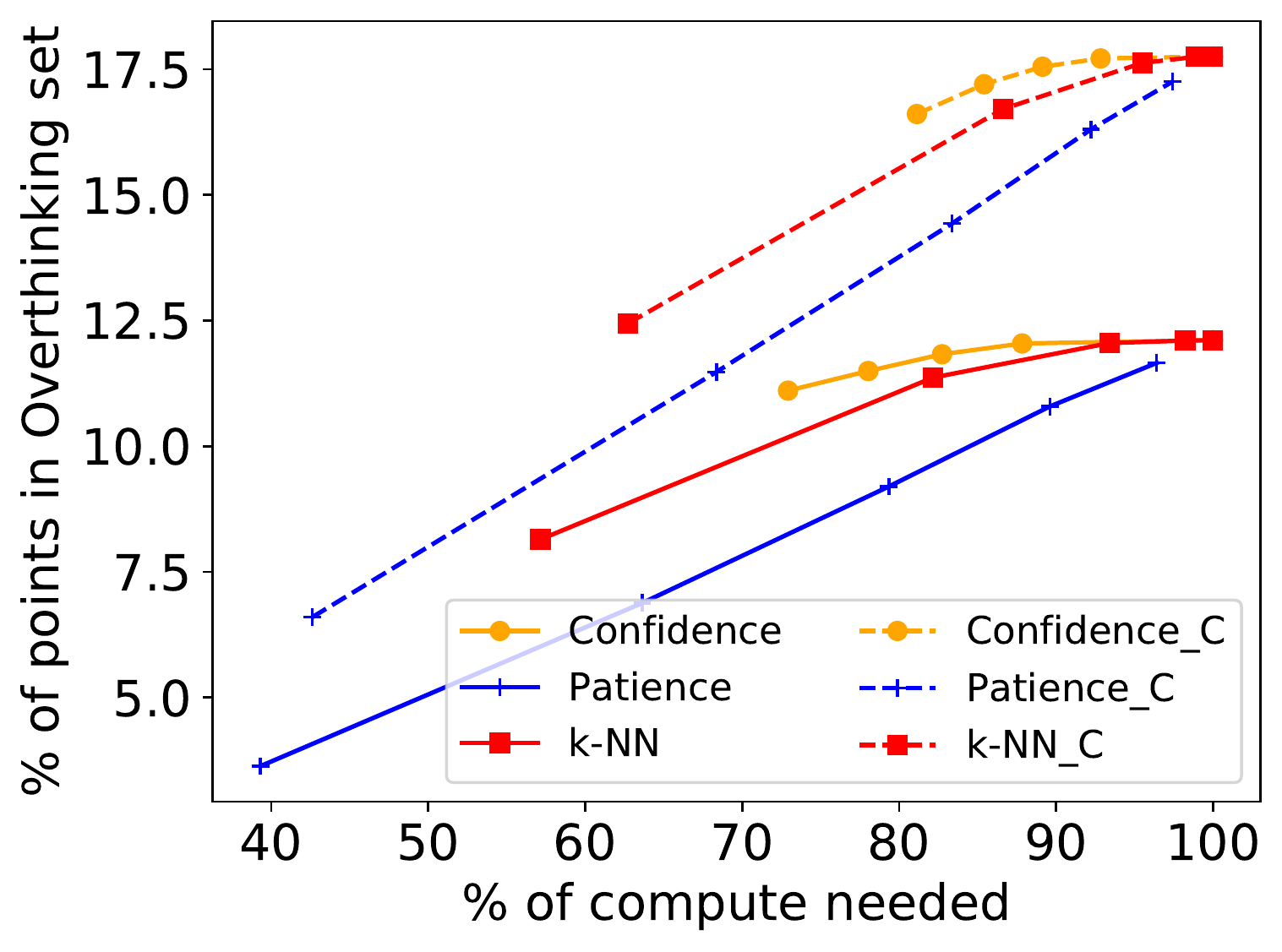}}
  }
  \caption{Decrease in the size of the overthinking set $OT$ of MEMs when trained with AugMix with two different architectures (VGG/ResNet-56) using various practical early-exit strategies on clean and corrupted CIFAR-10/100 datasets.
  }
  \label{fig:augmix_overthinking_dist_shift}
\end{figure*}

\begin{figure*}[tb]
  \centering{
  \subfigure[VGG on CIFAR-10]{\includegraphics[width=0.24\columnwidth]{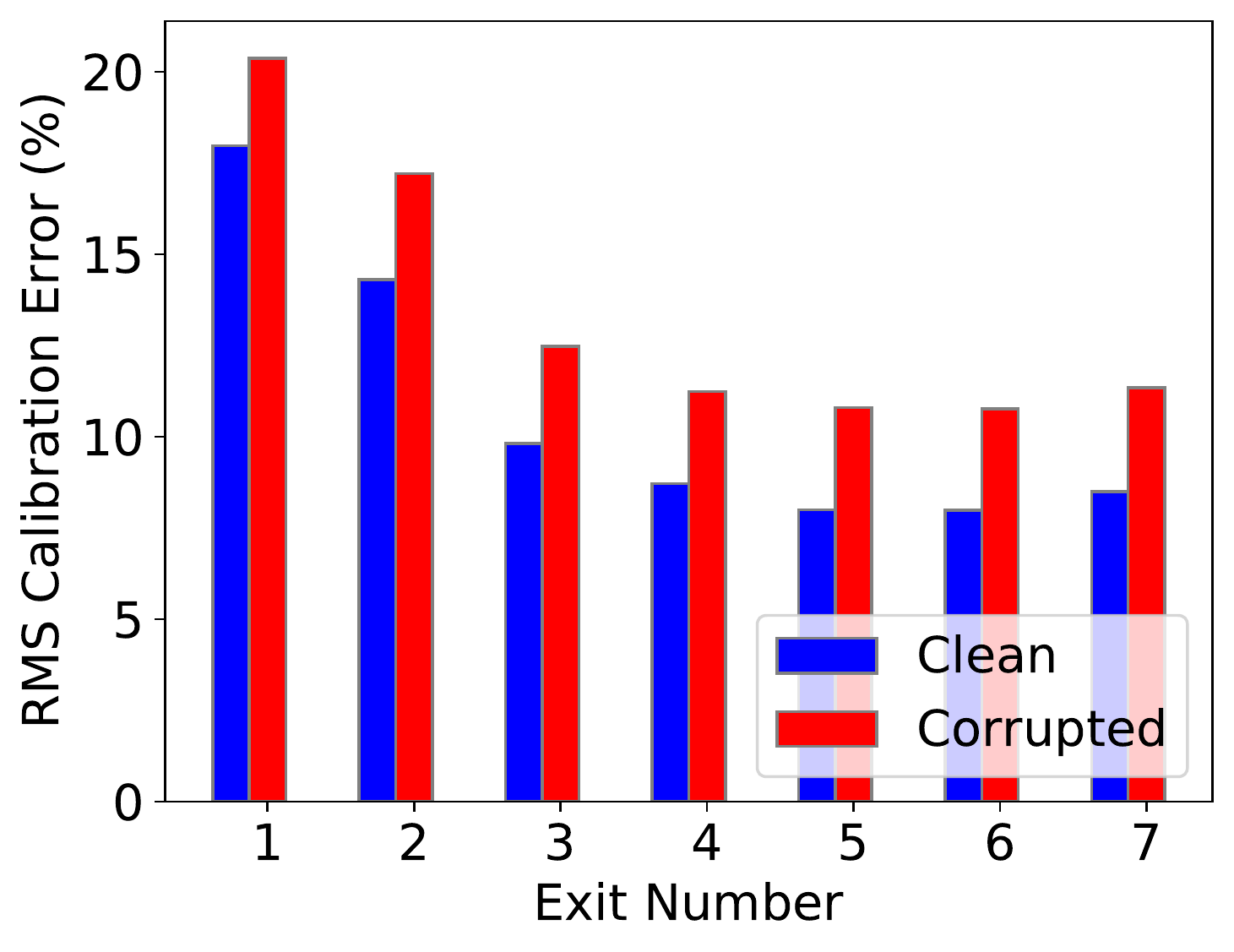}}
  \subfigure[ResNet on CIFAR-10]{\includegraphics[width=0.24\columnwidth]{Images/exit_wise_calibration_resnet_cifar10_augmixTrue.pdf}}
  \subfigure[VGG on CIFAR-100]{\includegraphics[width=0.24\columnwidth]{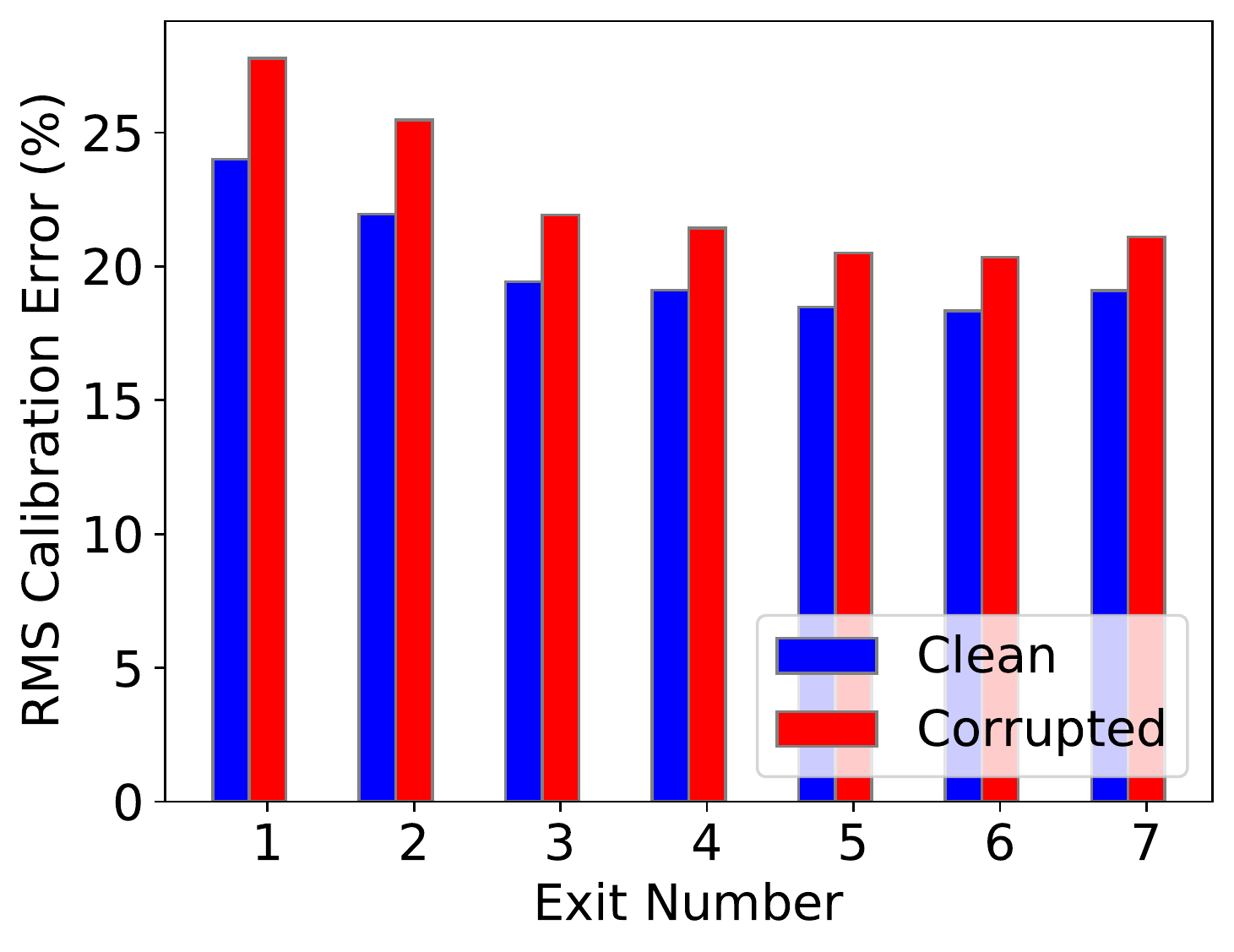}}
  \subfigure[ResNet on CIFAR-100]{\includegraphics[width=0.24\columnwidth]{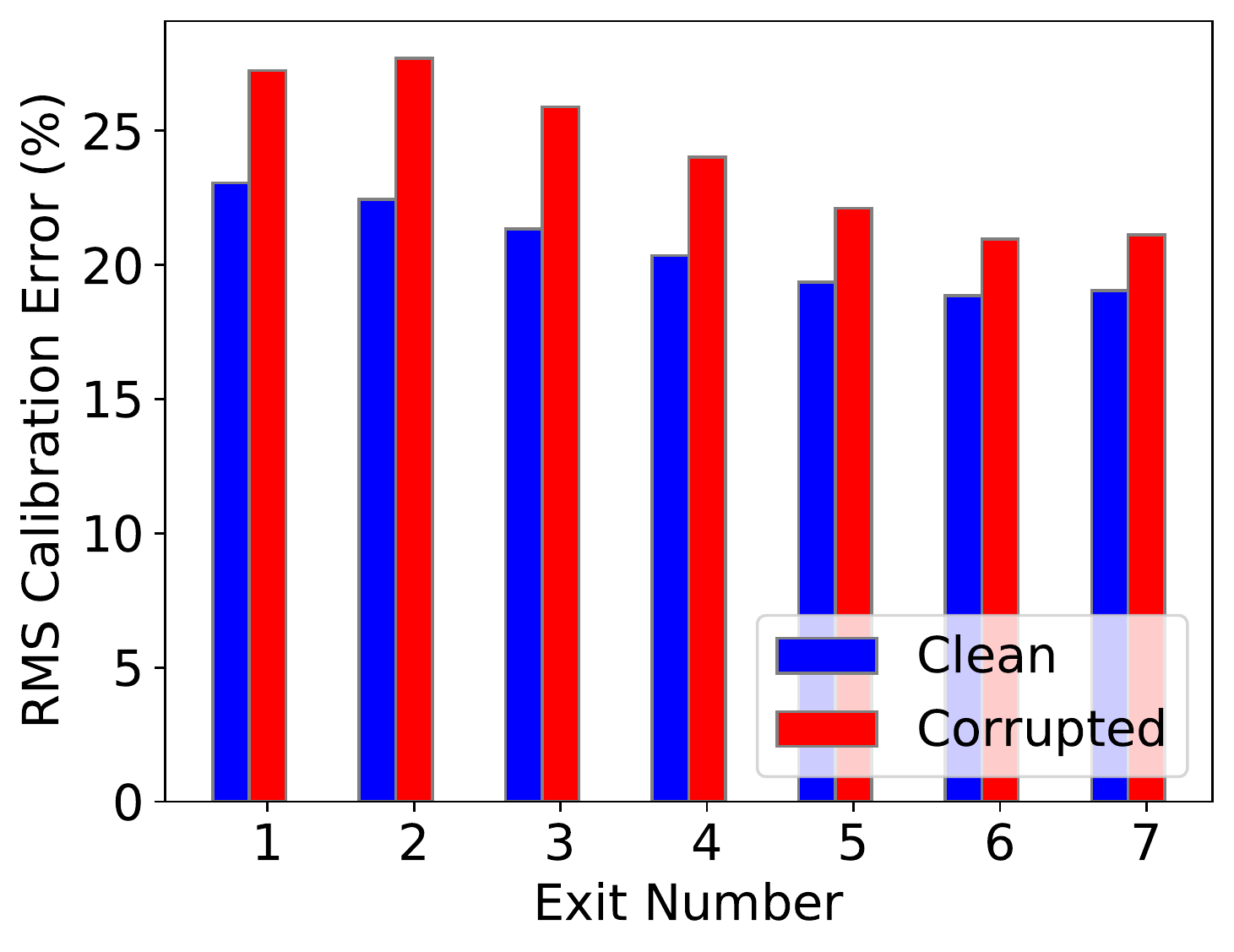}}
  }
  \caption{Improved RMS calibration error of all exits in the MEMs after training with AugMix with two different architectures (VGG/ResNet-56).
  }
  \label{fig:augmix_calibration_dist_shift}
\end{figure*}

\begin{figure*}[tb]
  \centering{
  \subfigure[VGG on CIFAR-10]{\includegraphics[width=0.24\columnwidth]{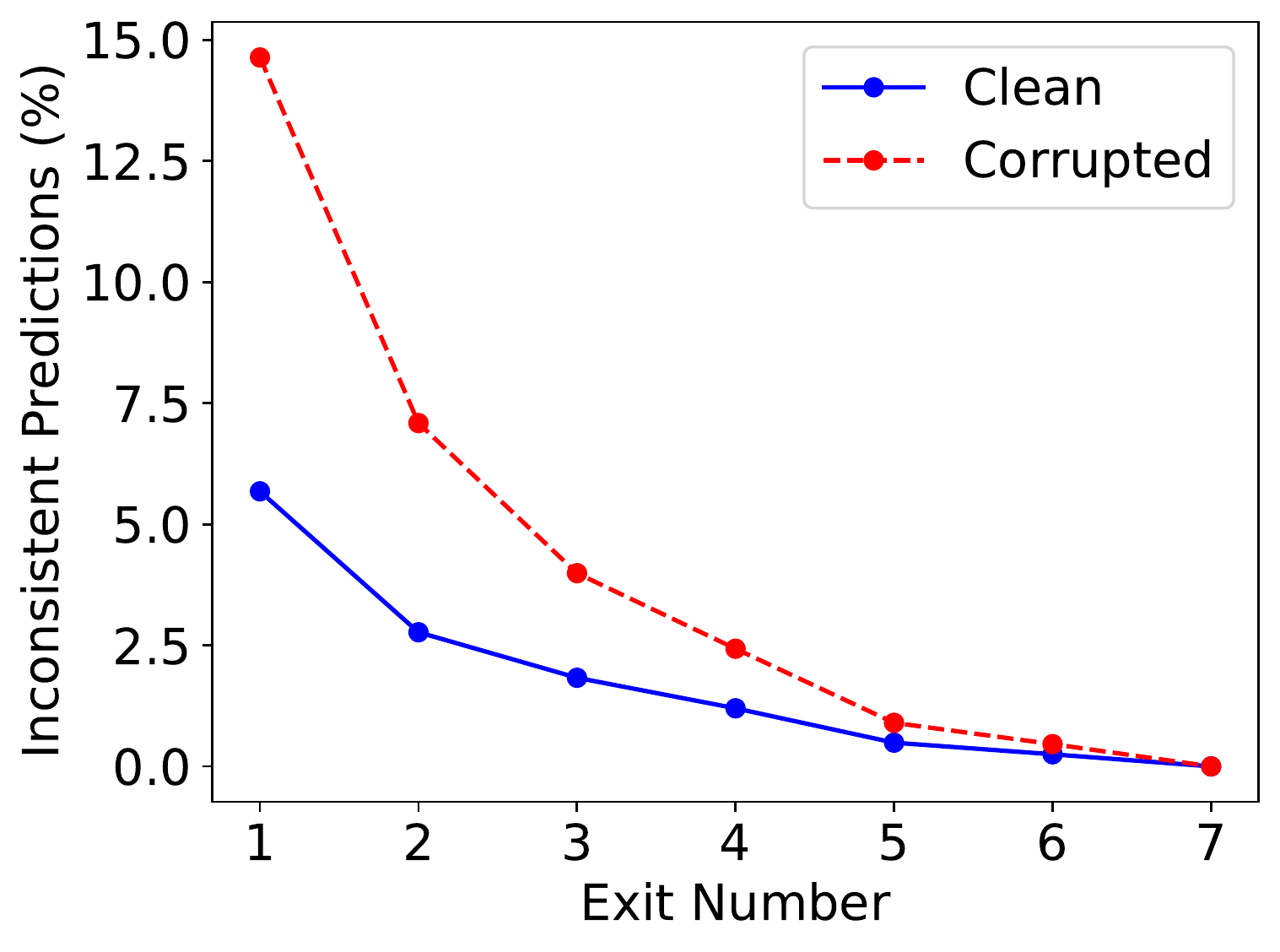}}
  \subfigure[ResNet on CIFAR-10]{\includegraphics[width=0.24\columnwidth]{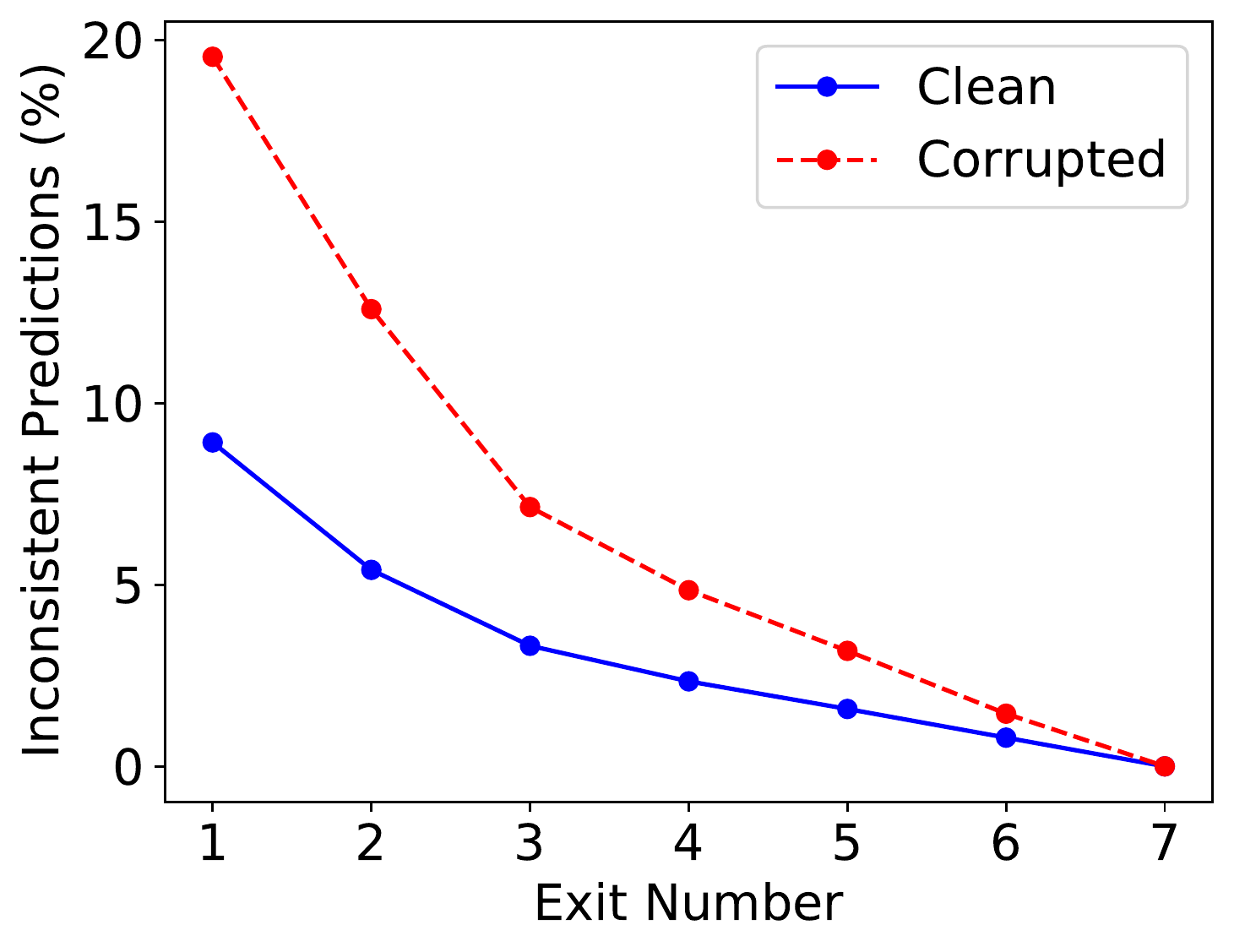}}
  \subfigure[VGG on CIFAR-100]{\includegraphics[width=0.24\columnwidth]{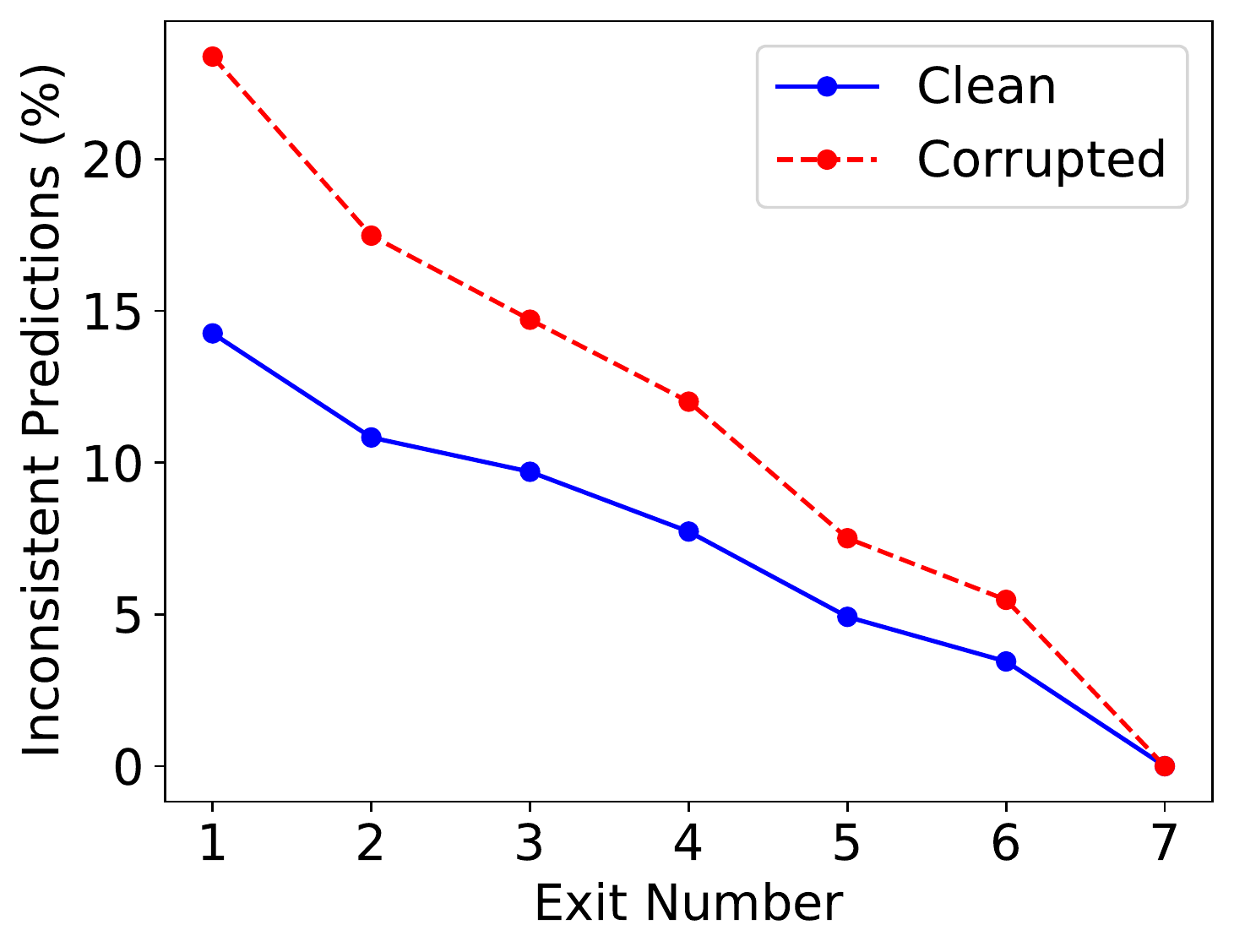}}
  \subfigure[ResNet on CIFAR-100]{\includegraphics[width=0.24\columnwidth]{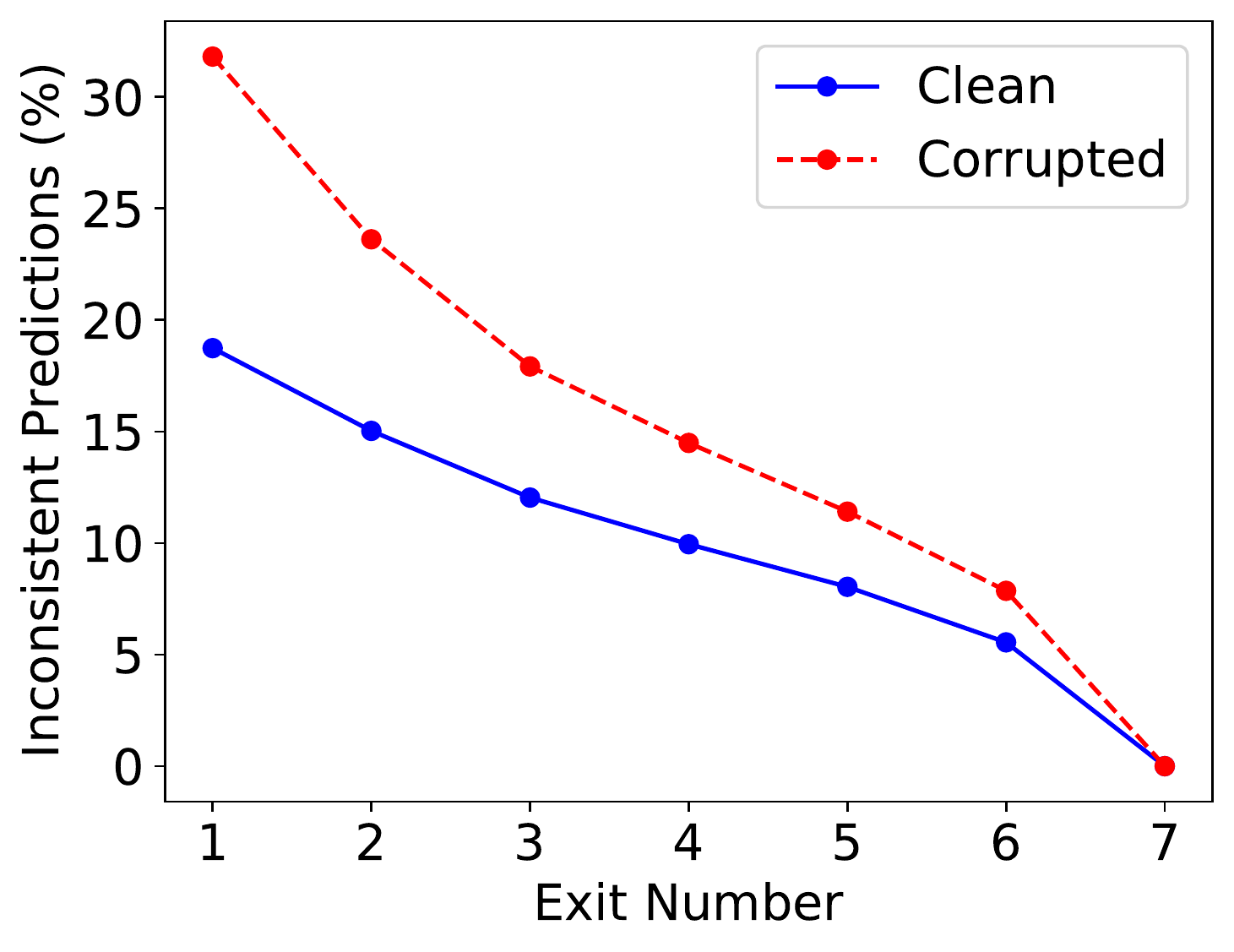}}
  }
  \caption{Decrease in the inconsistency of the predictions of all exits in the MEMs when trained with AugMix with two different architectures (VGG/ResNet-56).
  }
  \label{fig:augmix_inconsistent_predictions_dist_shift}
\end{figure*}

Additionally, we present the results for training the SDN-based MEM using AugMix~\cite{hendrycks2019augmix} which has been shown to improve the accuracy of DNNs on corrupted data. We use AugMix-based loss in every exit and train the backbone and all internal classifiers. This significantly improves the calibration of various exits in the MEM both on clean and corrupted data as shown in Figure~\ref{fig:augmix_calibration_dist_shift} which leads to a decrease in underthinking. Moreover, it also produces models with consistent predictions across the exits leading to a decrease in overthinking as seen in Figure~\ref{fig:augmix_inconsistent_predictions_dist_shift}.
Due to decreased underthinking, Figure~\ref{fig:augmix_underthinking_dist_shift}, and overthinking, Figure~\ref{fig:augmix_overthinking_dist_shift}, the gap between the accuracy (see Figure~\ref{fig:augmix_accuracy_dist_shift}) of the oracle-based and practical early-exit strategies also diminishes. Even though the gap is still not the same as that on in-distribution data, training with AugMix considerably improves the performance of MEMs in presence of distribution shift.

\begin{figure*}[tb]
  \centering{
  \subfigure[VGG on CIFAR-10]{\includegraphics[width=0.24\columnwidth]{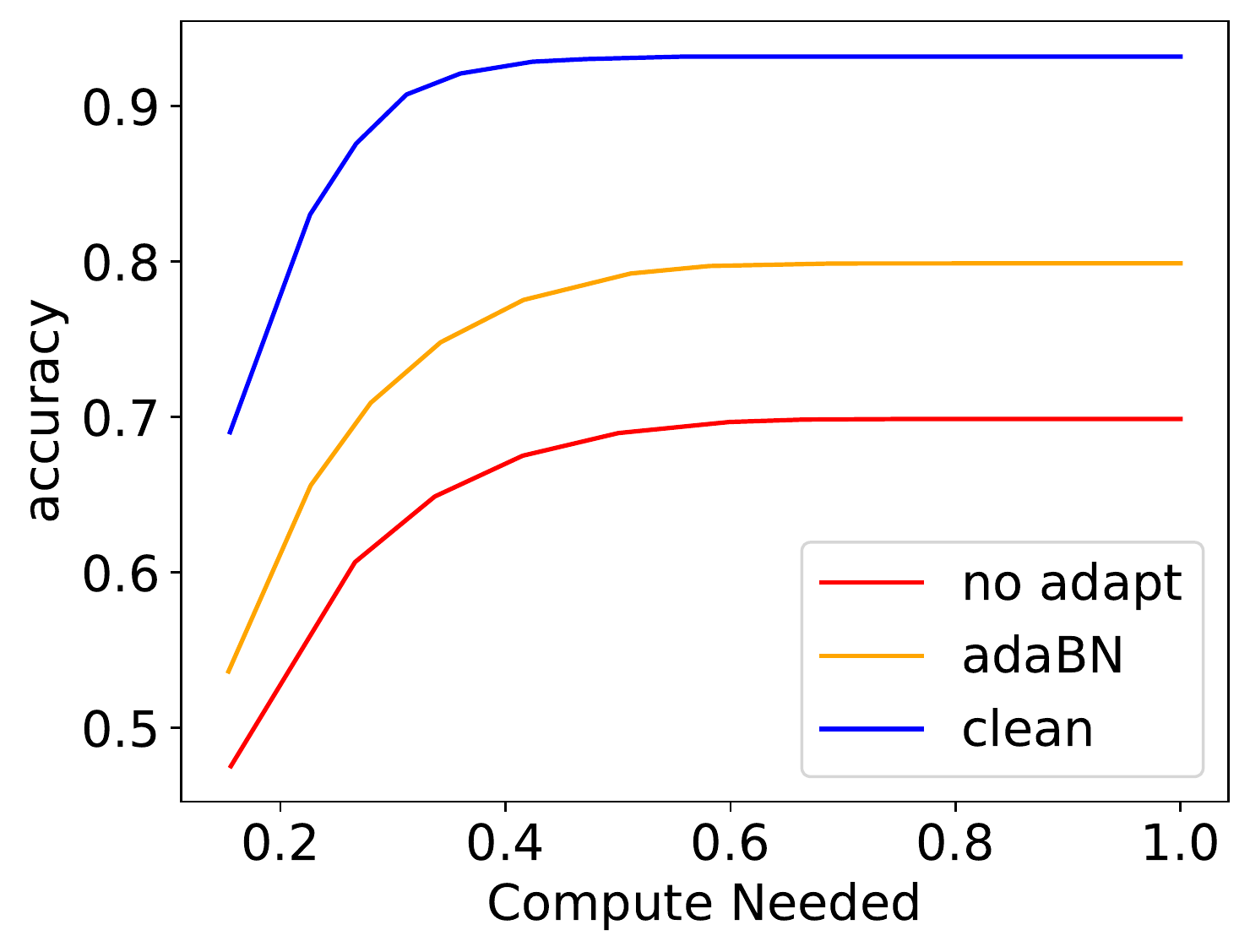}}
  \subfigure[ResNet on CIFAR-10]{\includegraphics[width=0.24\columnwidth]{Images/resnet_cifar10_accuracy_adaBN.pdf}}
  \subfigure[VGG on CIFAR-100]{\includegraphics[width=0.24\columnwidth]{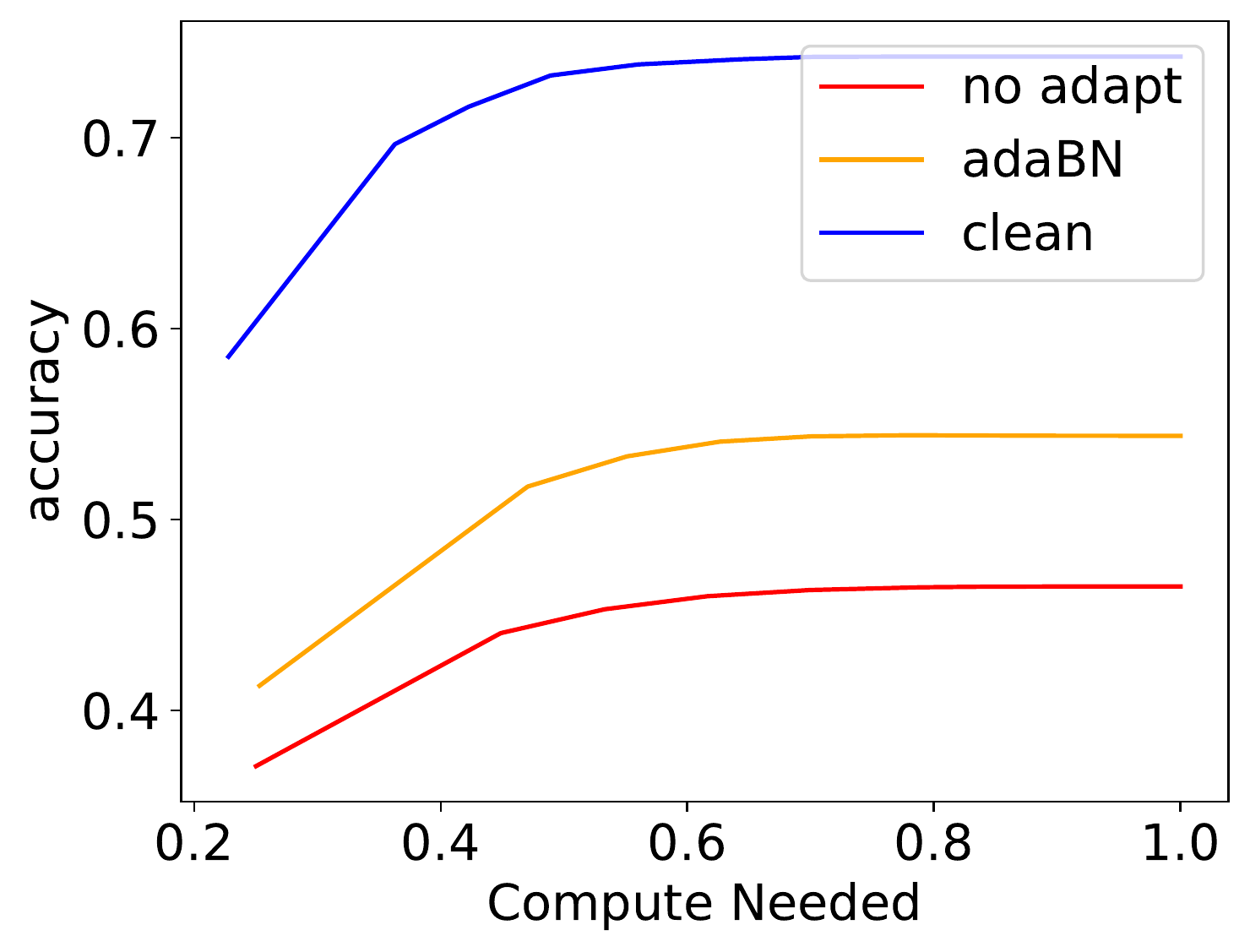}}
  \subfigure[ResNet on CIFAR-100]{\includegraphics[width=0.24\columnwidth]{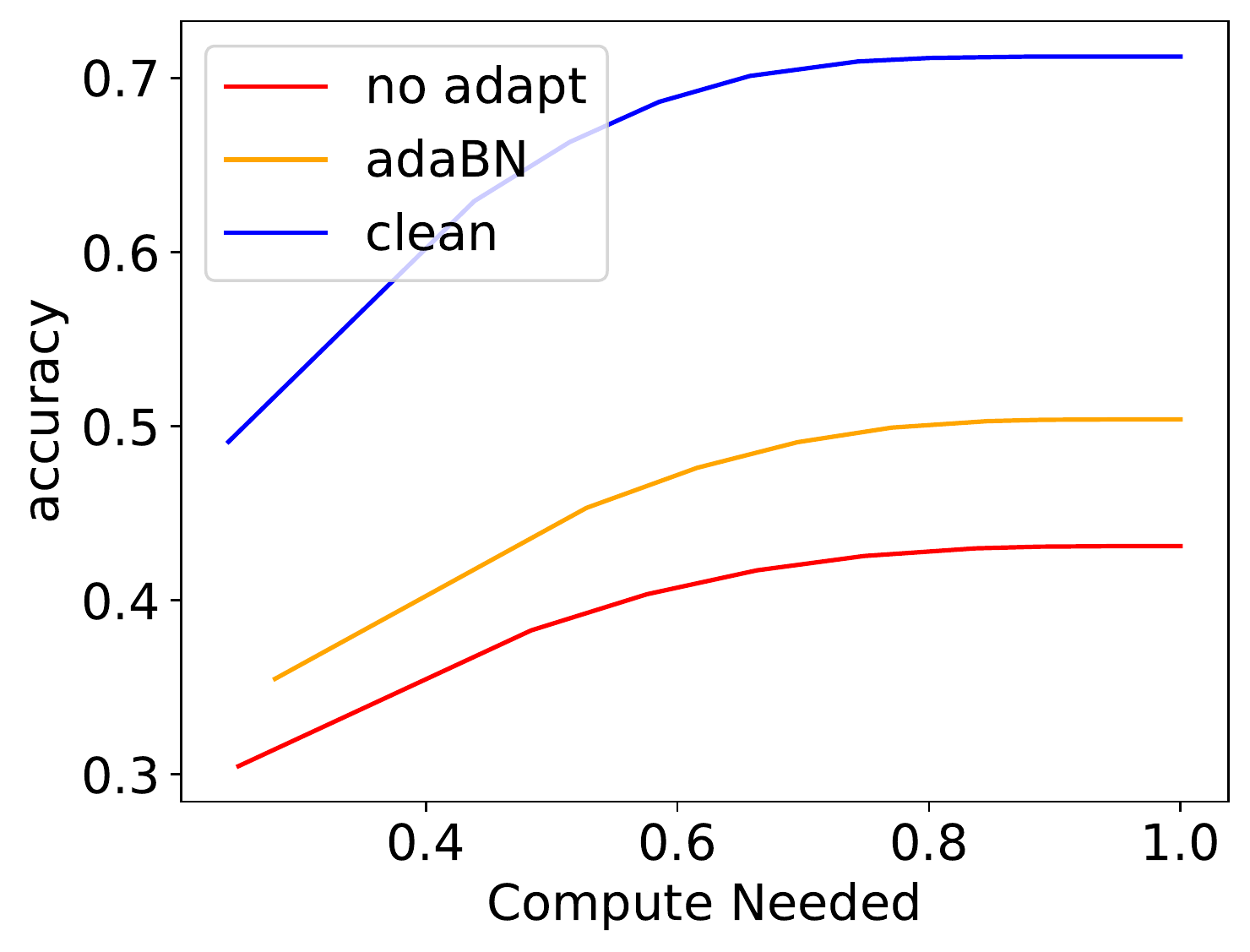}}
  }
  \caption{Comparison of accuracy of MEMs with adapted BN layers evaluated over corrupted data with a MEM evaluated on clean data and a MEM evaluated on corruptions without adaptation. MEMs use two different architectures (VGG/ResNet-56) and datasets (CIFAR-10/100).
  }
  \label{fig:adabn_accuracy_dist_shift}
\end{figure*}

\begin{figure*}[tb]
  \centering{
  \subfigure[VGG on CIFAR-10]{\includegraphics[width=0.24\columnwidth]{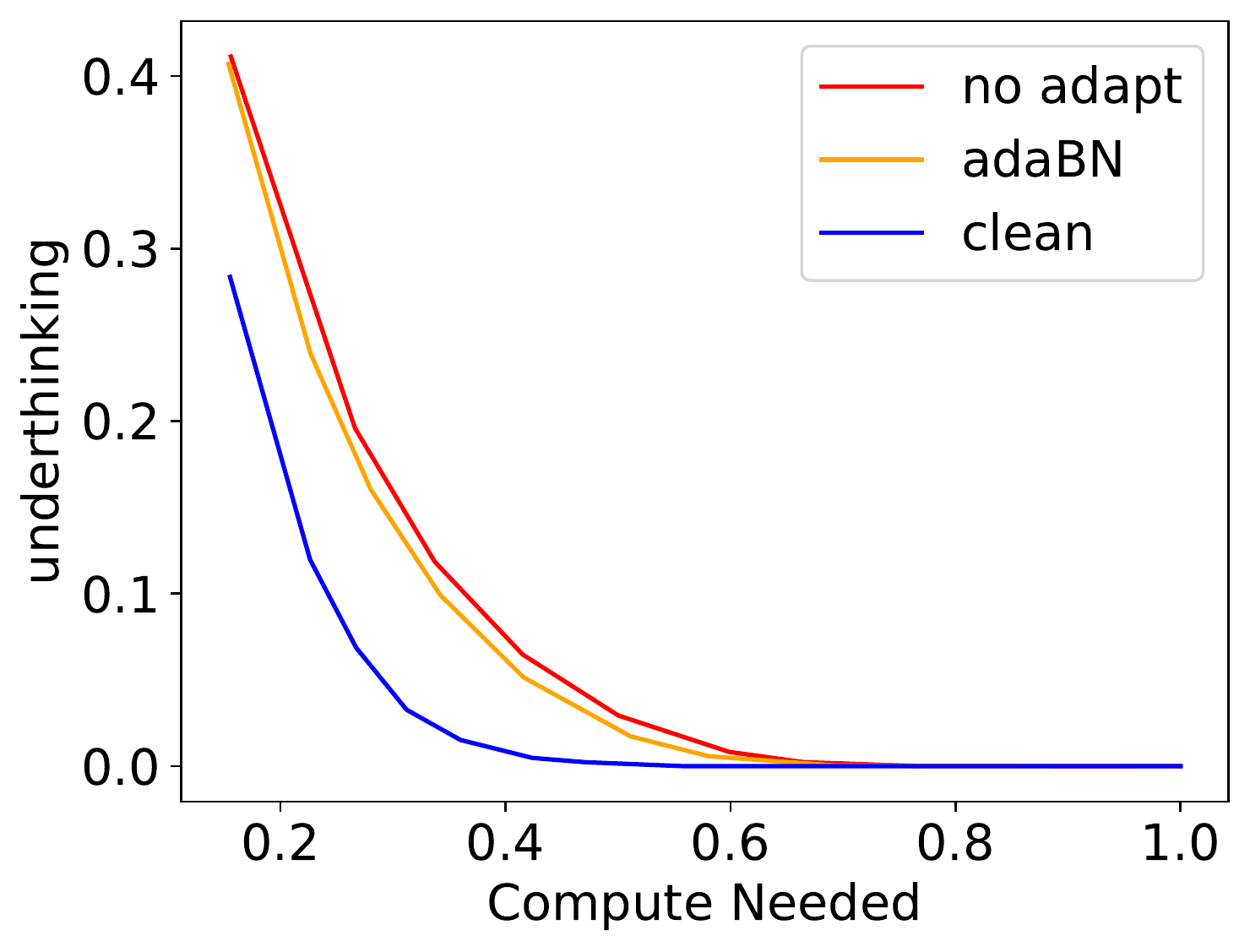}}
  \subfigure[ResNet on CIFAR-10]{\includegraphics[width=0.24\columnwidth]{Images/resnet_cifar10_underthinking_adaBN.pdf}}
  \subfigure[VGG on CIFAR-100]{\includegraphics[width=0.24\columnwidth]{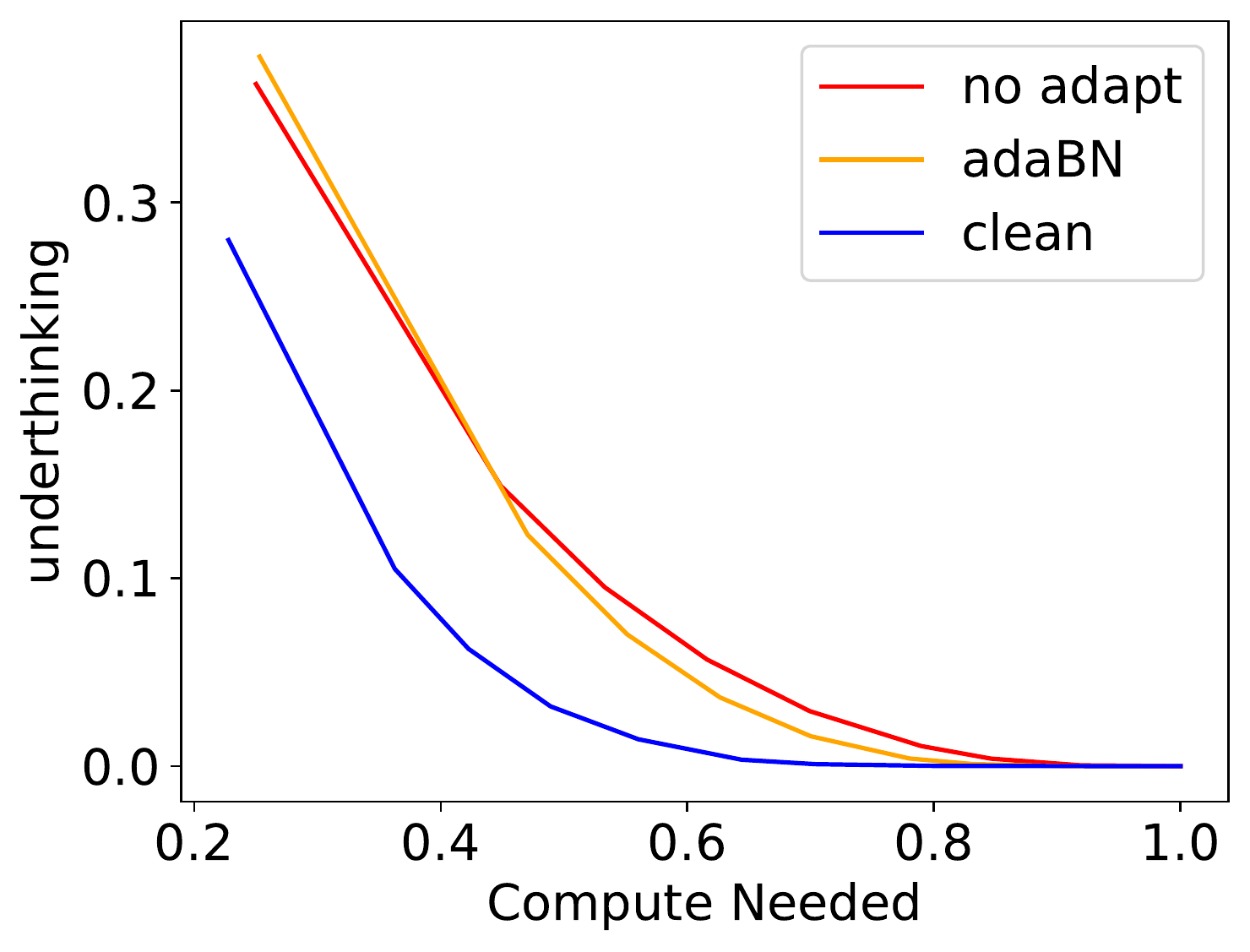}}
  \subfigure[ResNet on CIFAR-100]{\includegraphics[width=0.24\columnwidth]{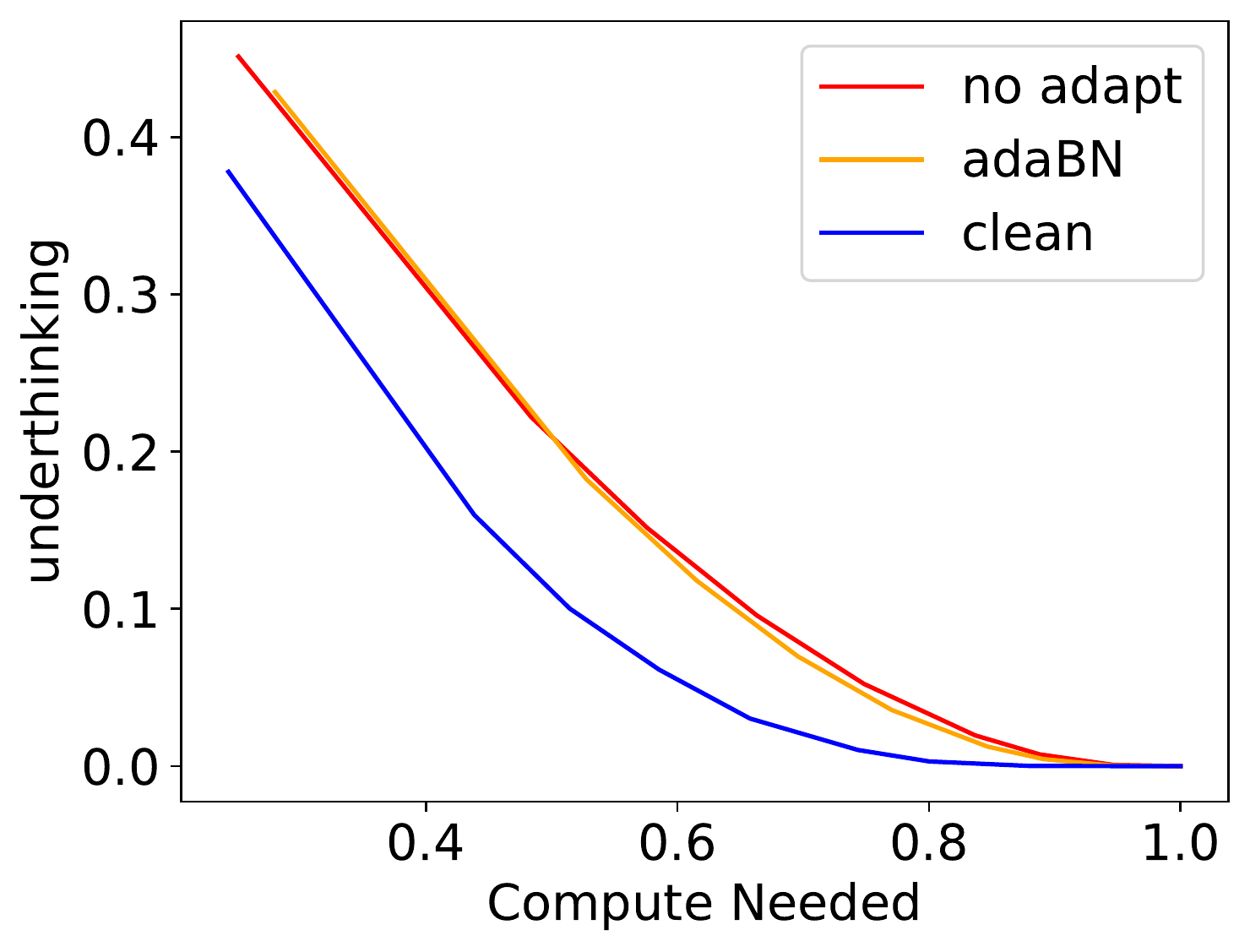}}
  }
  \caption{Comparison of underthinking of MEMs with adapted BN layers evaluated over corrupted data with a MEM evaluated on clean data and a MEM evaluated on corruptions without adaptation. MEMs use two different architectures (VGG/ResNet-56) and datasets (CIFAR-10/100).
  }
  \label{fig:adabn_underthinking_dist_shift}
\end{figure*}

\begin{figure*}[tb]
  \centering{
  \subfigure[VGG on CIFAR-10]{\includegraphics[width=0.24\columnwidth]{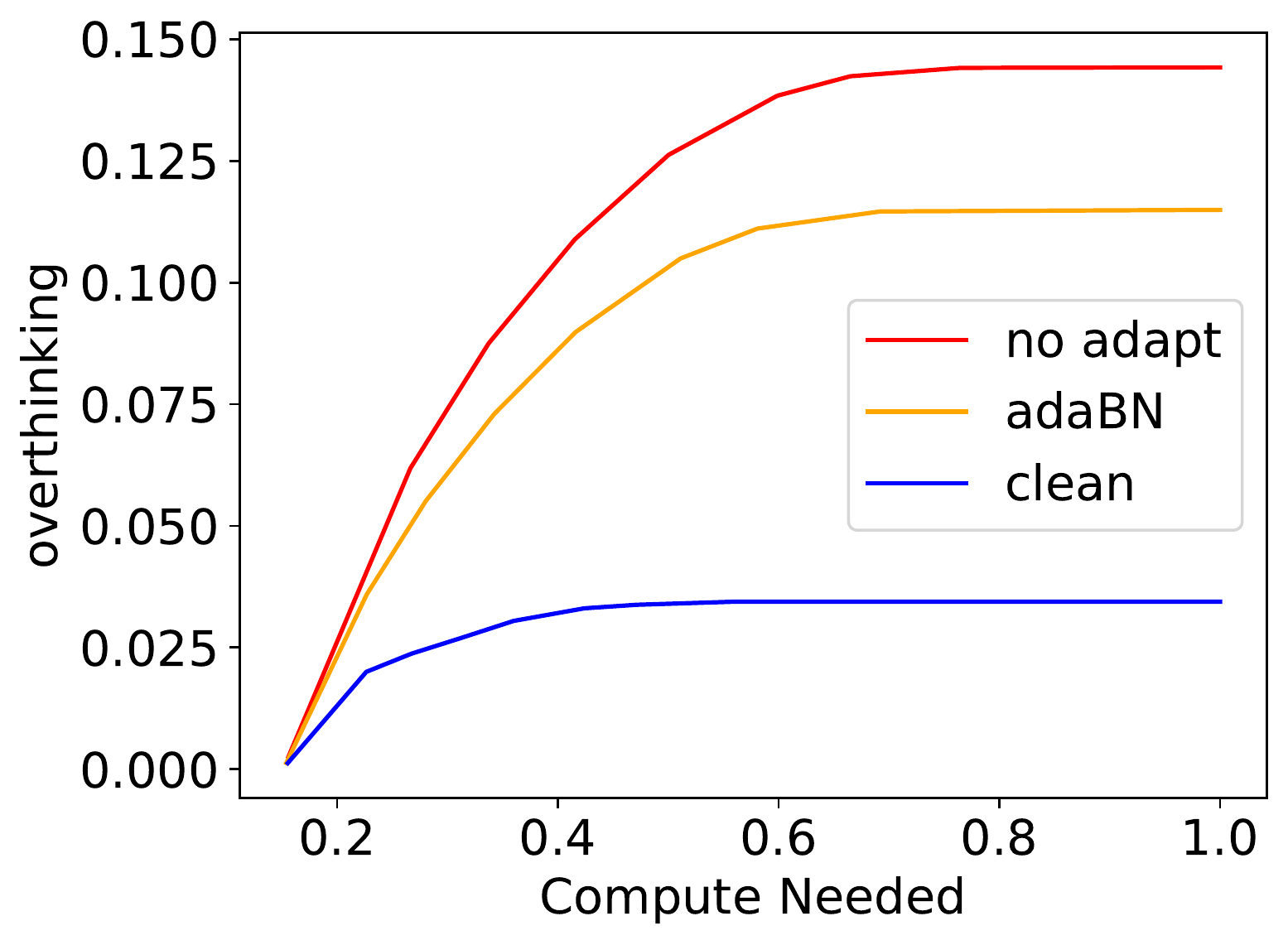}}
  \subfigure[ResNet on CIFAR-10]{\includegraphics[width=0.24\columnwidth]{Images/resnet_cifar10_overthinking_adaBN.pdf}}
  \subfigure[VGG on CIFAR-100]{\includegraphics[width=0.24\columnwidth]{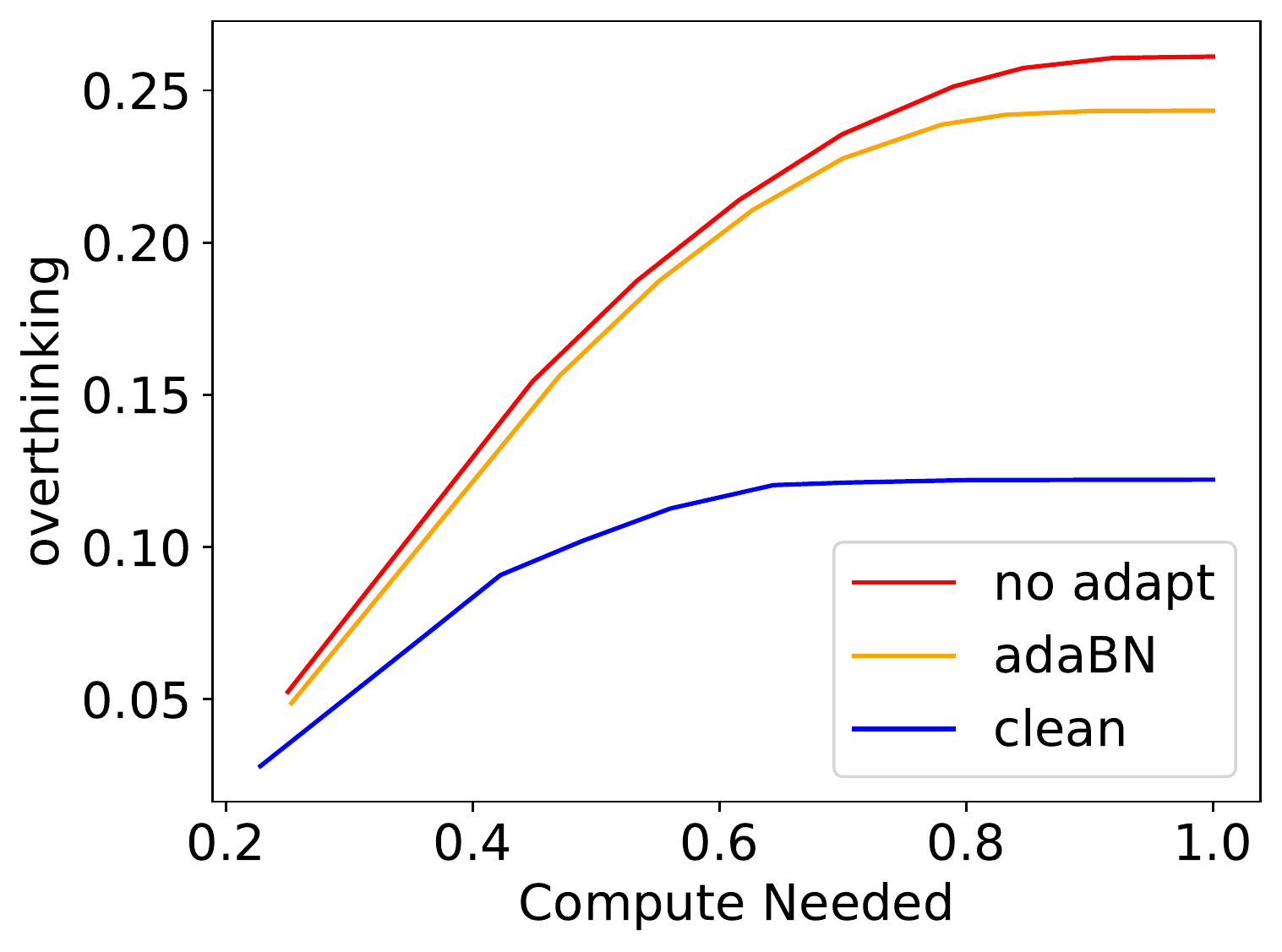}}
  \subfigure[ResNet on CIFAR-100]{\includegraphics[width=0.24\columnwidth]{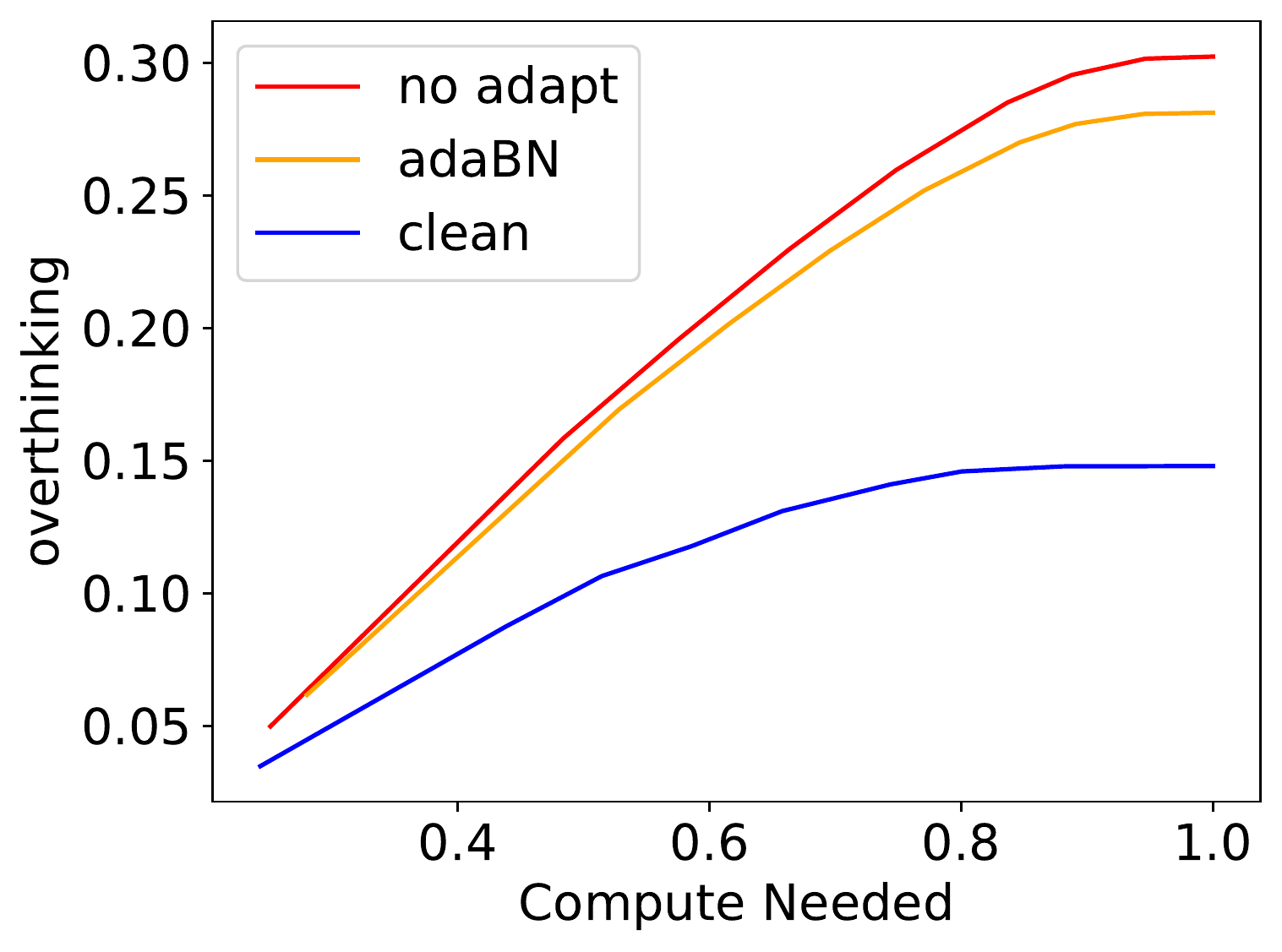}}
  }
  \caption{Comparison of overthinking of MEMs with adapted BN layers evaluated over corrupted data with a MEM evaluated on clean data and a MEM evaluated on corruptions without adaptation. MEMs use two different architectures (VGG/ResNet-56) and datasets (CIFAR-10/100).
  }
  \label{fig:adabn_overthinking_dist_shift}
\end{figure*}

\begin{figure*}[tb]
  \centering{
  \subfigure[VGG on CIFAR-10]{\includegraphics[width=0.24\columnwidth]{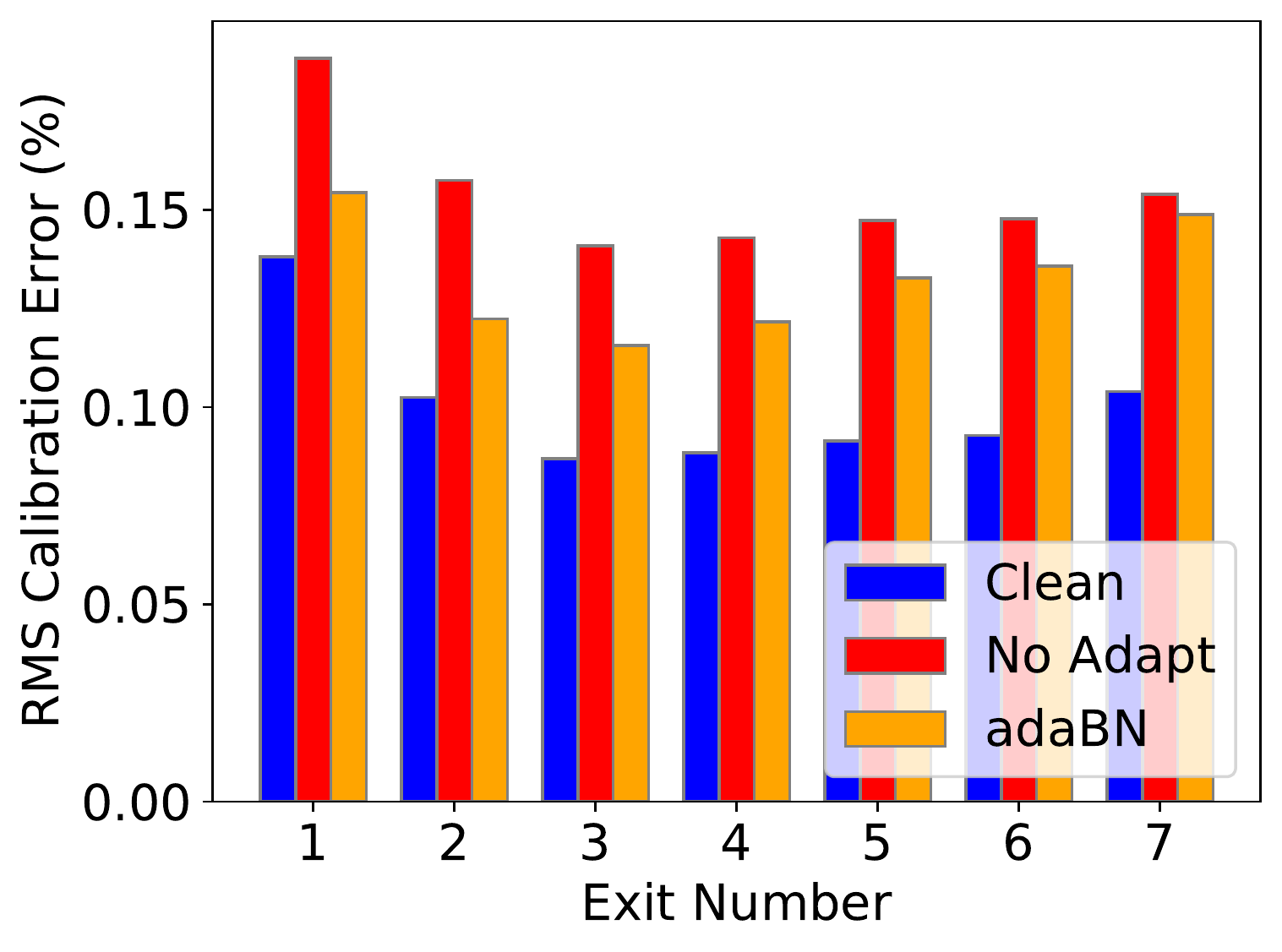}}
  \subfigure[ResNet on CIFAR-10]{\includegraphics[width=0.24\columnwidth]{Images/exit_wise_calibration_resnet_calibrate_False_cifar10_adaBN.pdf}}
  \subfigure[VGG on CIFAR-100]{\includegraphics[width=0.24\columnwidth]{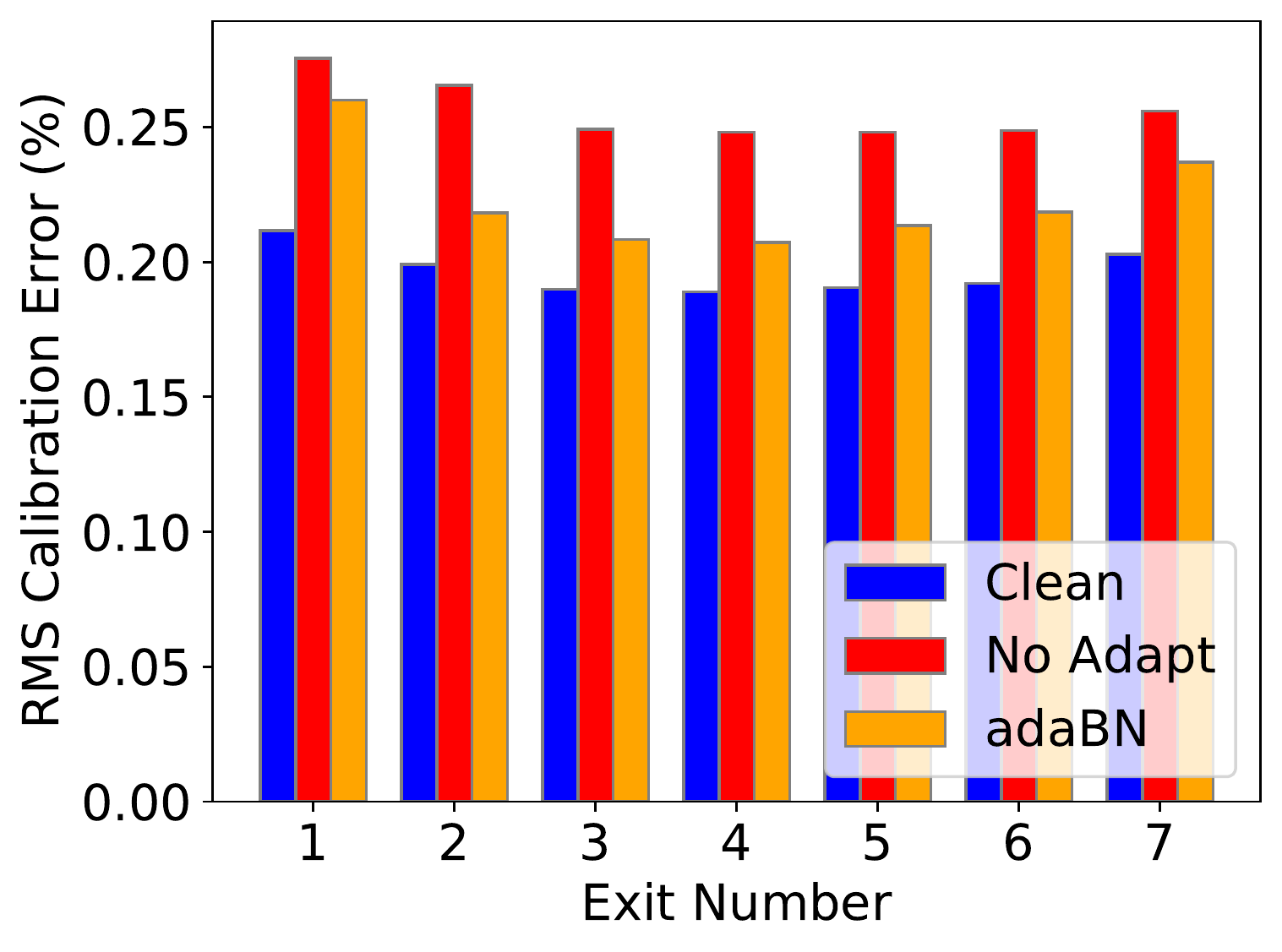}}
  \subfigure[ResNet on CIFAR-100]{\includegraphics[width=0.24\columnwidth]{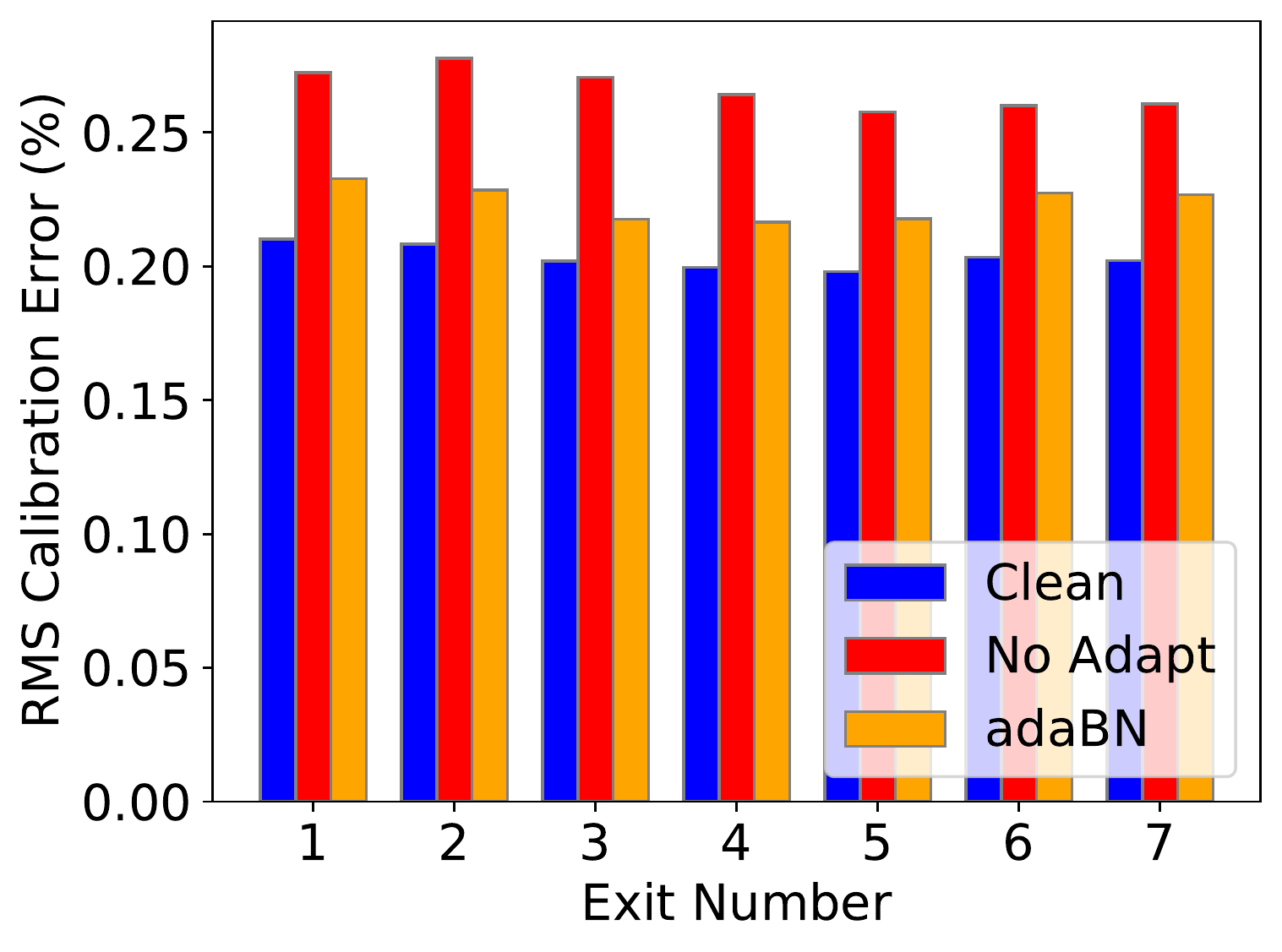}}
  }
  \caption{Improved RMS calibration error of most exits in the MEMs after adapting batch normalization layers with two different architectures (VGG/ResNet-56).
  }
  \label{fig:adabn_calibration_dist_shift}
\end{figure*}

\begin{figure*}[tb]
  \centering{
  \subfigure[VGG on CIFAR-10]{\includegraphics[width=0.24\columnwidth]{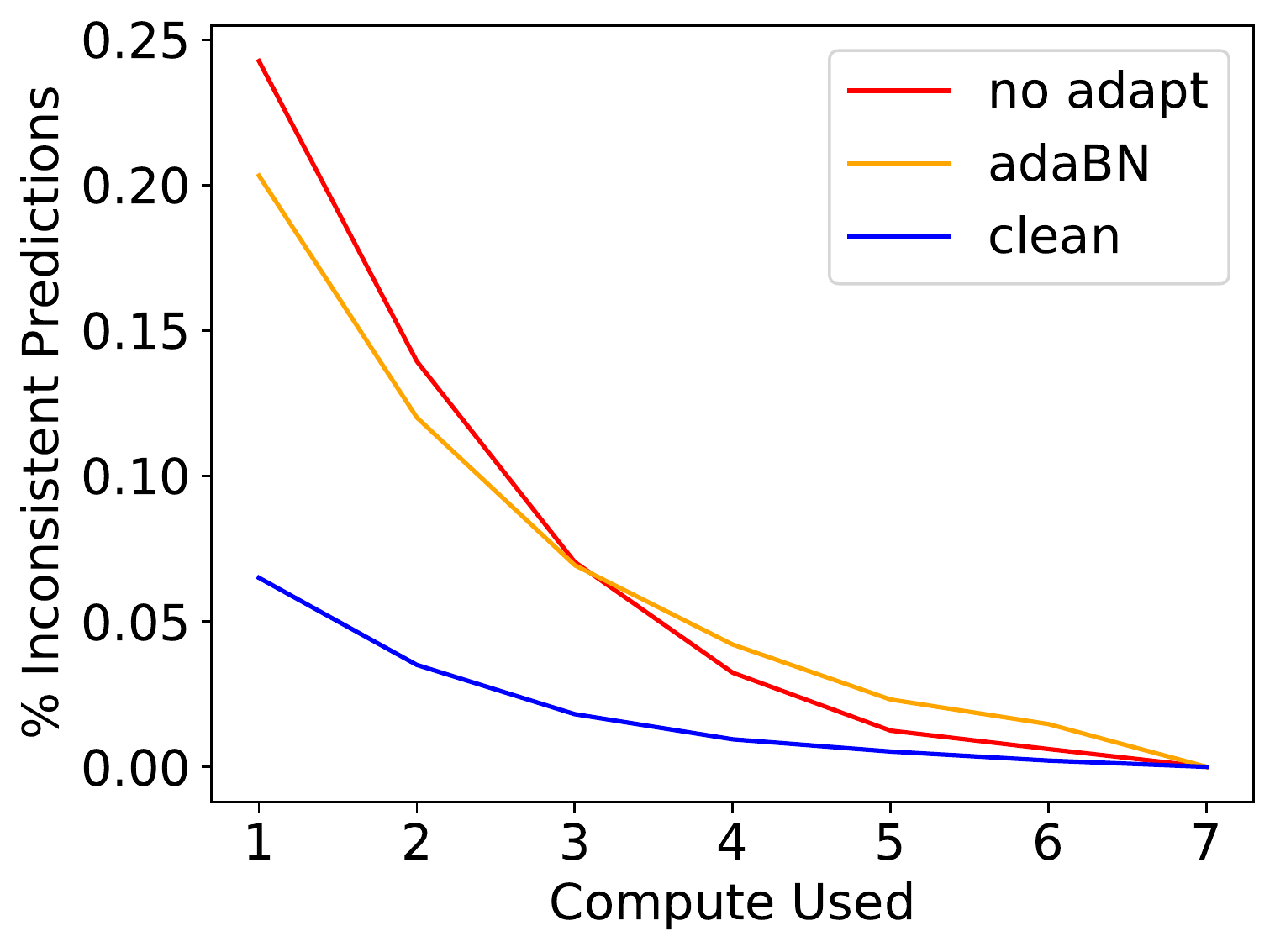}}
  \subfigure[ResNet on CIFAR-10]{\includegraphics[width=0.24\columnwidth]{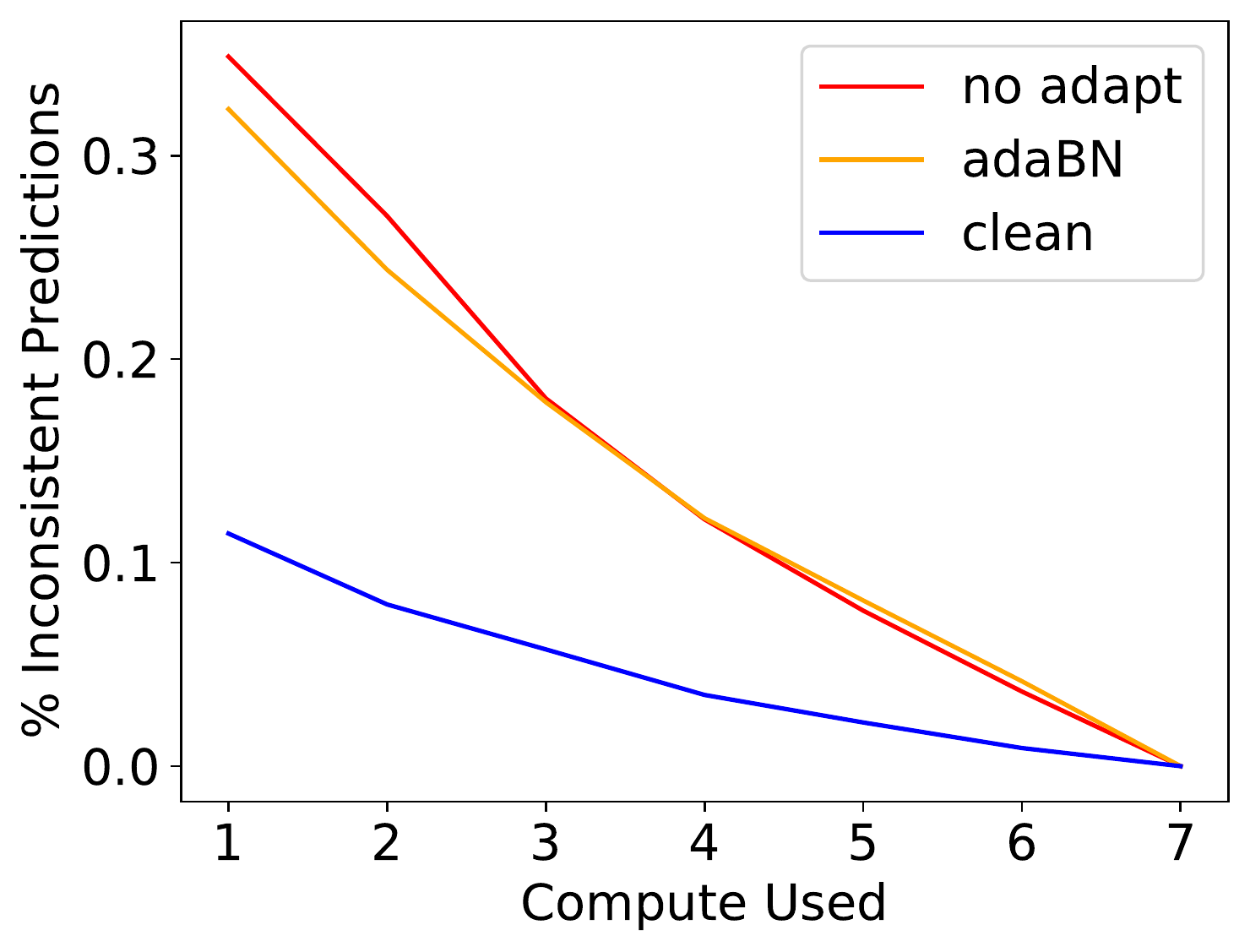}}
  \subfigure[VGG on CIFAR-100]{\includegraphics[width=0.24\columnwidth]{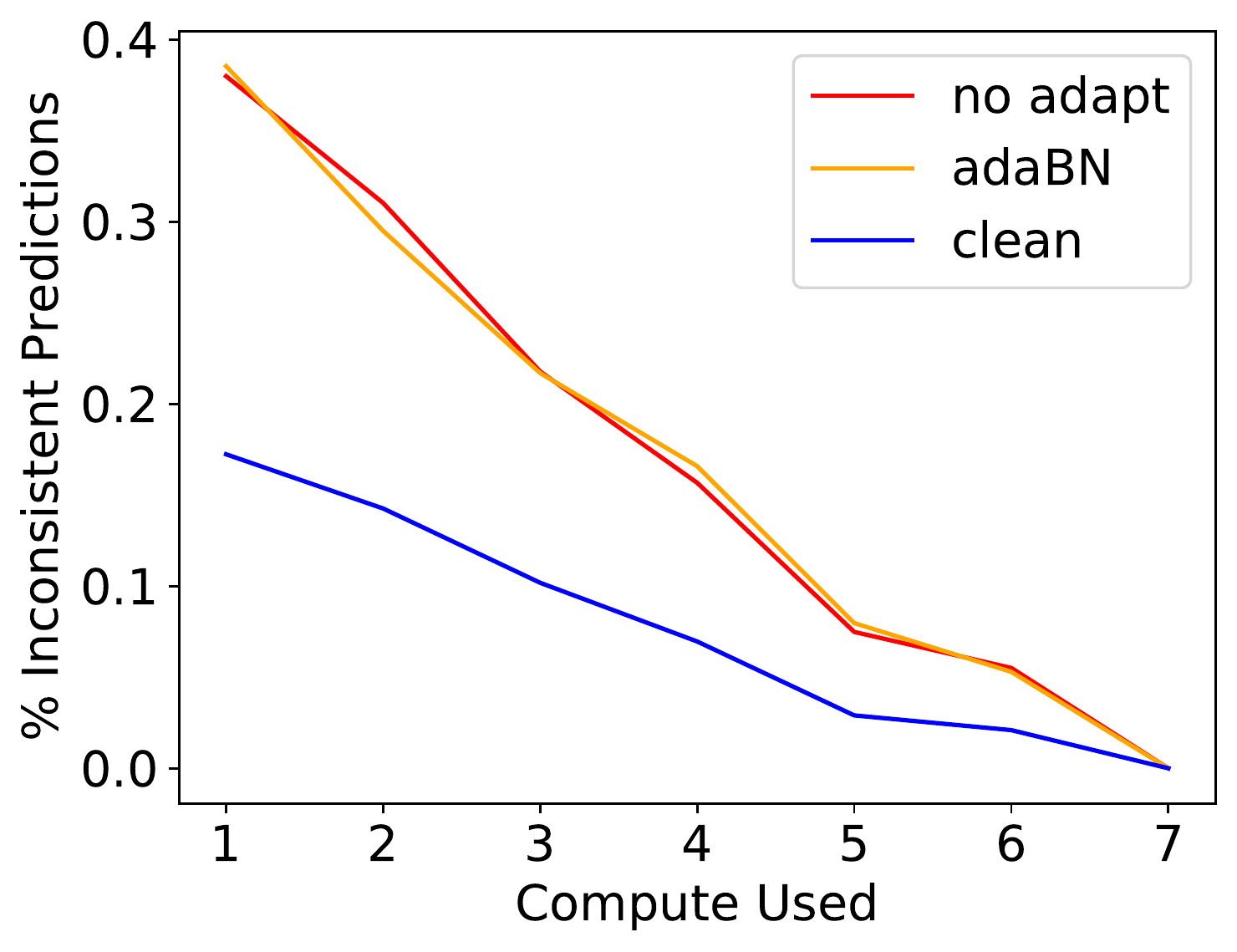}}
  \subfigure[ResNet on CIFAR-100]{\includegraphics[width=0.24\columnwidth]{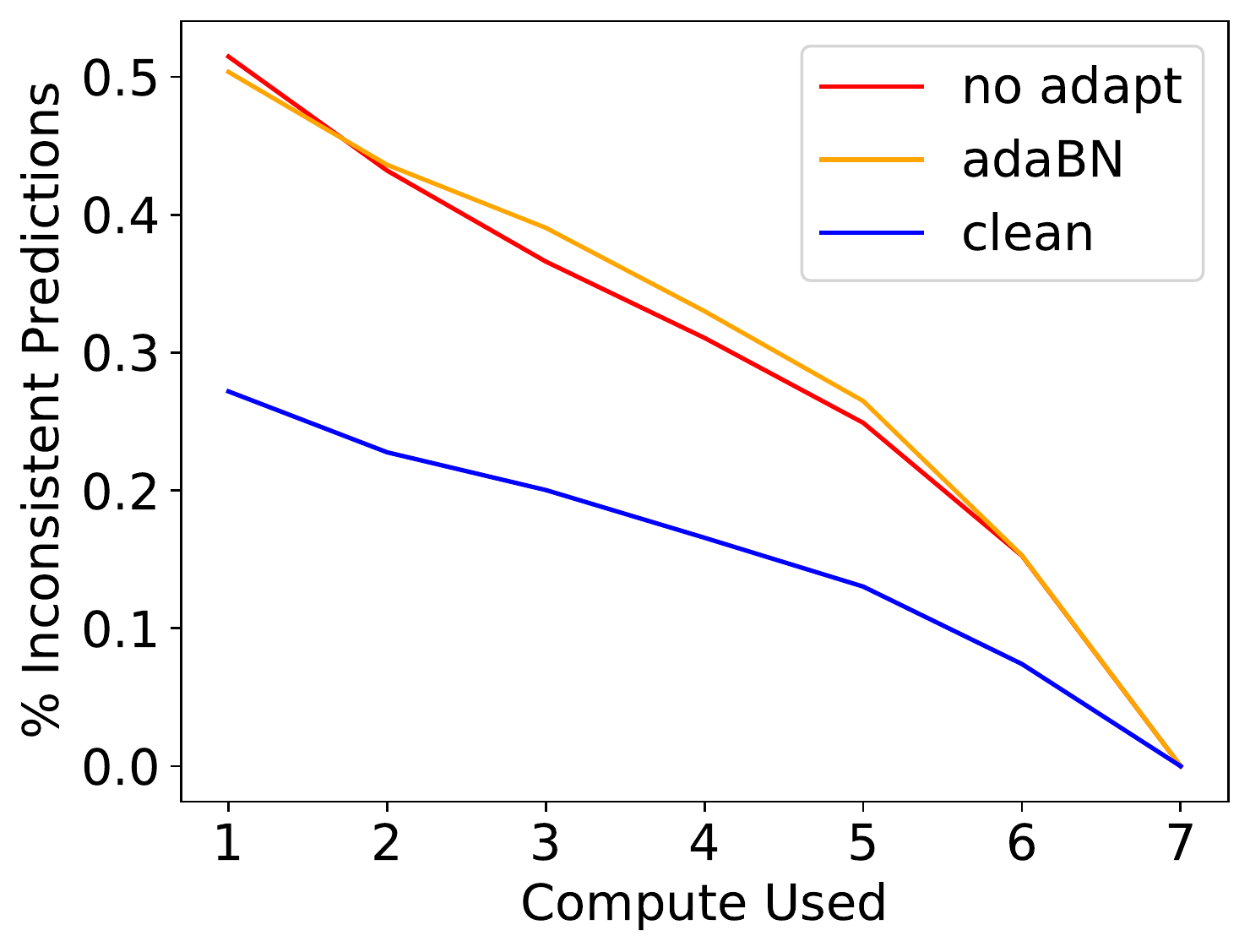}}
  }
  \caption{Decrease in the inconsistency of the predictions of all exits in the MEMs when trained with AugMix with two different architectures (VGG/ResNet-56).
  }
  \label{fig:adabn_inconsistent_predictions_dist_shift}
\end{figure*}

Finally, we present results for adapting batch normalization parameters (adaBN) \cite{benz2021revisiting} which has also been shown to improve robustness of DNNs.  Figure~\ref{fig:adabn_accuracy_dist_shift} indicates a signfiicant increase of $\sim 10\%$ over baseline evaluation on corrupt data, but performance is still short of clean data.  In contrast to AugMix evaluation, however, we note a subtle increase in underthinking (Figure~\ref{fig:adabn_underthinking_dist_shift}) and no difference in overthinking (Figure~\ref{fig:adabn_overthinking_dist_shift}).  Figure~\ref{fig:adabn_calibration_dist_shift} shows RMS calibration error has decreased slightly (especially for CIFAR-10), but overall there is little decrease in inconssitent predictions using adaBN (Figure~\ref{fig:adabn_inconsistent_predictions_dist_shift}).

\section{Experimental details}
\label{app:experimental_details}

All codes are written in Python using Pytorch. Dataset details, model architectures, and training hyperparameters used are described below.

\textbf{Dataset Details}: All models are trained with the CIFAR-10 or CIFAR-100 dataset.  We use the standard 50,000 training and 10,000 train/test split, and additionally create a  validation set using a random 5,000 samples from the training set for selecting exit thresholds.  To train all architectures, a standard data augmentation scheme using random cropping, random horizontal flip, and normalization is used.
For corrupted data, we used the corruptions proposed in the CIFAR-10/100-C dataset \cite{hendrycks2018benchmarking}, namely, \texttt{Gaussian noise, shot noise, impulse noise, glass blur, motion blur, defocus blur, zoom blur, snow, frost, fog, contrast, brightness, elastic transform, pixelate, jpeg compression}.

\textbf{Model Architectures}:  For the VGG and ResNet architectures, we use standard configurations for VGG-16-BN  and ResNet-56 following \cite{kaya2019shallow}.  To convert each network to its SDN variant, we pick the internal layers closest to 15\%, 30\%, 45\%, 60\%, 75\%, and 90\% of the full network's compute cost.  Internal classifiers consist of a mixed max-average pooling and linear classifier following \cite{kaya2019shallow}.

\textbf{Training Hyperparameters}: All DNN and SDN architectures are trained from scratch for 100 epochs using SGD with momentum ($0.9$) starting from an initial learning rate of $0.1$ and decaying by a factor of $0.1$ at epochs $35, 60,$ and  $85.$ Standard cross-entropy loss is used as the training objective.  

\begin{table}
  \caption{Improvement in the accuracy (\%) and reduction in the compute required (CR) (\%) relative to exiting at the last layer
  with SDN-based MEMs (with an oracle-based early-exit strategy) over standard deep neural networks (DNNs) on clean and corrupted (-C) versions of CIFAR-10/100 datasets (mean $\pm$ s.d. of 3 runs).}
  \label{Table:SDN_vs_DNN_orcale}
  \centering
  \small
  \resizebox{\columnwidth}{!}{
  \begin{tabular}{c|cc|cc|cc|cc}
    \toprule
   \multirow{3}{*}{Dataset}  & \multicolumn{4}{c|}{VGG-16} & \multicolumn{4}{c}{ResNet-56} \\
   & \multicolumn{2}{c|}{SDN} & \multicolumn{2}{c|}{DNN} &  \multicolumn{2}{c|}{SDN}& \multicolumn{2}{c}{DNN} \\
   & Accuracy & CR & Accuracy & CR & Accuracy & CR & Accuracy & CR \\
   \midrule
    C-10 & 96.44 $\pm$ 0.10 & 23.37 $\pm$ 0.13 &	93.19 $\pm$ 0.15 & 100.0 &	96.34 $\pm$ 0.14 & 25.95 $\pm$ 0.21	& 91.23 $\pm$ 0.15 & 100.0 \\
    C-10-C & 80.32 $\pm$ 0.51 & 39.41 $\pm$ 0.42 &	70.39 $\pm$ 0.25 & 100.0 &	80.14 $\pm$ 0.79 & 43.79 $\pm$ 0.74	& 66.70 $\pm$ 0.64 & 100.0 \\
    \midrule
    C-100 & 84.34 $\pm$ 0.15 & 37.98 $\pm$ 0.05	& 72.74 $\pm$ 0.39 & 100.0 &	83.62 $\pm$ 0.03 & 44.59 $\pm$ 0.20	& 68.91 $\pm$ 0.18 & 100.0 \\
    C-100-C & 61.45 $\pm$ 0.23 & 56.62 $\pm$ 0.14 &	46.43 $\pm$ 0.28 & 100.0 &	60.30 $\pm$ 0.16 & 62.41 $\pm$ 0.28 &	42.24 $\pm$ 0.41 & 100.0 \\
    \bottomrule
  \end{tabular}
  }
\end{table}

\begin{figure*}[tb]
  \centering{
  \subfigure[Exit 1]{\includegraphics[width=0.48\columnwidth]{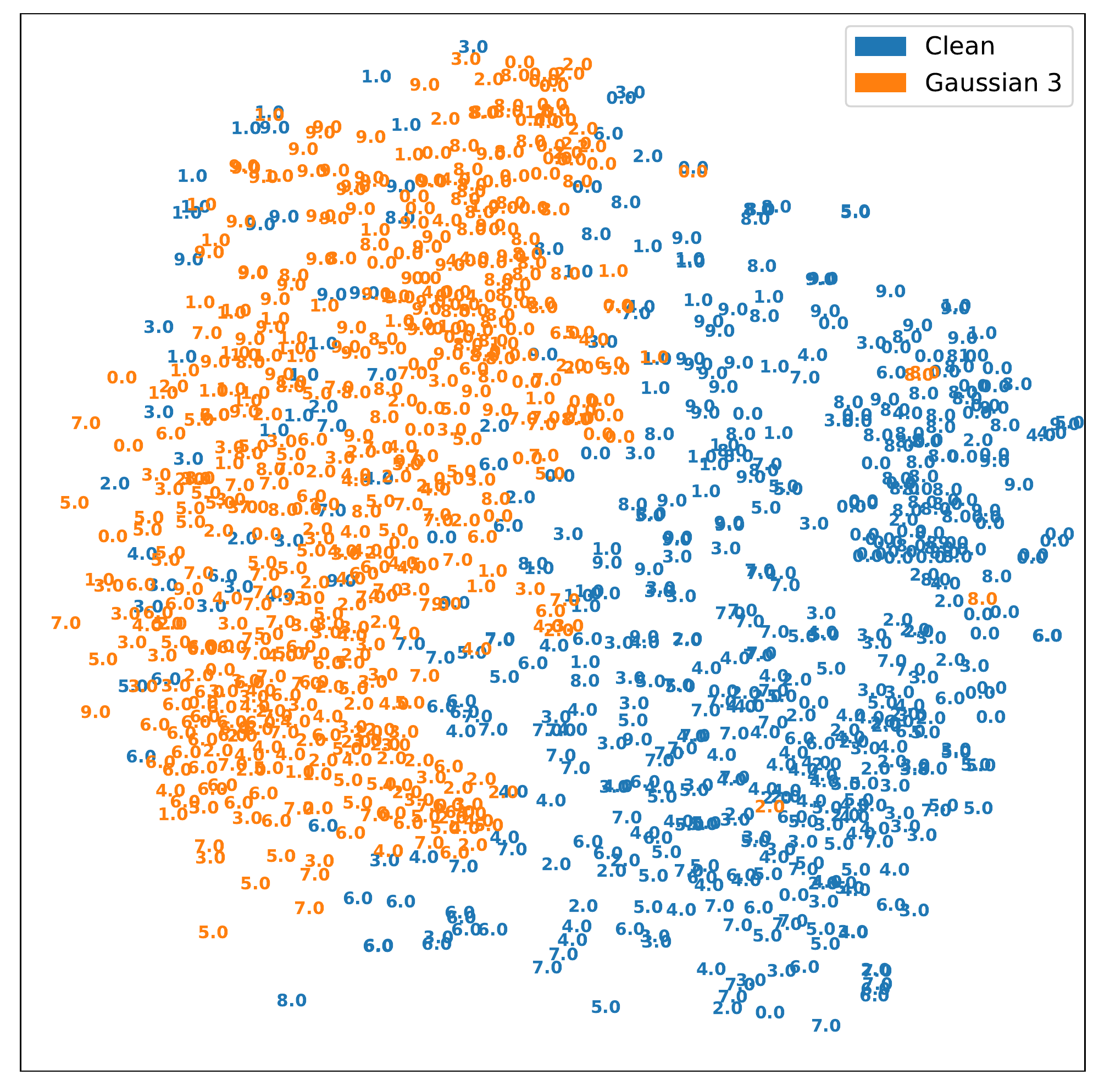}}
  \subfigure[Exit 3]{\includegraphics[width=0.48\columnwidth]{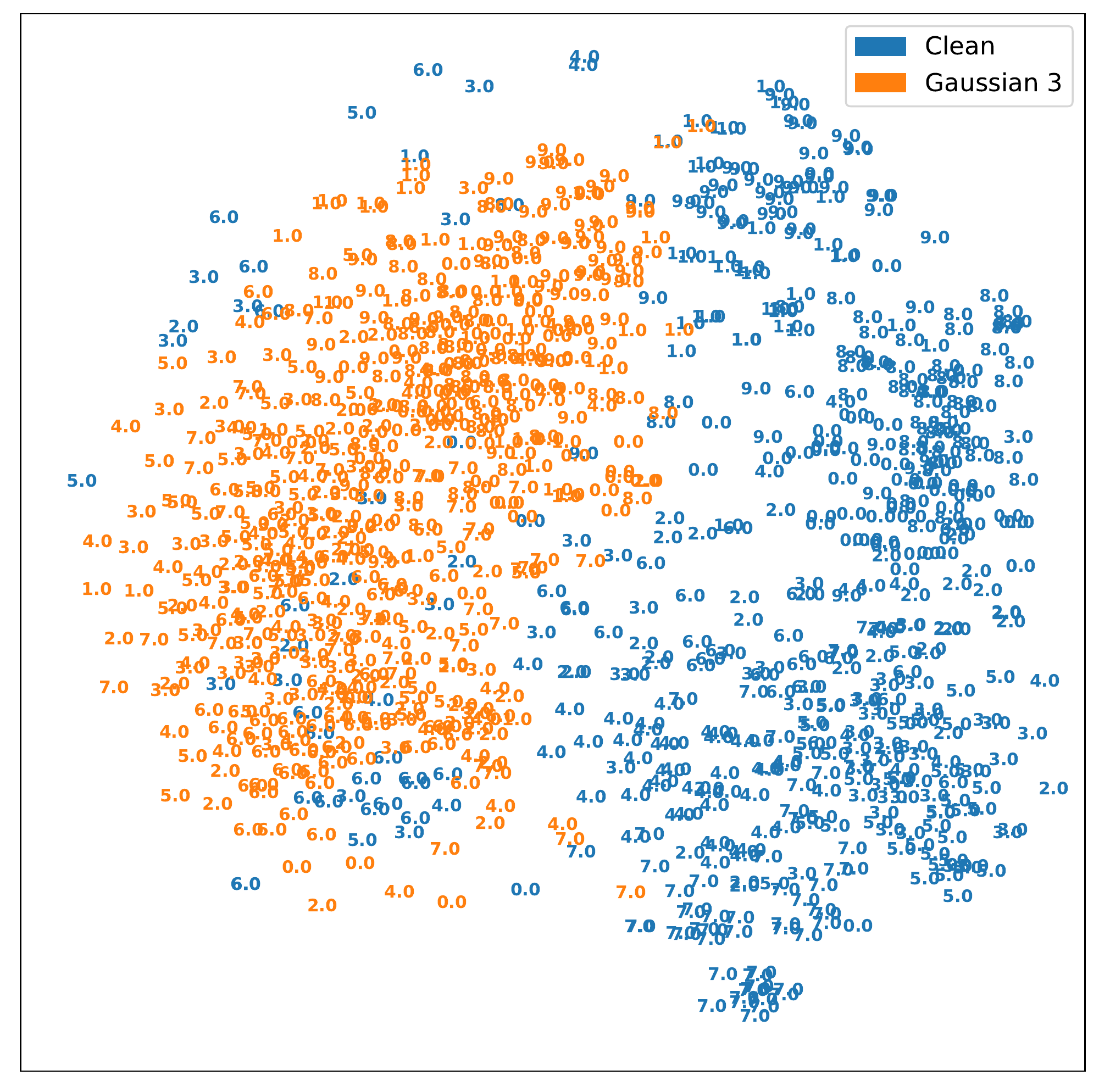}}
  \subfigure[Exit 5]{\includegraphics[width=0.48\columnwidth]{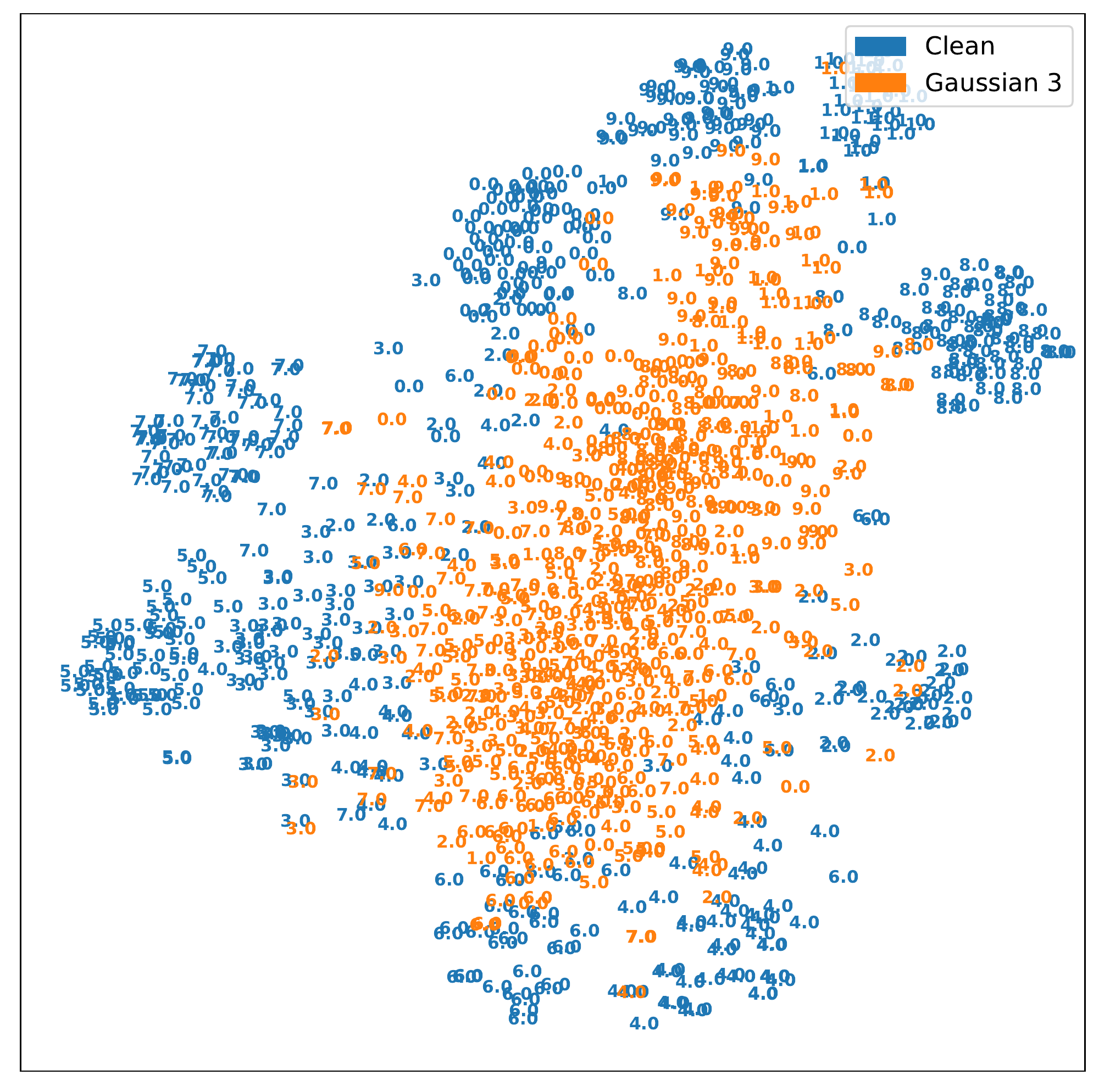}}
  \subfigure[Exit 7]{\includegraphics[width=0.48\columnwidth]{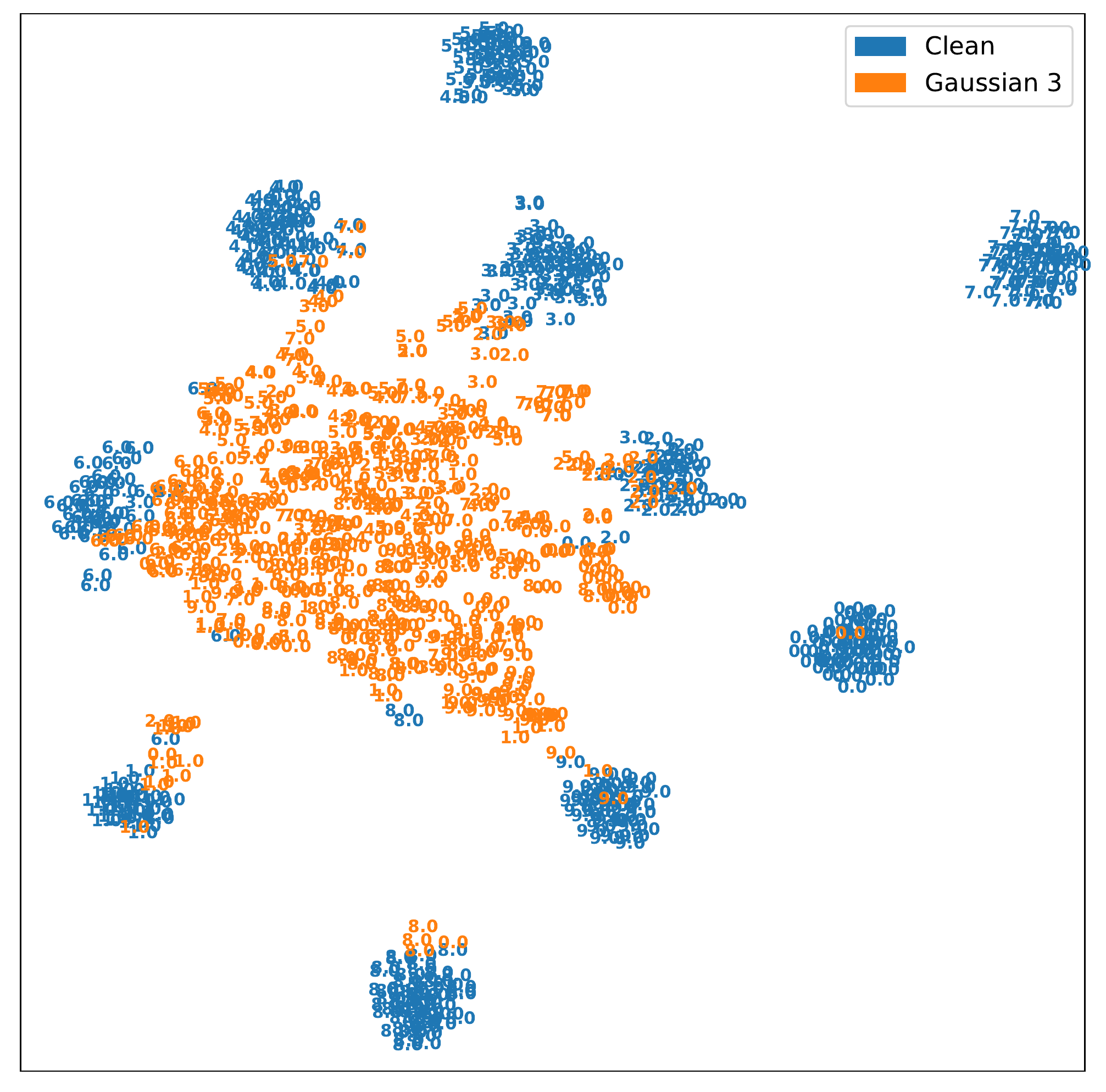}}
  }
  \caption{(Best viewed in color.) t-SNE embeddings of the penultimate layer of exits 1, 3, 5 and 7 (last layer), for clean and corrupted data (Gaussian noise with severity 3) of a ResNet-56 model trained with CIFAR-10. Each sample in the plot is denoted as the true label for the sample. Lack of clustering in earlier layers for corrupted data and incorrect classes from clean and corrupted data being closer makes the NN-based early-exit strategy under think.
  }
  \label{fig:knn_tsne}
\end{figure*}

\begin{figure*}[tb]
  \centering{
  \subfigure[VGG on CIFAR-10]{\includegraphics[width=0.24\columnwidth]{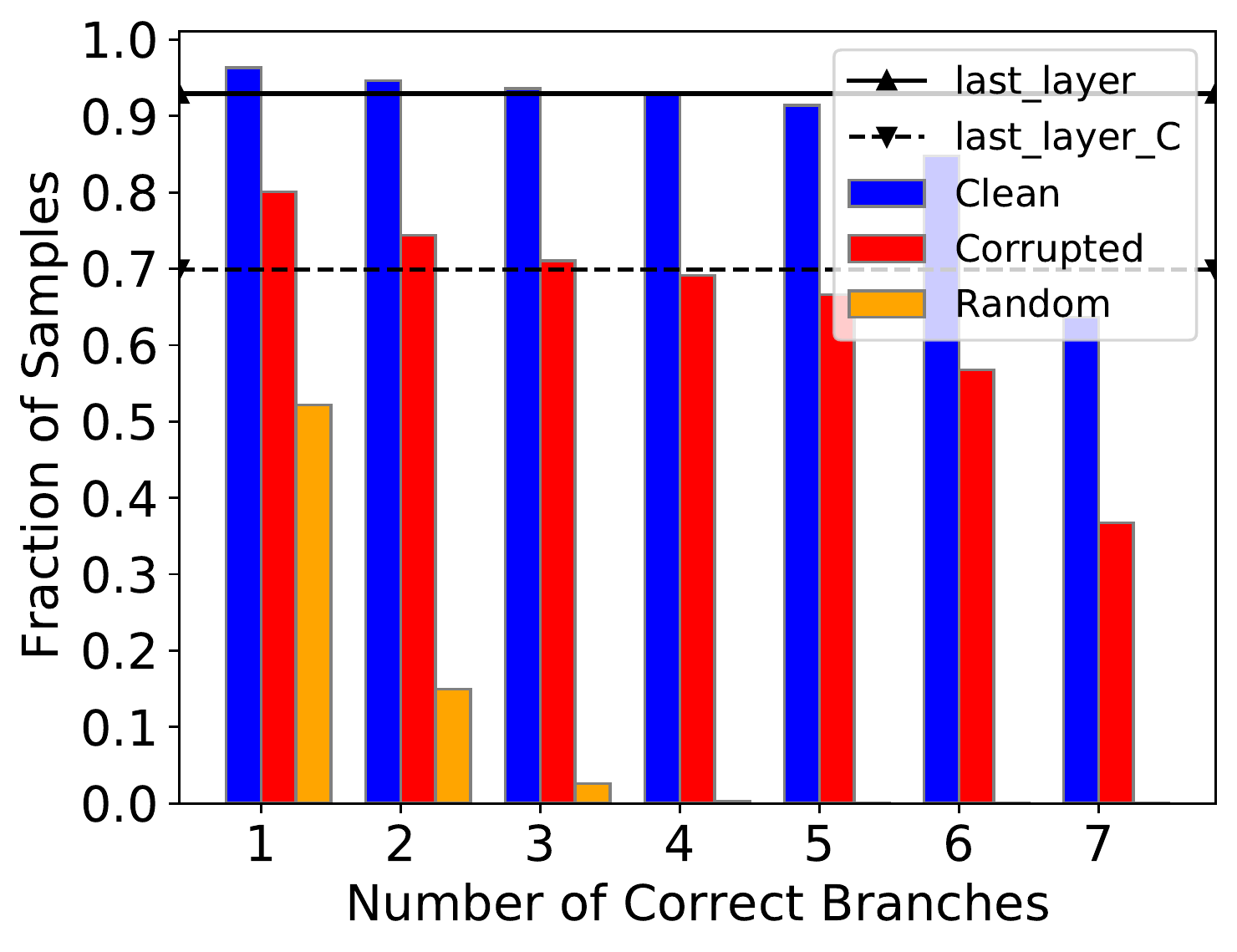}}
  \subfigure[ResNet on CIFAR-10]{\includegraphics[width=0.24\columnwidth]{Images/resnet_cifar10_barplot_num_exit_sum.pdf}}
  \subfigure[VGG on CIFAR-100]{\includegraphics[width=0.24\columnwidth]{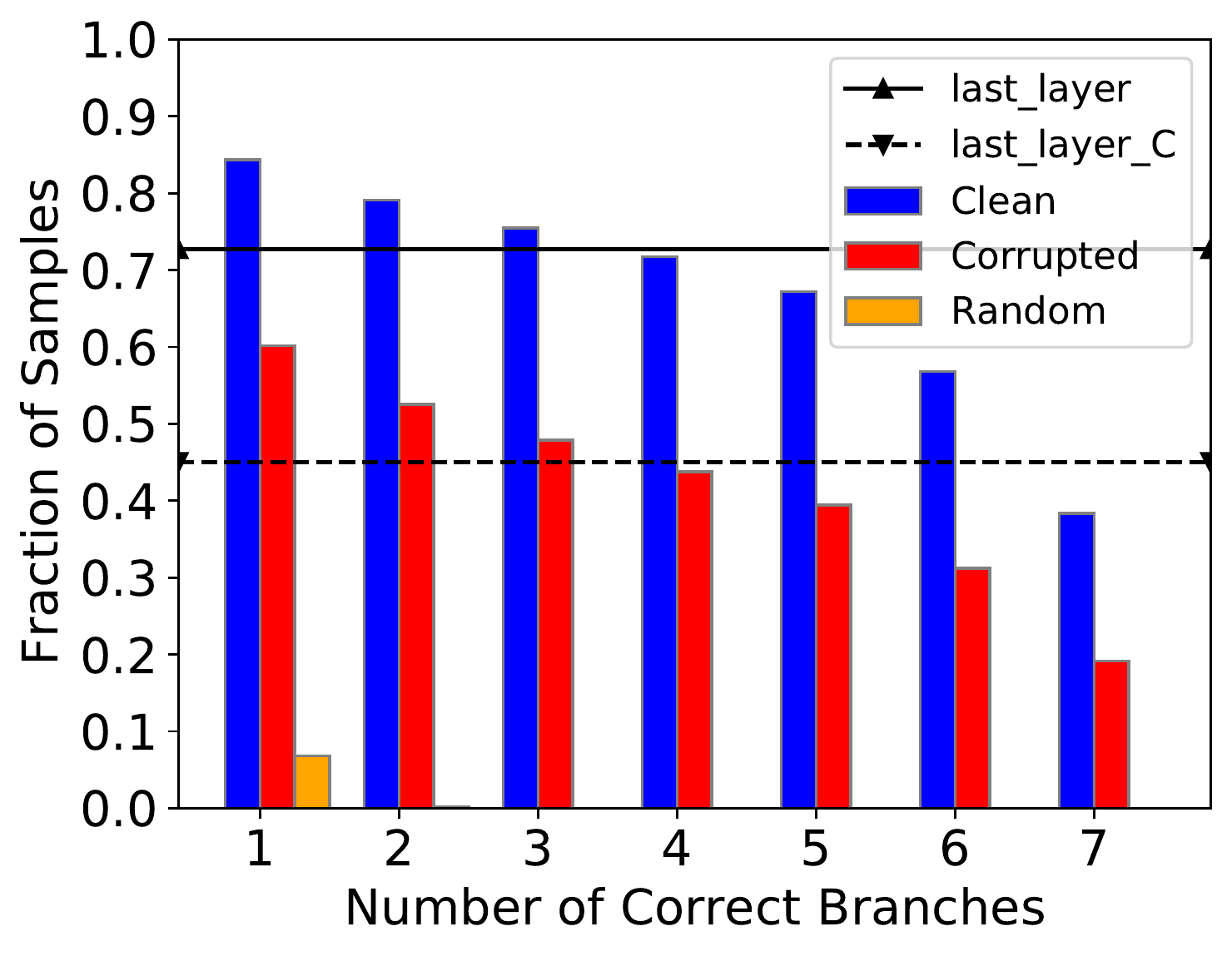}}
  \subfigure[ResNet on CIFAR-100]{\includegraphics[width=0.24\columnwidth]{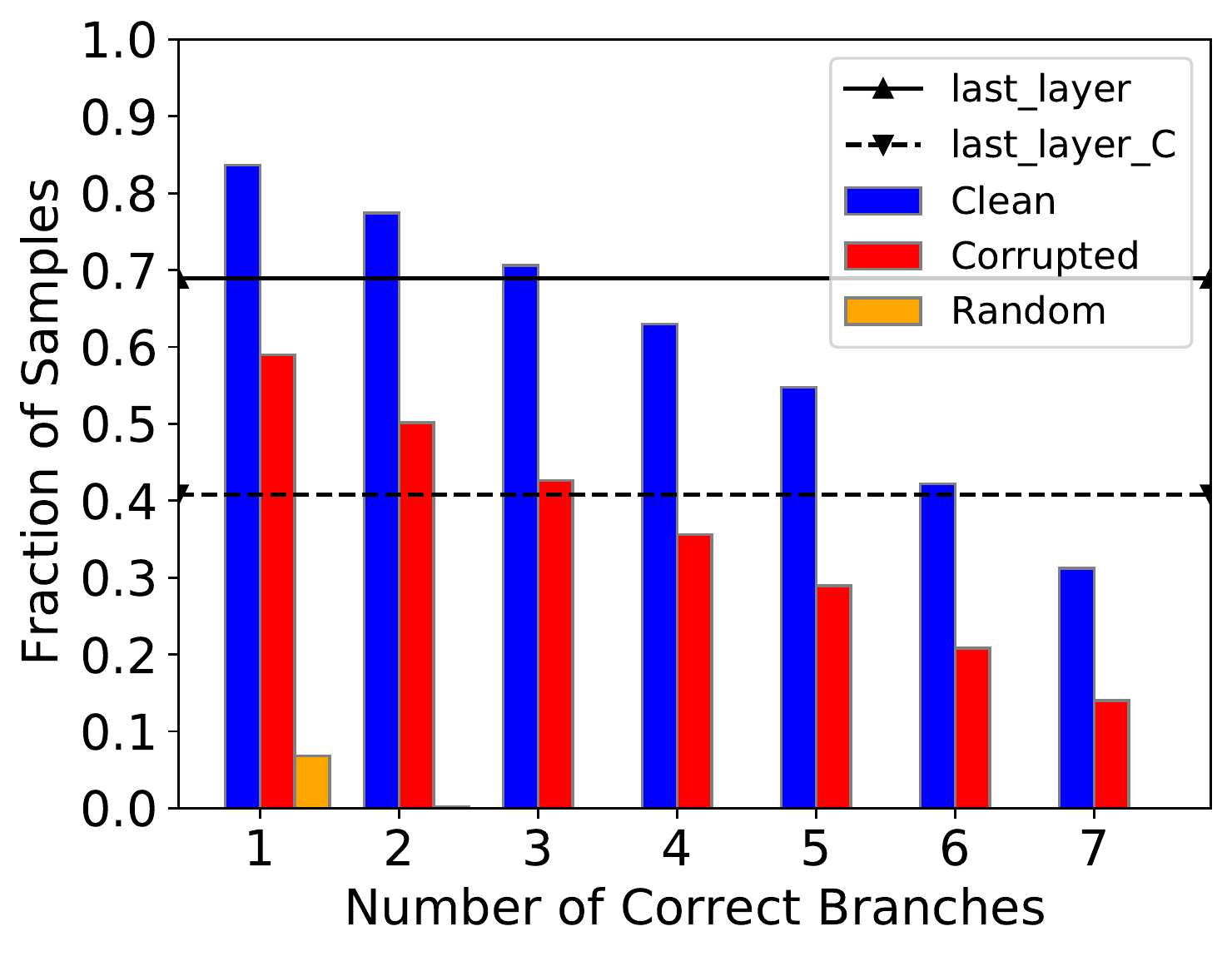}}
  }
  \caption{Correct predictions of a sample at multiple exits in MEMs with two different architectures (VGG-16/ResNet-56) on clean and corrupted CIFAR-10/100 datasets demonstrates the potential of early-exiting at improving the accuracy of DNNs compared to exiting only at the last layer. 
  }
  \label{fig:correct_pred_all}
\end{figure*}

\end{document}